%% file: bare_jrnl_compsoc.tex
\documentclass[10pt,journal,compsoc]{IEEEtran}
\ifCLASSOPTIONcompsoc
  \usepackage[nocompress]{cite}
\else
  \usepackage{cite}
\fi
\ifCLASSINFOpdf
\else
\fi
\usepackage{epsfig}
\usepackage{epstopdf}
\usepackage{graphicx}
\usepackage{amsmath}
\usepackage{bm}
\usepackage{amssymb}
\usepackage{color}
\usepackage[english]{babel}

\usepackage{algorithm} 
\usepackage{algpseudocode} 
\usepackage{booktabs}
\usepackage{verbatim}
\usepackage{gensymb}
\usepackage{pgffor}
\usepackage{tikz}
\usepackage{hyperref}

\usepackage{xspace}
\usepackage{multirow}
\usepackage{enumitem}
\usepackage{rotating}

\makeatletter
\DeclareRobustCommand\onedot{\futurelet\@let@token\@onedot}
\def\@onedot{\ifx\@let@token.\else.\null\fi\xspace}

\def\eg{{e.g}\onedot} 
\def\ie{{i.e}\onedot} 
 
\def\etc{{etc}\onedot} \def\vs{{vs}\onedot}
\def\wrt{w.r.t\onedot} 
\def\etal{{et al}\onedot}
\makeatother

\makeatletter
\let\MYcaption\@makecaption
\makeatother

\usepackage[font=footnotesize,labelformat=simple]{subcaption}

\makeatletter
\let\@makecaption\MYcaption
\makeatother

\makeatletter
\def\zapcolorreset{\let\reset@color\relax\ignorespaces}
\def\colorrows#1{\noalign{\aftergroup\zapcolorreset#1}\ignorespaces}
\makeatother

\begin{document}
%
\title{Learning 3D Human Shape and Pose from Dense Body Parts}
%
%
%
%

\author{Hongwen Zhang, Jie Cao, Guo Lu, 
Wanli Ouyang,~\IEEEmembership{Senior Member,~IEEE,} and
Zhenan Sun,~\IEEEmembership{Senior Member,~IEEE}
\IEEEcompsocitemizethanks{\IEEEcompsocthanksitem H. Zhang, J. Cao, and Z. Sun are with 
	CRIPAC, NLPR,
	Institute of Automation, Chinese Academy of Sciences, Beijing 100190,
	and aslo with the University of Chinese Academy of Sciences,
	Beijing 101408, China.
	E-mail: \mbox{hongwen.zhang@cripac.ia.ac.cn};
	\mbox{jie.cao@cripac.ia.ac.cn};
	\mbox{znsun@nlpr.ia.ac.cn}. (Corresponding author: Zhenan Sun.)
\IEEEcompsocthanksitem G. Lu is with the Beijing Institute of Technology, Beijing 100081, China. E-mail: \mbox{guo.lu@bit.edu.cn}.
\IEEEcompsocthanksitem W. Ouyang is with the University of Sydney, NSW 2006, Australia. E-mail: \mbox{wanli.ouyang@sydney.edu.au}.
}
}

%
%

\markboth{IEEE Transactions on Pattern Analysis and Machine Intelligence}%
{Zhang \MakeLowercase{\textit{et al.}}: Learning 3D Human Shape and Pose from Dense Body Parts}
%



\IEEEtitleabstractindextext{%
\begin{abstract}
Reconstructing 3D human shape and pose from monocular images is challenging despite the promising results achieved by the most recent learning-based methods. The commonly occurred misalignment comes from the facts that the mapping from images to the model space is highly non-linear and the rotation-based pose representation of body models is prone to result in the drift of joint positions. In this work, we investigate learning 3D human shape and pose from dense correspondences of body parts and propose a Decompose-and-aggregate Network (DaNet) to address these issues. DaNet adopts the dense correspondence maps, which densely build a bridge between 2D pixels and 3D vertices, as intermediate representations to facilitate the learning of 2D-to-3D mapping. The prediction modules of DaNet are decomposed into one global stream and multiple local streams to enable global and fine-grained perceptions for the shape and pose predictions, respectively. Messages from local streams are further aggregated to enhance the robust prediction of the rotation-based poses, where a position-aided rotation feature refinement strategy is proposed to exploit spatial relationships between body joints. Moreover, a Part-based Dropout (PartDrop) strategy is introduced to drop out dense information from intermediate representations during training, encouraging the network to focus on more complementary body parts as well as neighboring position features. The efficacy of the proposed method is validated on both indoor and real-world datasets including Human3.6M, UP3D, COCO, and 3DPW, showing that our method could significantly improve the reconstruction performance in comparison with previous state-of-the-art methods. Our code is publicly available at \url{https://hongwenzhang.github.io/dense2mesh}.
\end{abstract}

\begin{IEEEkeywords}
3D human shape and pose estimation, decompose-and-aggregate network, position-aided rotation feature refinement, part-based dropout.
\end{IEEEkeywords}}

\maketitle

\IEEEdisplaynontitleabstractindextext

%
\IEEEpeerreviewmaketitle

\input{tex/Introduction.tex}

\input{tex/RelatedWork.tex}

\input{tex/Methodology.tex}

\input{tex/Experiments.tex}

\section{Conclusion}
\label{conlusion}
In this work, a Decompose-and-aggregate Network is proposed to learn 3D human shape and pose from dense correspondences of body parts with the decomposed perception, aggregated refinement, and part-based dropout strategies.
All these new designs contribute to better part-based learning and effectively improve the reconstruction performance by providing well-suited part perception, leveraging spatial relationships for part pose refinement, and encouraging the exploitation of complementary body parts. 
Extensive experiments have been conducted to validate the efficacy of key components in our method.
In comparison with previous ones, our network can produce more accurate results, while being robust to extreme poses, heavy occlusions, and incomplete human bodies, etc.
In future work, we may explore integrating dense refinement~\cite{guler2019holopose} to further improve the shape and pose recovery results.


%



\ifCLASSOPTIONcompsoc
  \section*{Acknowledgments}
\else
  \section*{Acknowledgment}
\fi

The authors would like to thank the associate editor and reviewers for their helpful comments to improve this manuscript.
This work was supported in part by the National Natural Science Foundation of China (Grant No. U1836217, 61806197) and the National Key Research and Development Program of China (Grant No. 2017YFC0821602).
This work was done when H. Zhang visited the University of Sydney with the support of the Joint Ph.D. Training Program of the University of Chinese Academy of Sciences.

\ifCLASSOPTIONcaptionsoff
  \newpage
\fi



\bibliographystyle{IEEEtran}
\bibliography{IEEEabrv,egbib}
%



%




\end{document}

%% file: tex/Introduction.tex
\IEEEraisesectionheading{\section{Introduction}\label{sec:introduction}}

\begin{figure}[t]
	\centering
	\begin{subfigure}[b]{0.1\textwidth}
		\centering
		\includegraphics[height=20mm]{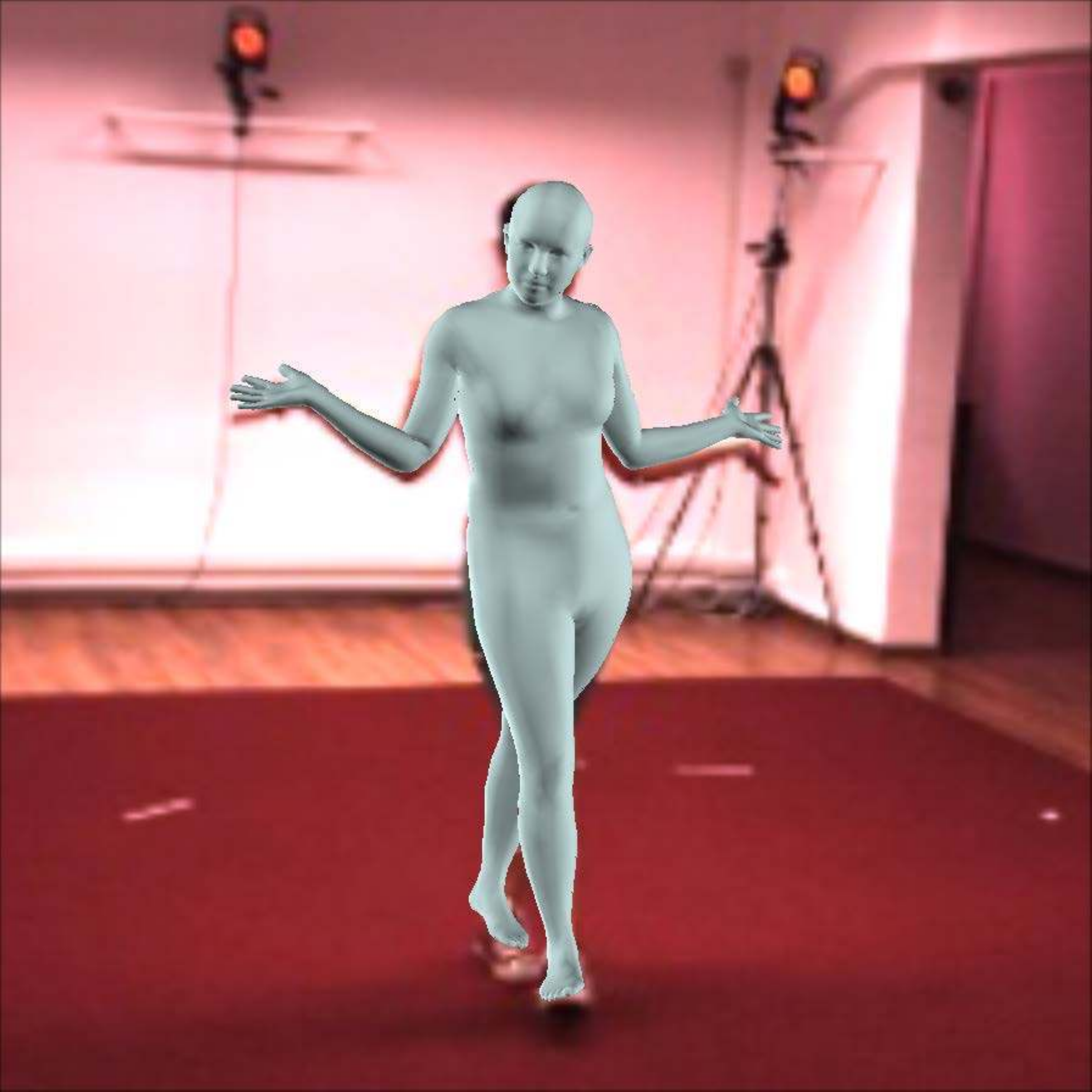}
		\caption{ }
		\label{fig:img_err_result}
	\end{subfigure}
    \begin{subfigure}[b]{0.32\textwidth}
		\centering
		\includegraphics[height=20mm]{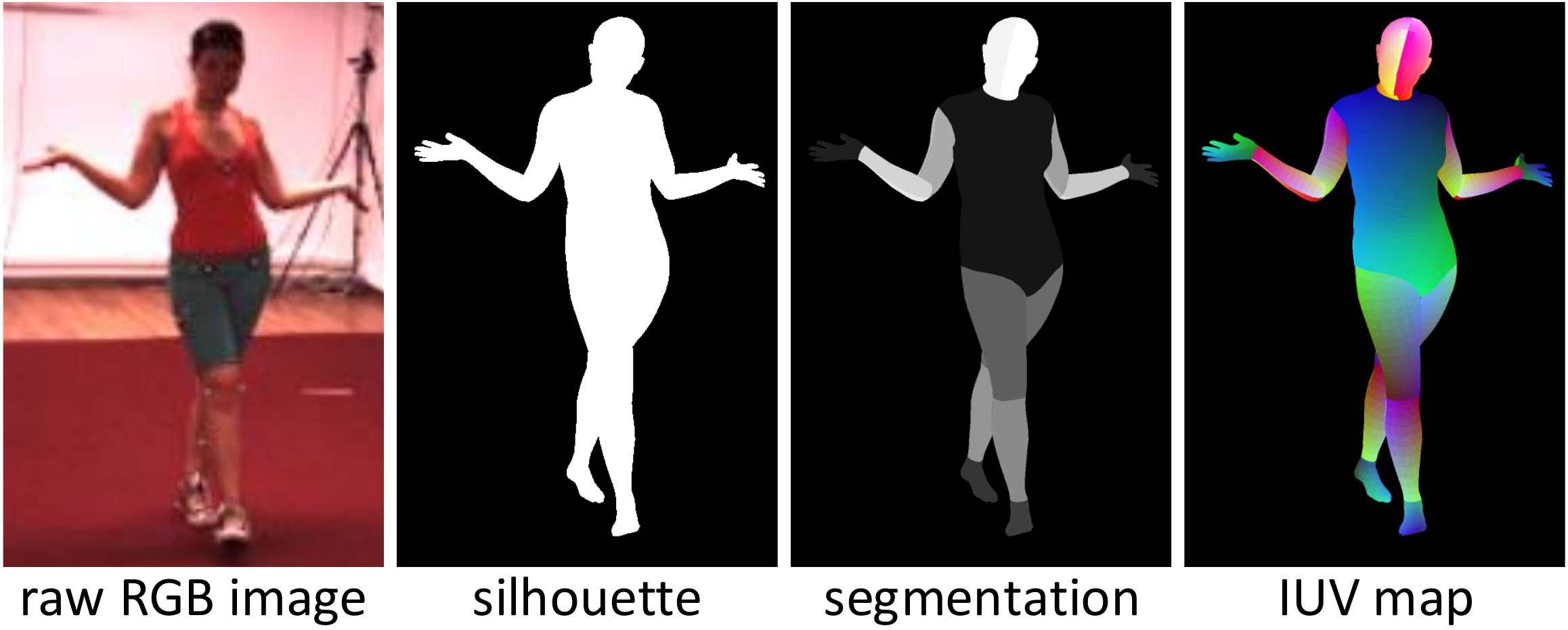}
		\caption{ }
		\label{fig:cpr_img_seg_uvi}
    \end{subfigure}

	\begin{subfigure}[b]{0.12\textwidth}
		\centering
		\includegraphics[height=22mm]{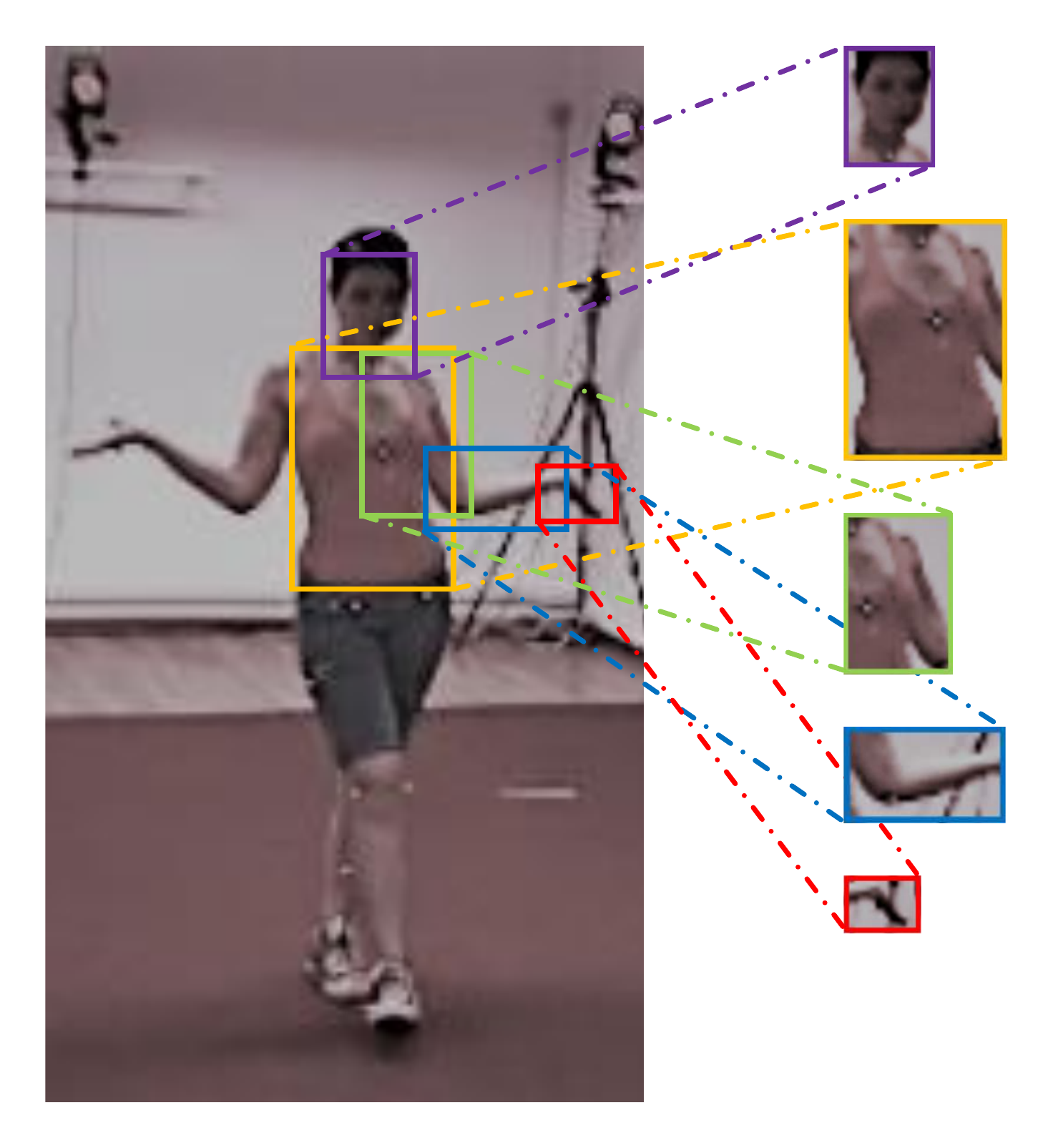}
		\caption{ }
		\label{fig:part_scale}
	\end{subfigure}
	\begin{subfigure}[b]{0.32\textwidth}
		\centering
		\includegraphics[height=22mm]{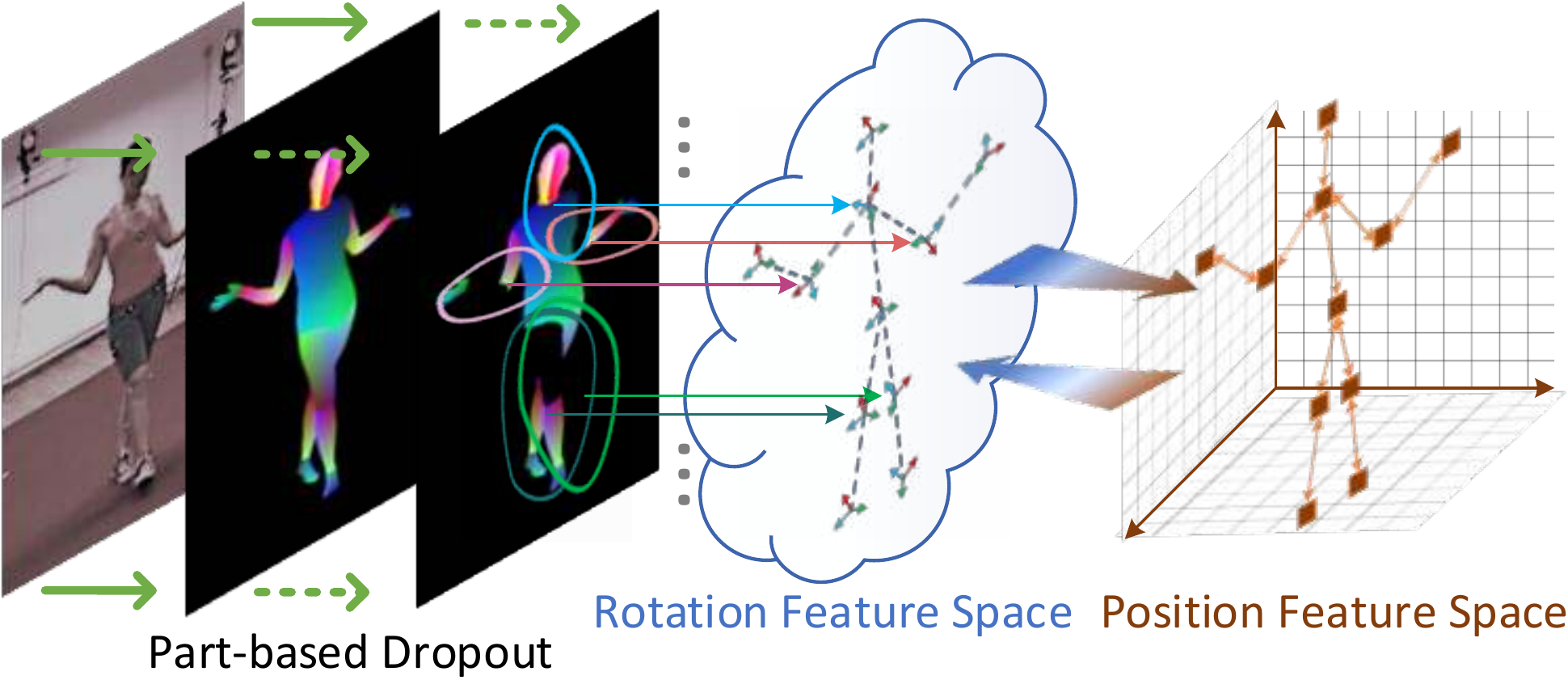}
		\caption{ }
		\label{fig:main_idea}
	\end{subfigure}
	\vspace{-3mm}
	\caption{Illustration of our main ideas. (a) A human image with a parametric body model.
	(b) Comparison of the raw RGB image, silhouette, segmentation, and IUV map. (c) Local visual cues are crucial for the perception of joint rotations. (d) Our DaNet learns 3D human shape and pose from IUV maps with decomposed perception, aggregated refinement, and part-based dropout strategies. 
	}
	\label{fig:motivation}
\end{figure}

\IEEEPARstart{R}{econstructing} human shape and pose from a monocular image is an appealing yet challenging task, which typically involves the prediction of the camera and parameters of a statistical body model (\eg the most commonly used SMPL~\cite{loper2015smpl} model). Fig.~\ref{fig:img_err_result} shows an example of the reconstructed result.
The challenges of this task come from the fundamental depth ambiguity, the complexity and flexibility of human bodies, and variations in clothing and viewpoint, \etc.
Classic optimization-based approaches~\cite{bogo2016keep,lassner2017unite} fit the SMPL model to 2D evidence such as 2D body joints or silhouettes in images, which involve complex non-linear optimization and iterative refinement.
Recently, regression-based approaches~\cite{tung2017self,kanazawa2018end,pavlakos2018learning,omran2018neural} integrate the SMPL model within neural networks and predict model parameters directly in an end-to-end manner.

Though great progress has been made, the direct prediction of the body model from the image space is still complex and difficult even for deep neural networks.
In this work, we propose to adopt IUV maps as intermediate representations to facilitate the learning of the mapping from images to models.
As depicted in Fig.~\ref{fig:cpr_img_seg_uvi}, compared with other 2D representations ~\cite{tung2017self,pavlakos2018learning,omran2018neural}, the IUV map could provide more rich information, because it encodes the dense correspondence between foreground pixels on 2D images and vertices on 3D meshes.
Such a dense semantic map not only contains essential information for shape and pose estimation from RGB images, but also eliminates the interference of unrelated factors such as appearance, clothing, and illumination variations.

The representation of 3D body model~\cite{anguelov2005scape,loper2015smpl} can be factorized into the shape and pose components, depicting the model at different scales.
The body shape gives an identity-dependent description about the model, while the body pose provides more detailed descriptions about the rotation of each body joint.
Previous regression-based methods~\cite{kanazawa2018end,omran2018neural} typically predict them simultaneously using global information from the last layer of the neural network.
We observe that the detailed pose of body joints should be captured by local visual cues instead of global information.
As shown in Fig.~\ref{fig:part_scale}, we can estimate the rotations of those visible body joints only based on local visual cues, while the information from other body joints and background regions would be irrelevant.

For the rotation-based pose representation of commonly used body models~\cite{anguelov2005scape,loper2015smpl}, small rotation errors accumulated along the kinematic chain could lead to large drift of position at the leaf joint.
Moreover, the rotation estimation is error-prone for those occluded body joints since their perceptions are less reliable under occlusions.
Hence, it is crucial to utilize information from visible body joints and the prior about the structure of human bodies.
As shown in previous work~\cite{chen2014articulated,chu2016structured}, the structural information at the feature level is helpful for more robust and accurate pose estimation.
However, it is non-trivial to apply these feature refinement methods to our case due to the weak correlation between rotation-based poses of different joints.
For instance, the shoulder, elbow, and wrist are three consecutive body joints, and one can hardly infer the relative rotation of wrist \wrt the elbow given the relative rotation of elbow \wrt the shoulder. 
On the other hand, we observe that the 3D locations of body joints have stronger correlations than the rotation of body joints. 
For instance, the positions of shoulder, elbow, and wrist are strongly constrained by the length of the arm.

Based on the observations above, we propose a Decompose-and-aggregate Network (DaNet) to learn 3D human shape and pose from dense correspondences of body parts. 
As illustrated in Fig.~\ref{fig:main_idea}, DaNet utilizes IUV maps as the intermediate information for more efficient learning, and decomposes the prediction modules into multiple streams considering that the prediction of different parameters requires the receptive fields with different sizes.
To robustly predict the rotations of body joints, DaNet aggregates messages from different streams and refines the rotation features via an auxiliary position feature space to exploit the spatial relationships between body joints.
For better generalization, a Part-based Dropout (PartDrop) strategy is further introduced to drop out dense information from intermediate representations during training, which could effectively regularize the network and encourage it to learn features from complementary body parts and leverage information from neighboring body joints.
As will be validated in our experiments, all the above new designs could contribute to better part-based learning and improve the reconstruction performance.
To sum up, the main contributions in this work are listed as follows.

\begin{itemize}[leftmargin=*]
	\item We comprehensively study the effectiveness of adopting the IUV maps in both global and local scales, which contains densely semantic information of body parts, as intermediate representations for the task of 3D human pose and shape estimation.
	\item Our reconstruction network is designed to have decomposed streams to provide global perception for the camera and shape prediction while detailed perception for pose prediction of each body joint.
	\item A part-based dropout strategy is introduced to drop dense information from intermediate representations during training. 
	Such a strategy can encourage the network to learn features from complementary body parts, which also has the potential for other structured image understanding tasks.
	\item A position-aided rotation feature refinement strategy is proposed to aggregate messages from different part features.
    It is more efficient to exploit the spatial relationship in an auxiliary position feature space since the correlations between position features are much stronger. 
\end{itemize}

An early version of this work appeared in~\cite{zhang2019danet}. We have made significant extensions to our previous work in three main aspects.
First, the methodology is improved to be more accurate and robust thanks to several new designs, including the part-based dropout strategy for better generalization performance and the customized graph convolutions for more efficient and better feature mapping and refinement.
Second, more extensive evaluations and comparisons are included to validate the effectiveness of our method, including evaluations on additional datasets and comparisons of the reconstruction errors across different human actions and model surface areas.
Third, more discussions are provided in our ablation studies, including comprehensive evaluations on the benefit of adopting IUV as intermediate representations and in-depth analyses on the refinement upon the rotation feature space and position feature space.

The remainder of this paper is organized as follows. Section~\ref{related_work} briefly reviews previous work related to ours. Section~\ref{preliminary} provides preliminary knowledge about the SMPL model and IUV maps. Details of the proposed network are presented in Section~\ref{methodology}. Experimental results and analyses are included in Section~\ref{experiments}. Finally, Section~\ref{conlusion} concludes the paper.

%% file: tex/RelatedWork.tex
\section{Related Work}
\label{related_work}

\subsection{3D Human Shape and Pose Estimation}
Early pioneering work on 3D human model reconstruction mainly focuses on the optimization of the fitting process.
Among them, \cite{sigal2008combined,guan2009estimating} fit the body model SCAPE~\cite{anguelov2005scape} with the requirement of ground truth silhouettes or manual initialization.
Bogo \etal~\cite{bogo2016keep} introduce the optimization method SMPLify and make the first attempt to automatically fit the SMPL model to 2D body joints by leveraging multiple priors.
Lassner \etal~\cite{lassner2017unite} extend this method and improve the reconstruction performance by incorporating silhouette information in the fitting procedure. 
These optimization-based methods typically rely on accurate 2D observations and the prior terms imposed on the shape and pose parameters, making the procedure time-consuming and sensitive to the initialization.
Alternatively, recent regression-based methods employ neural networks to predict the shape and pose parameters directly and learn the priors in a data-driven manner.
These efforts mainly focus on several aspects including intermediate representation leveraging, architecture designs, structural information modeling, and re-projection loss designs, \etc.
Our work makes contributions to the first three aspects above and is also complementary to the work focusing on the re-projection loss designs~\cite{tung2017self,guler2019holopose,rong2019delving}, reconstruction from videos or multi-view images~\cite{kanazawa2019learning,arnab2019exploiting,liang2019shape,pavlakos2019texturepose,kocabas2020vibe}, and detailed or holistic body model learning~\cite{joo2018total,pavlakos2019expressive,zhu2019detailed}.

\subsubsection{Intermediate Representation}
The recovery of the 3D human pose from a monocular image is challenging.
Common strategies use intermediate estimations as the proxy representation to alleviate the difficulty.
These methods can benefit from existing state-of-the-art networks for lower-level tasks.
For the recovery of 3D human pose or human model, 2D joint positions~\cite{martinez2017simple,nie2017monocular,moreno20173d,lee2018propagating}, silhouette~\cite{dibra2016hs,pavlakos2018learning,smith2019towards}, segmentation~\cite{omran2018neural}, depth maps~\cite{shotton2012efficient,gabeur2019moulding}, joint heatmaps~\cite{tekin2017learning,tung2017self,pavlakos2018learning}, volumetric representation~\cite{pavlakos2017coarse,varol2018bodynet,jackson20183d,zheng2019deephuman}, and 3D orientation fields~\cite{luo2018orinet,xiang2019monocular} are adopted in literature as intermediate representations to facilitate the learning task.
Though the aforementioned representations are helpful for the task, detailed information contained within body parts is missing in these coarse representations, which becomes the bottleneck for fine-grained prediction tasks.
Recently, DensePose~\cite{alp2018densepose} regresses the IUV maps directly from images, which provides the dense correspondence mapping from the image to the human body model.
However, the 3D model cannot be directly retrieved from such a 2.5D projection.
In our work, we propose to adopt such a densely semantic map as the intermediate representation for the task of 3D human shape and pose estimation.
To the best of our knowledge, we are among the first attempts~\cite{kolotouros2019convolutional,rong2019delving,xu2019denserac} to leverage IUV maps for 3D human model recovery.
In comparison, the major differences between concurrent efforts and ours lie in three aspects:
1) \cite{kolotouros2019convolutional,rong2019delving,xu2019denserac} obtain IUV predictions from a pretrained network of DensePose~\cite{alp2018densepose}, while our work augments the annotations of 3D human pose datasets with the rendered ground-truth IUV maps and imposes dense supervisions on the intermediate representations;
2) \cite{kolotouros2019convolutional,rong2019delving,xu2019denserac} only leverage global IUV maps, while our work exploits using IUV maps in both global and local scales;
3) DenseRaC~\cite{xu2019denserac} resorts to involving more synthetic IUV maps as additional training data while our work introduces the part-based dropout upon IUV maps to improve generalization.
We believe these concurrent work complement each other and enrich the research community.

\subsubsection{Architecture Design}
Existing approaches to 3D human shape and pose estimation have designed a number of network architectures for more effective learning of the highly nonlinear image-to-model mapping.
Tan \etal~\cite{tan2018indirect} develop an encoder-decoder based framework where the decoder learns the SMPL-to-silhouette mapping from synthetic data and the encoder learns the image-to-SMPL mapping with the decoder frozen.
Kanazawa \etal~\cite{kanazawa2018end} present an end-to-end framework HMR to reconstruct the SMPL model directly from images using a single CNN with an iterative regression module.
Kolotouros \etal~\cite{kolotouros2019learning} enhance HMR with the fitting process of SMPLify~\cite{bogo2016keep} to incorporate regression- and optimization-based methods.
Pavlakos \etal~\cite{pavlakos2018learning} propose to predict the shape and pose parameters from the estimated silhouettes and joint locations respectively.
Sun \etal~\cite{sun2019human} also leverage joint locations and further involve deep features into the prediction process.
Instead of regressing the shape and pose parameters directly, Kolotouros \etal~\cite{kolotouros2019convolutional} employ a Graph CNN~\cite{kipf2017semi} to regress the 3D coordinates of the human mesh vertices, while Yao \etal~\cite{yao2019densebody} regress the 3D coordinates in the form of an unwrapped position map.
All aforementioned regression-based methods predict the pose in a global manner.
In contrast, our DaNet predicts joint poses from multiple streams, hence the visual cues could be captured in a fine-grained manner.
Recently, Güler \etal~\cite{guler2019holopose} also introduce a part-based reconstruction method to predict poses from the deep features pooled around body joints.
In comparison, the pooling operation of our DaNet is performed on intermediate representations, enabling detailed perception for better pose feature learning.
Moreover, existing approaches for rotation-based pose estimation do not consider feature refinement, while DaNet includes an effective rotation feature refinement scheme for robust pose predictions.

\subsubsection{Structural Information Modeling}
Leveraging the articulated structure information is crucial for human pose modeling~\cite{pishchulin2013poselet,yang2011articulated,zuffi2015stitched}.
Recent deep learning-based approaches to human pose estimation~\cite{chen2014articulated,tompson2014joint,chu2016structured,fang2018learning,zhao2019semantic} incorporate the structured feature learning in their network architecture designs.
All these efforts exploit the relationship between the \emph{position features} of body joints and their feature refinement strategies are only validated on the position-based pose estimation problem.
Our work is complementary to them by investigating the refinement for \emph{rotation features} under the context of the rotation-based pose representation, which paves a new way to impose structural constraints upon rotation features.
Our solution aggregates the rotation features into a position feature space, where the aforementioned structural feature learning approaches could be easily applied.

For more geometrically reasonable pose predictions, different types of pose priors~\cite{akhter2015pose,zhou2016deep,sun2017compositional,zhou2017towards,yang20183d} are also employed as constraints in the learning procedure.
For instance, Akhter and Black~\cite{akhter2015pose} learn the pose prior in the form of joint angle constraints.
Sun \etal~\cite{sun2017compositional} design handcrafted constraints such as limb-lengths and their proportions.
Similar constraints are exploited in~\cite{zhou2017towards} under the weakly-supervised setting.
For the rotation-based pose representation in the SMPL model, though it inherently satisfies structure constraints such as limb proportions, the pose prior is still essential for better reconstruction performance.
SMPLify~\cite{bogo2016keep} imposes several penalizing terms on predicted poses to prevent unnatural results.
Kanazawa \etal~\cite{kanazawa2018end} introduce an adversarial prior for guiding the prediction to be realistic.
All these methods consider the pose prior at the \emph{output level}.
In our work, we will exploit the relationship at the \emph{feature level} for better 3D pose estimation in the SMPL model. 

\subsection{Regularization in Neural Networks}
Regularization is important to neural networks for better generalization performance.
A number of regularization techniques have been proposed to remove features from neural networks at different granularity levels.
Among them, dropout~\cite{srivastava2014dropout} is commonly used at the fully connected layers of neural networks to drop unit-wise features independently.
The introduction of dropout has inspired the development of other dropping out strategies with structured forms.
For instance, SpatialDropout~\cite{tompson2015efficient} drops channel-wise features across the entire feature map, while DropBlock~\cite{ghiasi2018dropblock} drops block-wise features in contiguous regions.
Different from these techniques, our PartDrop strategy drops part-wise features at the granularity level of semantic body parts.
Such a part-wise dropping strategy could remove patterns in a more structured manner and perform better in our learning task.
Moreover, our PartDrop strategy is applied on intermediate representations, which is also different from data augmentation methods such as Cutout~\cite{devries2017improved}.

%% file: tex/Methodology.tex
\begin{figure}[t]
	\begin{center}
    \begin{tabular}[b]{cc}
        \begin{tabular}[b]{c}
            \begin{subfigure}[b]{0.3825\textwidth}
        		\includegraphics[width=1.0\textwidth]{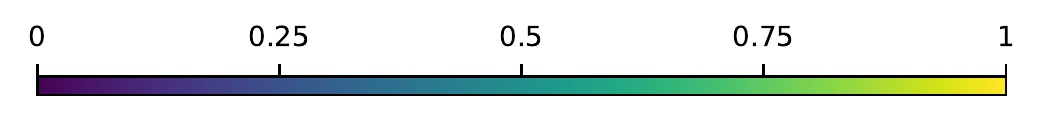}
        	\end{subfigure}\\
            \begin{subfigure}[b]{0.1071\textwidth}
    			\includegraphics[width=1\textwidth]{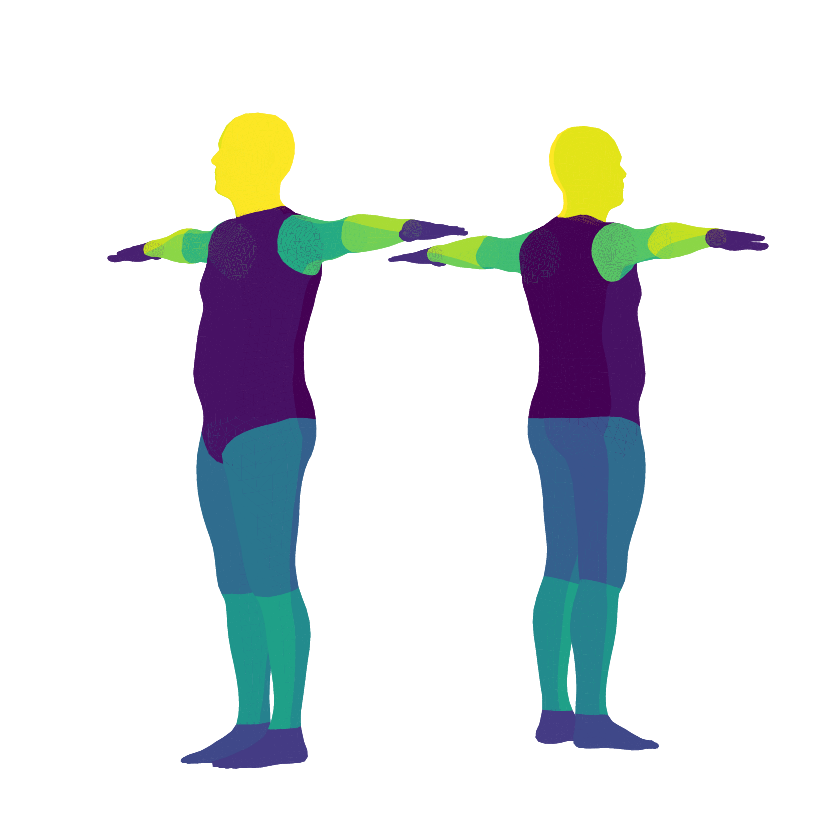}
    			\caption{}
    			\label{fig:smpl_i}
    		\end{subfigure}
    		\begin{subfigure}[b]{0.1071\textwidth}
    			\includegraphics[width=1\textwidth]{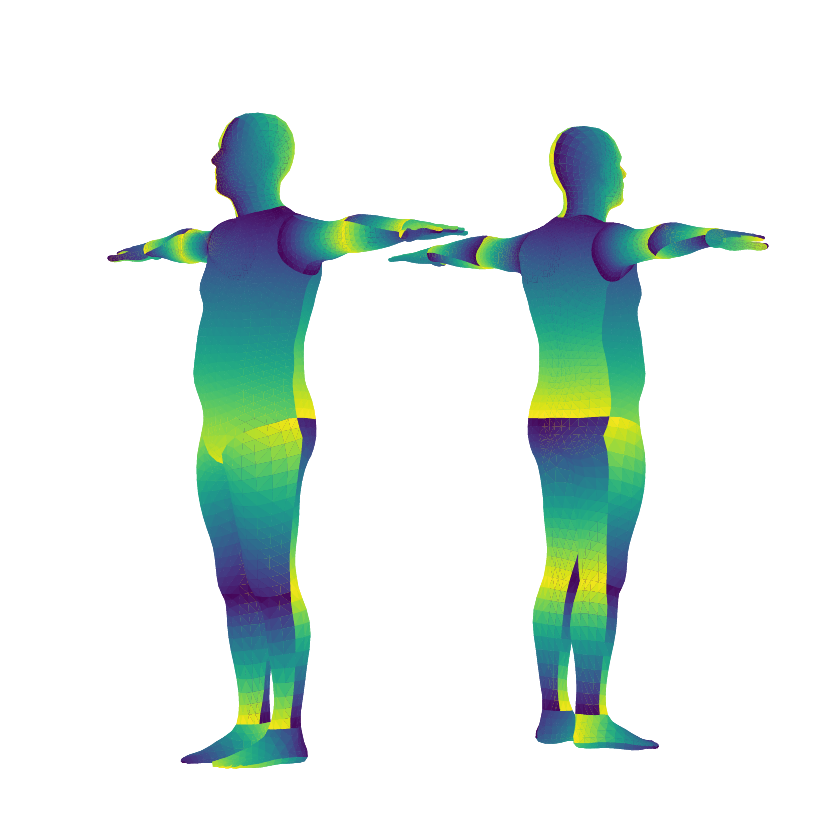}
    			\caption{}
    			\label{fig:smpl_u}
    		\end{subfigure}
    		\begin{subfigure}[b]{0.1071\textwidth}
    			\includegraphics[width=1\textwidth]{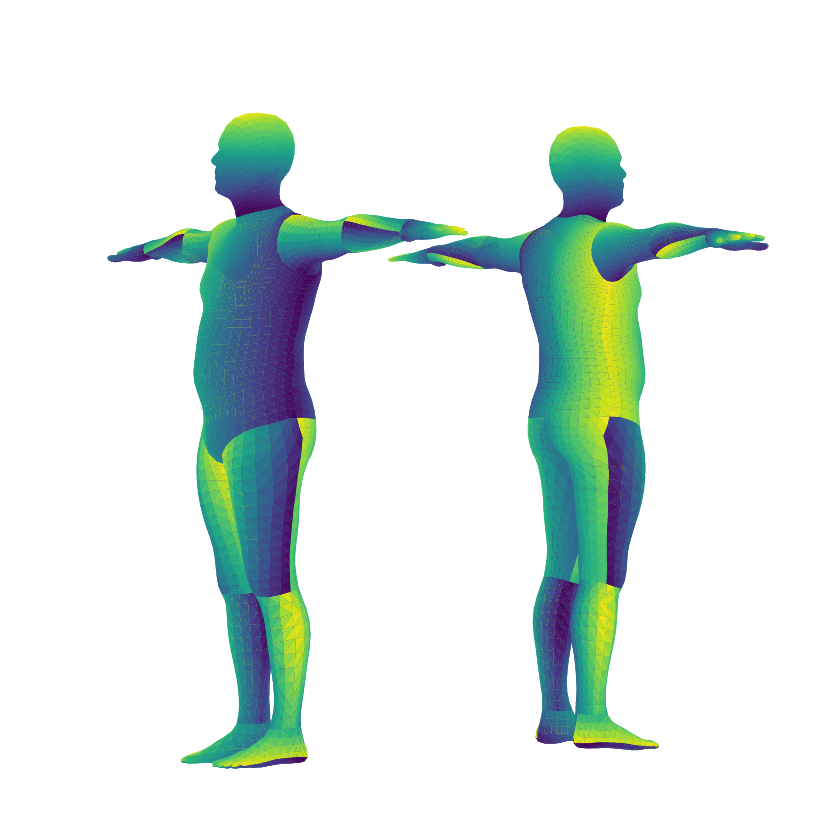}
    			\caption{}
    			\label{fig:smpl_v}
    		\end{subfigure}\\
    		\begin{subfigure}[b]{0.3213\textwidth}
    			\includegraphics[width=1\textwidth]{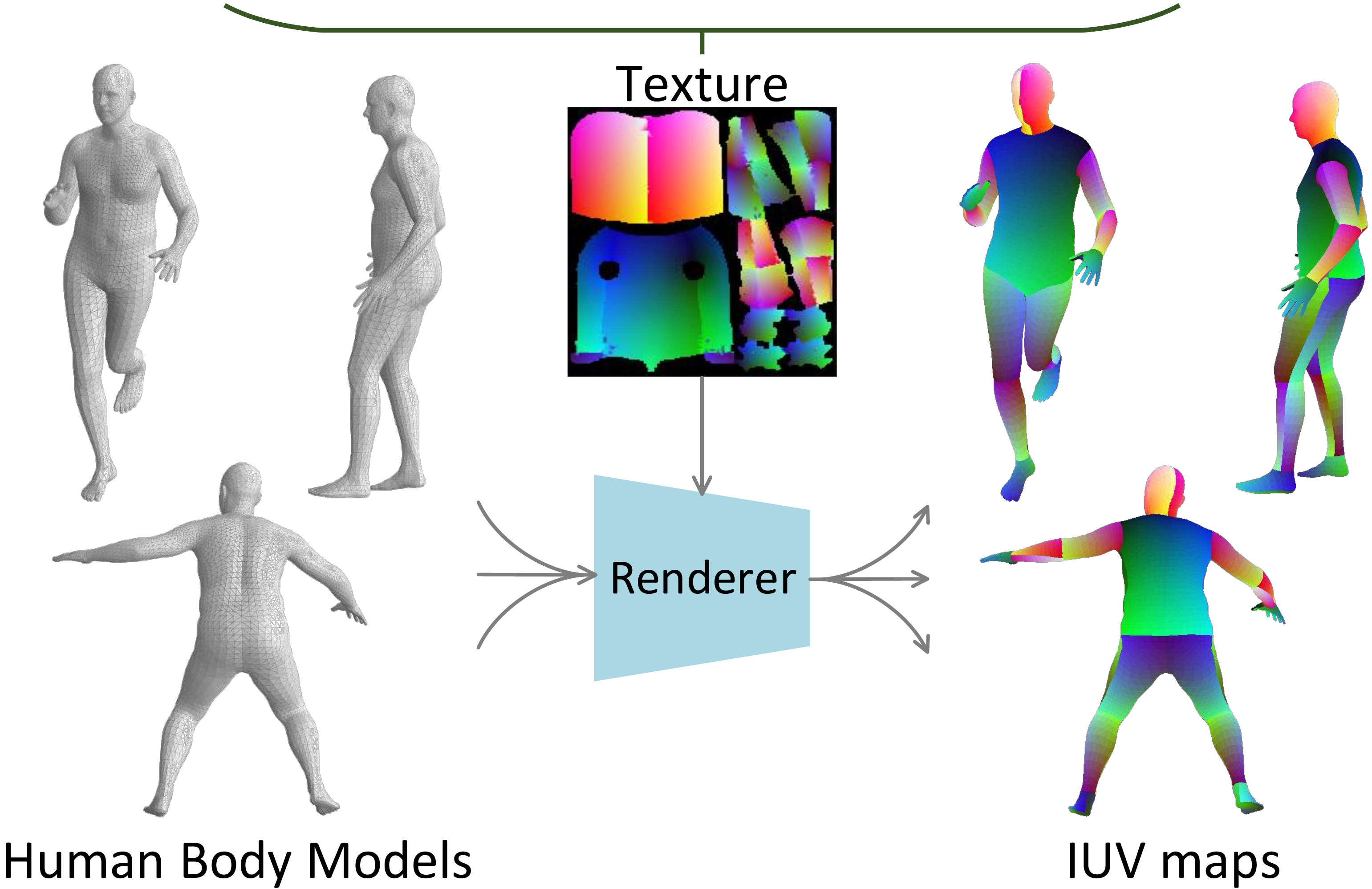}
    			\caption{}
    			\label{fig:render_vis}
    		\end{subfigure}
    	\end{tabular}
	\end{tabular}
	\end{center}
	\vspace{-5mm}
    \caption{Illustration of the preparation of ground truth IUV maps. (a)(b)(c) show the $\textit{Index}$, $U$, and $V$ values defined in DensePose~\cite{alp2018densepose}, respectively. Note that the original $\textit{Index}$ values (range from 1 to 24) are also normalized into the $[0, 1]$ interval. (d) Generation of ground truth IUV Maps for 3D human body models.
    }
\label{fig:smpl_uvi}
\end{figure}

\begin{figure*}[t]
	\begin{center}
		\includegraphics[height=55mm]{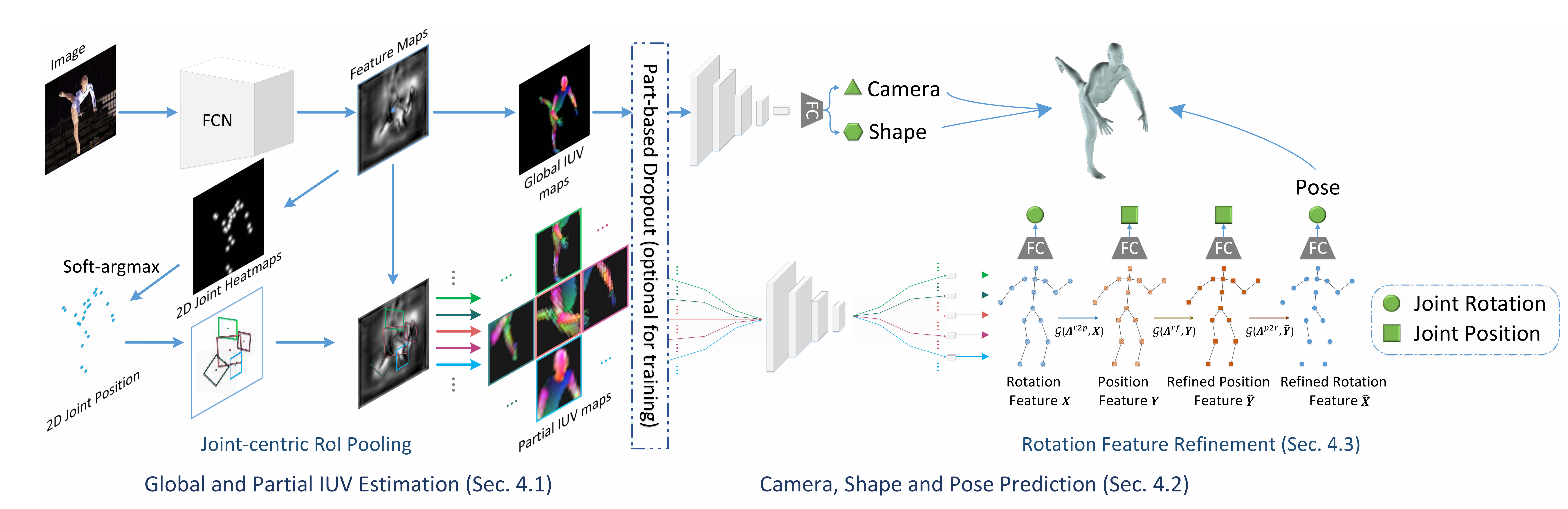}
		\caption{Overview of the proposed Decompose-and-aggregate Network (DaNet).}
		\label{fig:framework}
	\end{center}
\end{figure*}

\section{SMPL Model and IUV Maps}
\label{preliminary}
\textbf{SMPL Model.} The Skinned Multi-Person Linear model (SMPL)~\cite{loper2015smpl} is one of the widely used statistical human body models, which represents the body mesh with two sets of parameters, \ie, the shape and pose parameters.
The shape indicates the model's height, weight and limb proportions while the pose indicates how the model deforms with the rotated skeleton joints.
Such decomposition of shape and pose makes it convenient for algorithms to focus on one of these two factors independently.
In the SMPL model, the shape parameters $\bm{\beta} \in \mathbb{R}^{10}$ denote the coefficients of the PCA basis of the body shape.
The pose parameters $\bm{\theta} \in \mathbb{R}^{3K}$ denote the axis-angle representations of the relative rotation of $K$ skeleton joints with respect to their parents in the kinematic tree, where $K = 24$ in the SMPL model.
For simplicity, the root orientation is also included as the pose parameters of the root joint in our formulation.
Given the pose and shape parameters, the model deforms accordingly and generates a triangulated mesh with $N = 6890$ vertices $\mathcal{M}(\bm{\theta}, \bm{\beta}) \in \mathbb{R}^{3\times N}$.
The deformation process $\mathcal{M}(\bm{\theta}, \bm{\beta})$ is differentiable with respect to the pose $\bm{\theta}$ and shape $\bm{\beta}$,
which means that the SMPL model could be integrated within a neural network as a typical layer without any learnable weights.
After obtaining the final mesh, vertices could be further mapped to sparse 3D keypoints by a pretrained linear regressor.

\textbf{IUV Maps.}
Reconstructing the 3D object model from a monocular image is ambiguous, but there are determinate correspondences between foreground pixels on 2D images and vertices on 3D surfaces.
Such correspondences could be represented in the form of UV maps, where the foreground pixels contain the corresponding UV coordinate values. 
In this way, the pixels on the foreground could be projected back to vertices on the template mesh according to a predefined bijective mapping between the 3D surface space and the 2D UV space.
For the human body model, the correspondence could have finer granularity by introducing the $\textit{Index}$ of the body parts~\cite{alp2017densereg,alp2018densepose}, which results in the IUV maps $\bm{H} = (\bm{H}^{i} | \bm{H}^{u} | \bm{H}^{v}) \in \mathbb{R}^{(1+P) \times h_{iuv} \times w_{iuv} \times 3}$, where $P$ denotes the number of body parts, $h_{iuv}$ and $w_{iuv}$ denote the height and width of IUV maps.
The $\textit{Index}$ channels $\bm{H}^{i}$ indicates whether a pixel belongs to the background or a specific body part, while the $UV$ channels $\bm{H}^{u}$ and $\bm{H}^{v}$ contain the corresponding $U$, $V$ values of visible body parts respectively.
The IUV maps $\bm{H}$ encode $\textit{Index}$, $\textit{U}$, and $\textit{V}$ values individually for $P$ body parts in a one-hot manner along $(1+P)$ ways.
The $\textit{Index}$ values for body parts count from 1 and $\textit{Index}$ 0 is reserved for the background.
For each body part, the UV space is independent so that the representation could be more fine-grained.
The IUV annotation of the human body is firstly introduced in DenseReg~\cite{alp2017densereg} and DensePose~\cite{alp2018densepose}.
Figs.~\ref{fig:smpl_i}\subref{fig:smpl_u}\subref{fig:smpl_v} show the \textit{Index}, \textit{U}, and \textit{V} values on the SMPL model as defined in DensePose~\cite{alp2018densepose}.

\textbf{Preparation of IUV Maps for 3D Human Pose Datasets.}
Currently, there is no 3D human pose dataset providing IUV annotations.
In this work, for those datasets providing SMPL parameters with human images, we augment their annotations by rendering the corresponding ground-truth IUV maps based on the same IUV mapping protocol of DensePose~\cite{alp2018densepose}.
Specifically, we first construct a template texture map from IUV values of each vertex on the SMPL model, and then employ a renderer to generate IUV maps.
As illustrated in Fig.~\ref{fig:render_vis}, for each face in the triangulated mesh, the texture values used for rendering is a triplet vector denoting the corresponding $\textit{Index}$, $U$, and $V$ values.
Then, given SMPL models, the corresponding IUV maps can be generated by existing rendering algorithms such as~\cite{loper2014opendr,kato2018neural}.
Specifically, the renderer takes the template texture map and 3D model as inputs and output a rendered image with the size of $h_{iuv} \times w_{iuv} \times 3$.
Afterwards, the rendered image is reorganized as the shape of $(1+P) \times h_{iuv} \times w_{iuv} \times 3$ by converting values into one-hot representations.

\section{Methodology}
\label{methodology}

As illustrated in Fig.~\ref{fig:framework}, our DaNet decomposes the prediction task into one global stream for the camera and shape predictions and multiple local streams for joint pose predictions.
The overall pipeline involves two consecutive stages.
In the first stage, the IUV maps are estimated from global and local perspectives in consideration of the different sizes of the receptive fields required by the prediction of different parameters.
In the second stage, the global and local IUV maps are used for different feature extraction and prediction tasks.
The global features are extracted from global IUV maps and then directly used to predict camera and body shape.
The rotation features are extracted from partial IUV maps and further fed into the aggregated refinement module before the final prediction of joint poses.
During training, the part-based dropout is applied to the estimated IUV maps between the above two stages.

Overall, our objective function is a combination of three objectives:
\begin{equation}
\mathcal{L} = \mathcal{L}_{inter} + \mathcal{L}_{target} + \mathcal{L}_{refine},
\end{equation}
where $\mathcal{L}_{inter}$ is the objective for estimating the intermediate representations (Sec. \ref{uvi_est}),
$\mathcal{L}_{target}$ is the objective for predicting the camera and SMPL parameters (Sec. \ref{final_pred}),
$\mathcal{L}_{refine}$ is the objective involving in the aggregated refinement module (Sec. \ref{feat_refine}).
In the following subsections, we will present the technical details and rationale of our method.

\subsection{Global and Partial IUV Estimation}
\label{uvi_est}
\begin{figure}[t]
	\begin{subfigure}[b]{0.07\textwidth}
 		\centering
		\includegraphics[height=14mm]{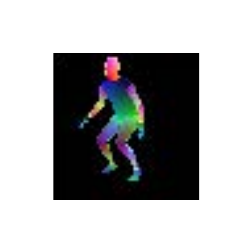}
		\caption{}
		\label{fig:uvi_vis_gl}
	\end{subfigure}
	\hspace{0.2mm}
	\begin{subfigure}[b]{0.18\textwidth}
		\centering
		\includegraphics[height=14mm]{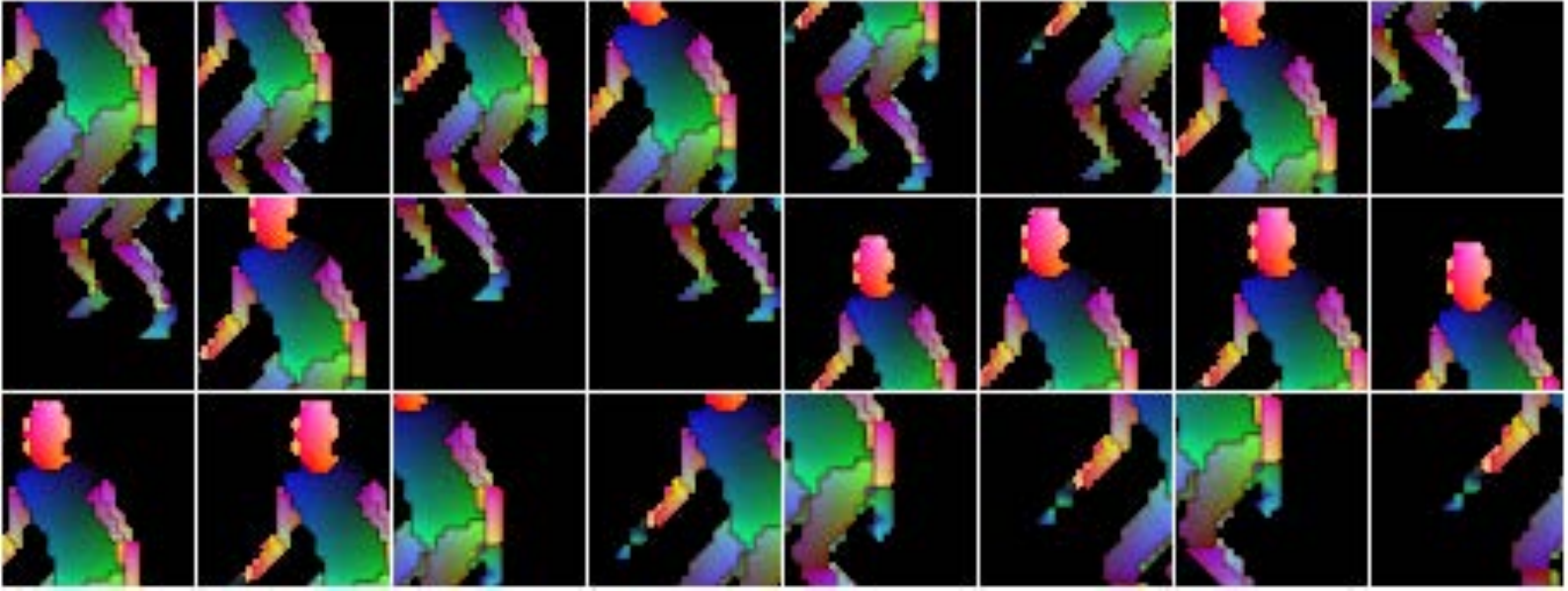}
		\caption{}
		\label{fig:uvi_vis_lc}
	\end{subfigure}
	\hspace{3mm}
	\begin{subfigure}[b]{0.18\textwidth}
		\centering
		\includegraphics[height=14mm]{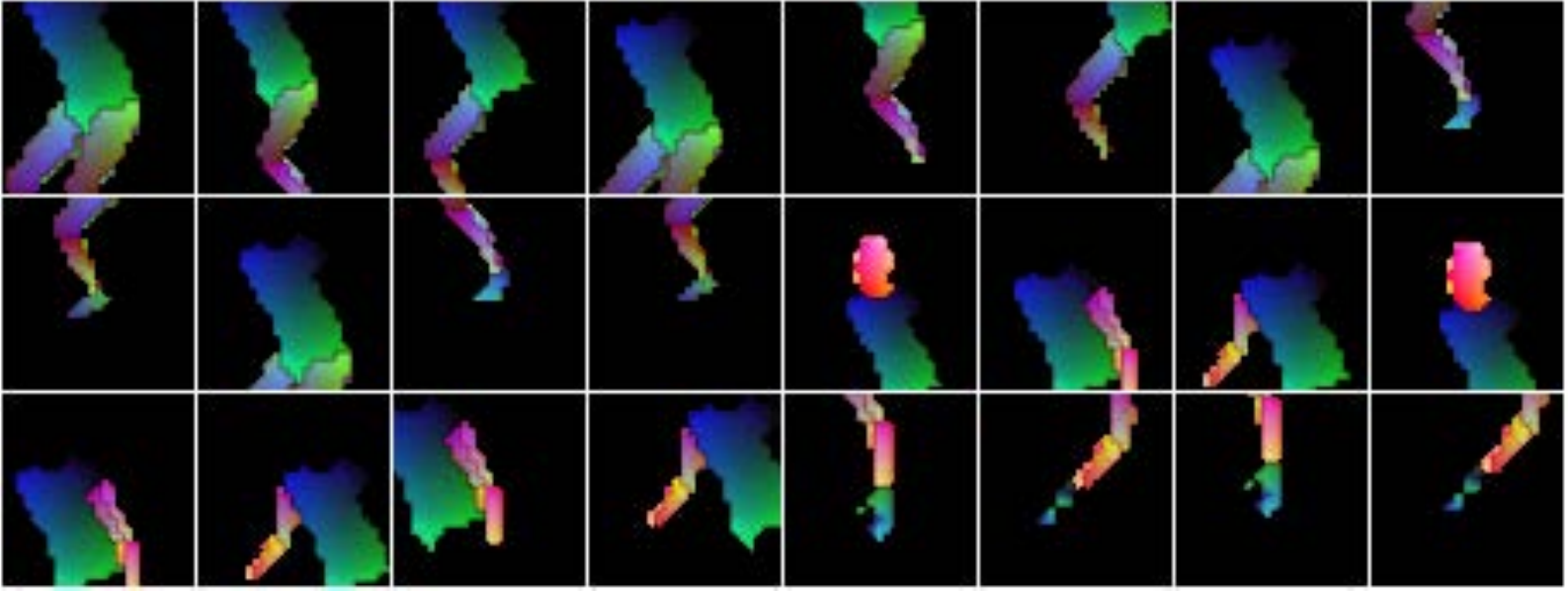}
		\caption{}
		\label{fig:uvi_vis_lc_sp}
	\end{subfigure}
	\vspace{-3mm}
    \caption{Visualization of (a) global, (b) partial, and (c) simplified partial IUV maps.}
\label{fig:uvi_vis}
\end{figure}

The first stage in our method aims to estimate corresponding IUV maps from input images for subsequent prediction tasks.
Specifically, a fully convolutional network is employed to produce $K+1$ sets of IUV maps, including one set of global IUV maps and $K$ sets of partial IUV maps for the corresponding $K$ body joints.
The global IUV maps are aligned with the original image through up-sampling, while the partial IUV maps are centered around the body joints.
Fig.~\ref{fig:uvi_vis} visualizes a sample of the global and partial IUV maps.
The feature maps outputted from the last layer of the FCN would be shared by the estimation tasks of both global and partial IUV maps.
The estimation of the global IUV maps is quite straightforward since they could be obtained by simply feeding these feature maps into a convolutional layer.
For the estimation of each set of partial IUV maps, a joint-centric RoI pooling would be first performed on these feature maps to extract appropriate sub-regions, which results in $K$ sets of partial feature maps.
Then, the $K$ sets of partial IUV maps would be estimated independently from these partial feature maps.
Now, we will give details about the RoI pooling process for partial IUV estimation.

\textbf{Joint-centric RoI Pooling.}
For pose parameters in the SMPL model, they represent the relative rotation of each body joint with respect to its parent in the kinematic tree.
Hence, the perception of joint poses should individually focus on corresponding body parts.
In other words, globally zooming, translating the human in the image should have no effect on the pose estimation of body joints.
Moreover, the ideal scale factors for the perception of joint poses should vary from one joint to another since the proportions of body parts are different.
To this end, we perform joint-centric RoI pooling on feature maps for partial IUV estimation.
Particularly, for each body joint, a sub-region of the feature maps is extracted and spatially transformed to a fixed resolution for subsequent partial IUV map estimation and joint pose prediction.
In our implementation, the RoI pooling is accomplished by a Spatial Transformer Network (STN)~\cite{jaderberg2015spatial}.
In comparison with the conventional STNs, the pooling process in our network is learned in an explicitly supervised manner.

\begin{figure}[t]
	\begin{subfigure}[b]{0.3\textwidth}
		\includegraphics[height=35mm]{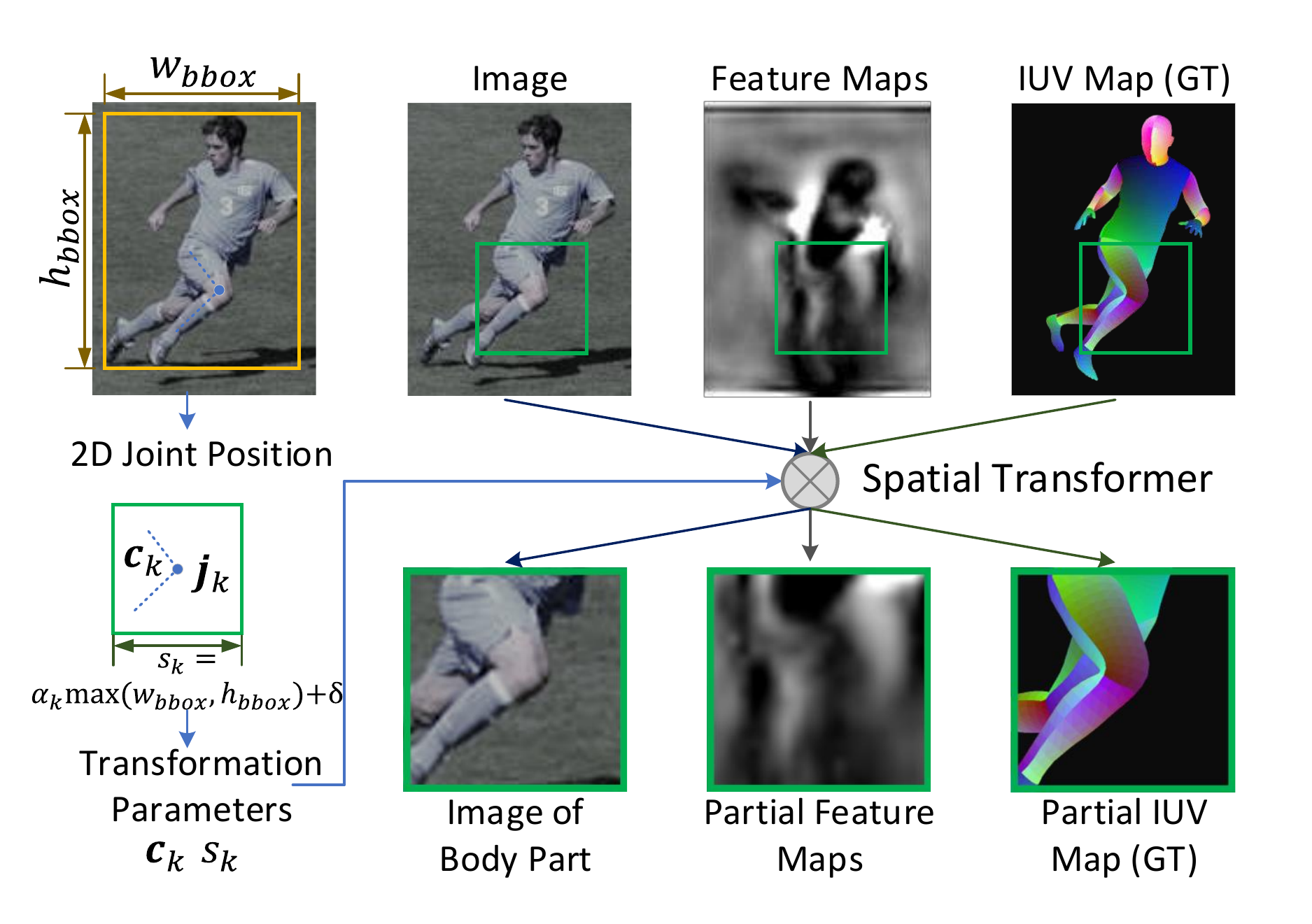}
		\caption{}
		\label{fig:roi_pooling}
	\end{subfigure}
	\hspace{-4mm}
	\begin{subfigure}[b]{0.15\textwidth}
		\includegraphics[height=35mm]{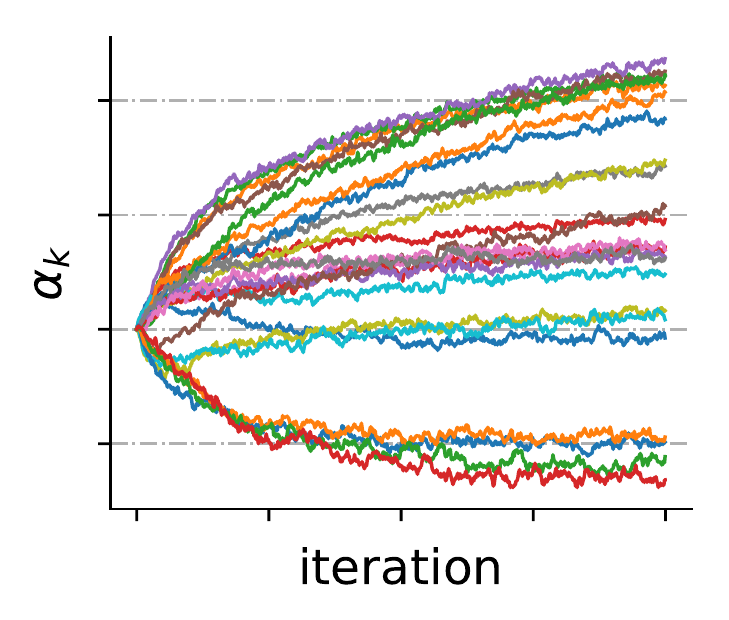}
		\caption{}
		\label{fig:learned_ratio}
	\end{subfigure}
	\vspace{-3mm}
    \caption{Joint-centric RoI pooling. (a) The RoI pooling is implemented as an STN. (b) The evolution of $\alpha_k$s of different body joints over learning iterations.}
\end{figure}

As illustrated in Fig.~\ref{fig:roi_pooling}, the joint-centric RoI pooling operations are guided by 2D joint positions so that each sub-region is centered around the target joint.
Specifically, 2D joint heatmaps are estimated along with the global IUV maps in a multi-task learning manner, and 2D joint positions are retrieved from heatmaps using the soft-argmax~\cite{sun2018integral} operation.
Without loss of generality, let $\bm{j}_{k}$ denote the position of the $k$-th body joint.
Then, the center and scale parameters used for spatial transformation are determined individually for each set of partial IUV maps.
Specifically, for the $k$-th set of partial IUV maps, the center $\bm{c}_k$ is the position of the $k$-th joint,
while the scale $s_k$ is proportional to the size of the foreground region, \ie,
\begin{equation}
\begin{aligned}
\bm{c}_k &= \bm{j}_{k}, \\
s_k &= \alpha_{k} \max(w_{bbox}, h_{bbox})+ \delta,
\end{aligned}
\label{stn_para}
\end{equation}
where $\alpha_k$ and $\delta$ are two constants, $w_{bbox}$ and $h_{bbox}$ denote the width and height of the foreground bounding box respectively.
In our implementation, the foreground is obtained from the part segmentation (\ie, $\textit{Index}$ channels of estimated IUV maps).
Compared with our previous work~\cite{zhang2019danet} calculating $s_k$ from 2D joints, the $s_k$s determined by foreground regions here are more robust to 2D joint localization.

Note that the above constants $\alpha_k$ and $\delta$ can be handcrafted or learned in the STN by taking ground-truth IUV maps as inputs.
For learned $\alpha_k$s, Fig.~\ref{fig:learned_ratio} shows how the values of different body joints evolve over learning iterations.
It can be observed that $\alpha_k$s are enlarged for some joints while shrunk for others, which provides more suitable RoI sizes for each body joint.

After obtaining the transformation parameters in Eq.~\ref{stn_para}, the feature maps extracted from the last layer of fully convolutional network are spatially transformed to a fixed resolution and used to estimate the partial IUV maps,
where the corresponding ground-truth ones are also extracted from the ground-truth global IUV maps using the same pooling process.

Considering that the pose of a body joint is only related to its adjacent body parts, we can further simplify partial IUV maps by discarding those irrelevant body parts.
For each set of partial IUV maps, we retain specific channels corresponding to those body parts surrounding the target joint.
The partial IUV maps before and after the simplification are depicted in Fig.~\ref{fig:uvi_vis_lc} and Fig.~\ref{fig:uvi_vis_lc_sp} respectively.

\textbf{Loss Functions.} 
A classification loss and several regression losses are involved in the training of this stage.
For both global and partial IUV maps, the loss is calculated in the same manner and denoted as $\mathcal{L}_{iuv}$.
Specifically, a classification loss is imposed on the $\textit{Index}$ channels of IUV maps, where a $(1+P)$-way cross-entropy loss is employed to classify a pixel belonging to either background or one among $P$ body parts.
For the $UV$ channels of IUV maps, an $L_1$ based regression loss is adopted, and is only taken into account for those foreground pixels.
In other words, the estimated $UV$ channels are firstly masked by the ground-truth $\textit{Index}$ channel before applying the regression loss.
For the 2D joint heatmaps and 2D joint positions estimated for RoI pooling, an $L_1$ based regression loss is adopted and denoted as $\mathcal{L}_{roi}$.
Overall, the objective in the IUV estimation stage involves two main losses:
\begin{equation}
\mathcal{L}_{inter} = \lambda_{iuv}\mathcal{L}_{iuv} + \lambda_{roi}\mathcal{L}_{roi},
\end{equation}
where $\lambda_{iuv}$ and $\lambda_{roi}$ are used to balance the two terms.

\subsection{Camera, Shape and Pose Prediction}
\label{final_pred}

After obtaining the global and partial IUV maps, the camera and shape parameters would be predicted in the global stream, while pose parameters would be predicted in the local streams.

The global stream consists of a ResNet~\cite{he2016deep} as the backbone network and a fully connected layer added at the end with 13 outputs, corresponding to the camera scale $s \in \mathbb{R}$, translation $\bm{t} \in \mathbb{R}^{2}$ and the shape parameters $\bm{\beta} \in \mathbb{R}^{10}$.
In the local streams, a tailored ResNet acts as the backbone network shared by all body joints and is followed by $K$ residual layers for rotation feature extraction individually.
For the $k$-th body joint, the extracted rotation features would be refined (see Sec.~\ref{feat_refine}) and then used to predict the rotation matrix $\bm{R}_k \in \mathbb{R}^{3\times 3}$ via a fully connected layer.
Here, we follow previous work~\cite{pavlakos2018learning,omran2018neural} to predict the rotation matrix representation of the pose parameters $\bm{\theta}$ rather than the axis-angle representation defined in the SMPL model.
Note that using other rotation representations such as the 6D continuous representation~\cite{zhou2019continuity} is also feasible.
An $L_1$ loss is imposed on the predicted camera, shape, and pose parameter, and we denote it as $\mathcal{L}_{smpl}$.

Following previous work~\cite{kanazawa2018end,pavlakos2018learning,omran2018neural}, we also add additional constraint and regression objective for better performance.
For the predicted rotation matrix, 
we impose an orthogonal constraint loss $\mathcal{L}_{orth}=\sum_{k=0}^{K-1}{\left\|\bm{R}_k \bm{R}_k^{\mathsf{T}} - \bm{I}\right\|_{2}}$ upon the predicted rotation matrices $\left\{\bm{R}_k \right\}_{k=0}^{K-1}$ to guarantee their orthogonality.
Moreover, given the predicted SMPL parameters, the performance could be further improved by adding supervision explicitly on the resulting model $\mathcal{M}(\bm{\theta}, \bm{\beta})$.
Specifically, three $L_1$ based loss functions are used to measure the difference between the ground-truth positions and the predicted ones.
The corresponding losses are denoted as $\mathcal{L}_{vert}$ for vertices on 3D mesh, $\mathcal{L}_{3Dkp}$ for sparse 3D human keypoints, and $\mathcal{L}_{reproj}$ for the reprojected 2D human keypoints, respectively.
For the sparse 3D human keypoints, the predicted positions are obtained via a pretrained linear regressor by mapping the mesh vertices to the 3D keypoints defined in human pose datasets.
Overall, the objective in this prediction stage is the weighted sum of multiple losses:
\begin{equation}
\begin{aligned}
    \mathcal{L}_{target} &= \lambda_{smpl}\mathcal{L}_{smpl} + \lambda_{orth}\mathcal{L}_{orth} \\
    &+ \lambda_{point}\left(\mathcal{L}_{vert} + \mathcal{L}_{3Dkp} + \mathcal{L}_{reproj}\right),
\end{aligned}
\end{equation}
where $\lambda_{smpl}$, $\lambda_{orth}$, and $\lambda_{point}$ are balance weights.

\textbf{Part-based Dropout.}
Our approach learn the shape and pose from the IUV intermediate representation, which contains dense correspondences of the body parts.
Following previous work on data augmentation~\cite{devries2017improved} and model regularization~\cite{srivastava2014dropout,ghiasi2018dropblock}, we introduce a Part-based Dropout (PartDrop) strategy to drop out semantic information from intermediate representations during training.
PartDrop has a dropping rate $\gamma$ as the probability of dropping values in the estimated IUV maps.
In contrast to other dropping out strategies such as Dropout~\cite{srivastava2014dropout} and DropBlock~\cite{ghiasi2018dropblock}, the proposed PartDrop strategy drops features in contiguous regions at the granularity level of body parts.
Specifically, for each training sample, the index subset $\mathcal{I}_{drop}$ of the body parts to be dropped is randomly selected from $\{1, 2, \dots, P\}$ with the probability of $\gamma$.
Then, for both global and partial IUV maps, the estimated IUV values of selected body parts are dropped out by setting corresponding body parts as zeros:
\begin{equation}
\bm{H}[p,:,:,:] = 0,~\text{for}~p \in \mathcal{I}_{drop},
\end{equation}
where $\bm{H}[p,:,:,:]$ denotes IUV maps with the part index of $p$.

\begin{figure}[t]
	\begin{center}
        \centering
        \begin{subfigure}[b]{0.09\textwidth}
    		\centering
    		\includegraphics[height=19mm]{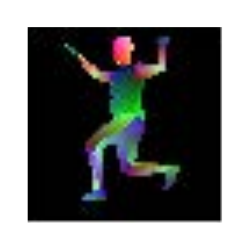}
    		\caption{}
    	\end{subfigure}
    	\foreach \img/\msk in {2/5,3/6,4/7} {
    	    \hspace{1mm}
        	\begin{subfigure}[b]{0.09\textwidth}
        		\centering
        		\includegraphics[height=19mm]{Img/drop_vis/drop_vis_Part\msk.pdf}

        		\includegraphics[height=19mm]{Img/drop_vis/drop_vis_Part\img.pdf}
        		\caption{}
        	\end{subfigure}
    	}
	\end{center}
	\vspace{-3mm}
    \caption{Comparison of different dropping out strategy. (a) Original IUV map. (b)(c)(d) PartDrop (ours), DropBlock~\cite{ghiasi2018dropblock} and Dropout~\cite{srivastava2014dropout} drop IUV values in part-wise, block-wise, and unit-wise manners, respectively. The corresponding binary masks are shown on the top row.}
\label{fig:drop_vis}
\end{figure}

PartDrop is motivated by the observation that the estimated IUV maps on real-world images typically have errors on irregular parts in challenging cases such as heavy occlusions.
Fig.~\ref{fig:drop_vis} visualizes how PartDrop, DropBlock~\cite{ghiasi2018dropblock}, and Dropout~\cite{srivastava2014dropout} drop values in part-wise, block-wise, and unit-wise manners.
It can be observed that, Dropout leaves obvious pepper-like artifacts after dropping, DropBlock introduces unwanted square patterns, while PartDrop brings much less visual artifacts in the resulting IUV maps.
In comparison with DropBlock and Dropout, the proposed PartDrop can remove semantic information from foreground areas in a more structured manner, which consequently enforces the neural network to learn features from complementary body parts and improves its generalization.

\subsection{Rotation Feature Refinement}
\label{feat_refine}

\begin{figure}[t]
	\begin{center}
    	\centering
    	\begin{subfigure}[b]{0.45\textwidth}
    		\centering
    		\includegraphics[width=1\textwidth]{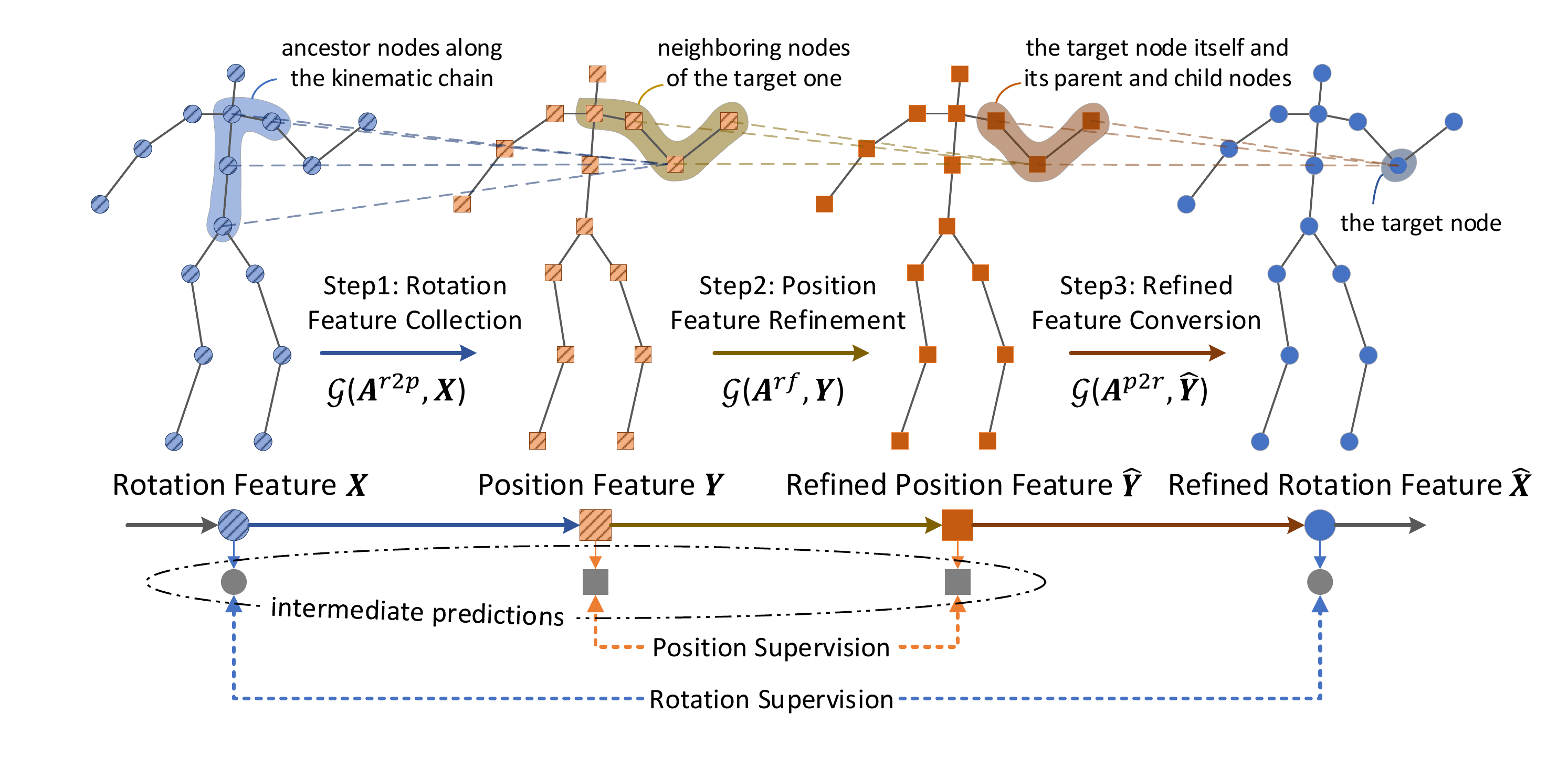}
    		\caption{Position-aided Rotation Feature Refinement}
    		\label{fig:rot2pos2rot}
    	\end{subfigure}
    	\begin{subfigure}[b]{0.115\textwidth}
    		\centering
    		\includegraphics[width=1.1\textwidth]{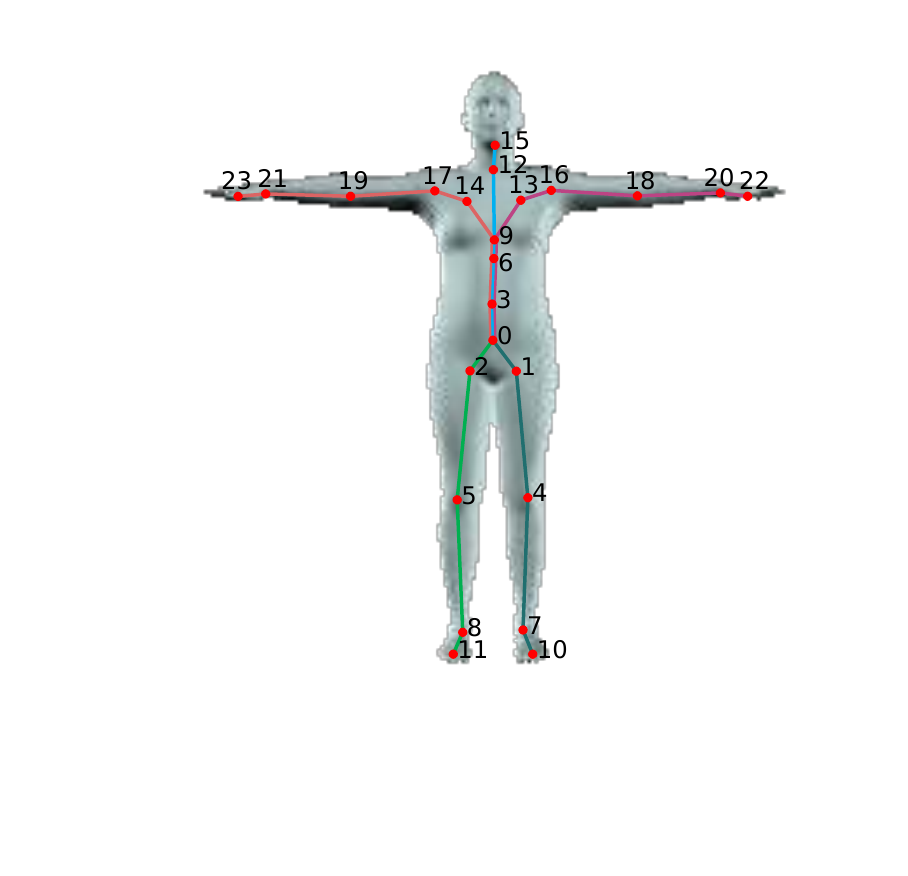}
    		\caption{Kine. Tree}
    		\label{fig:smpl_joint}
    	\end{subfigure}
    	\begin{subfigure}[b]{0.115\textwidth}
    		\centering
    		\includegraphics[width=1.1\textwidth]{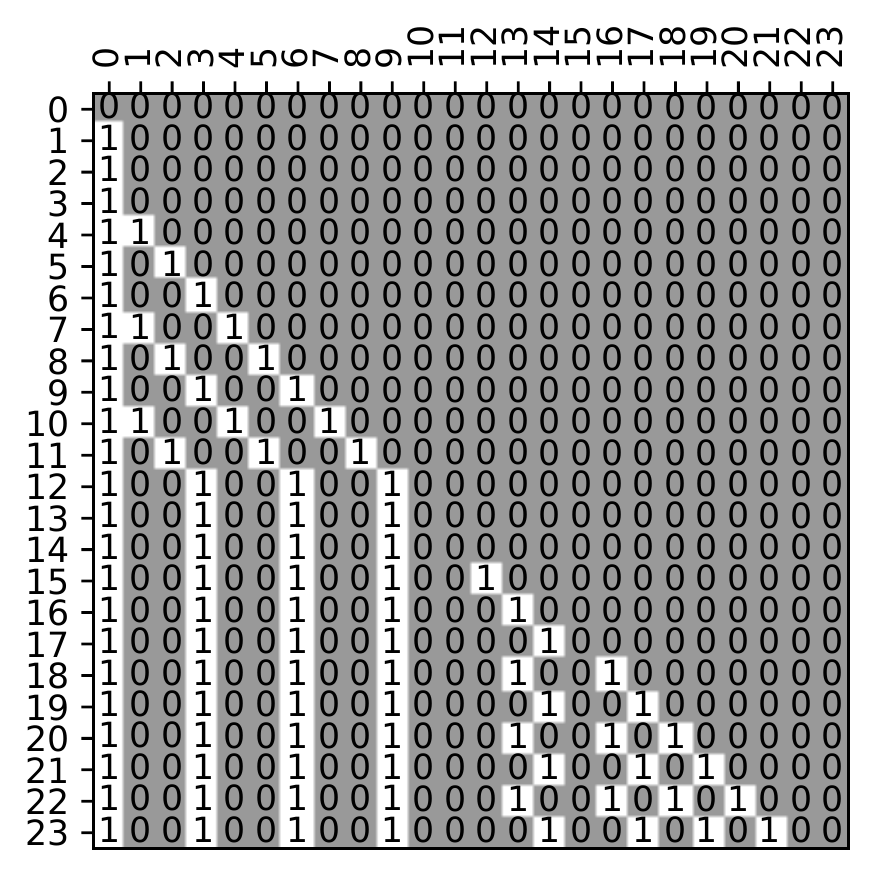}
    		\caption{$\bm{A}^{r2p}$}
    		\label{fig:graph_r2p}
    	\end{subfigure}
    	\begin{subfigure}[b]{0.115\textwidth}
    		\centering
    		\includegraphics[width=1.1\textwidth]{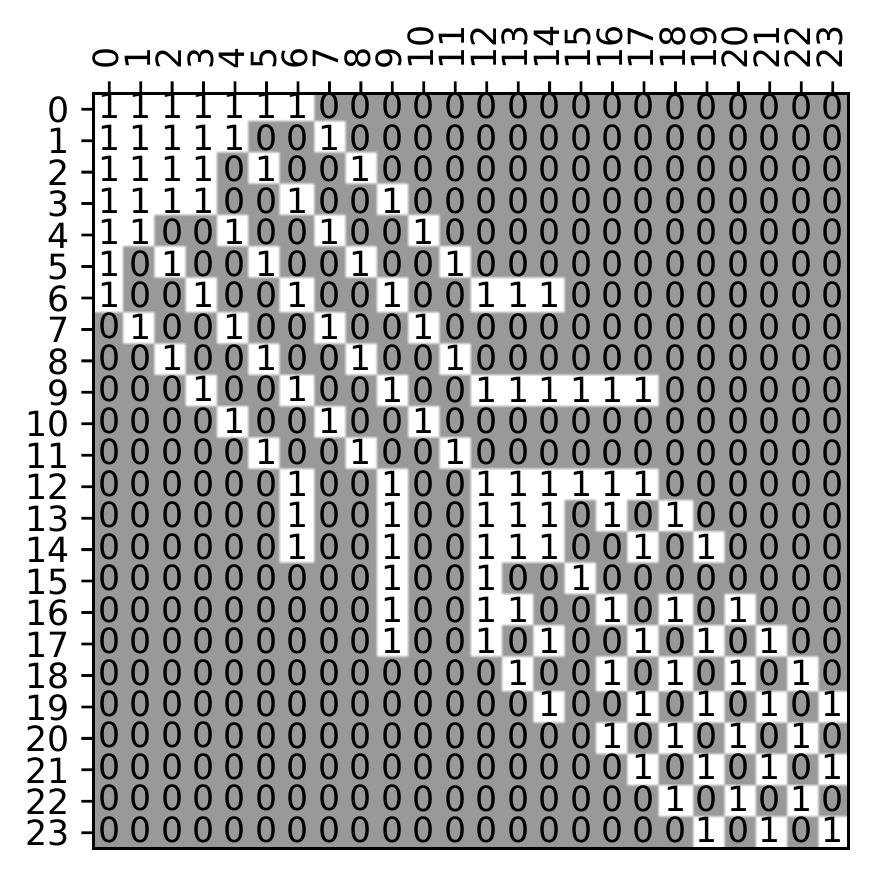}
    		\caption{$\bm{A}^{rf}$}
    		\label{fig:graph_rf}
    	\end{subfigure}
    	\begin{subfigure}[b]{0.115\textwidth}
    		\centering
    		\includegraphics[width=1.1\textwidth]{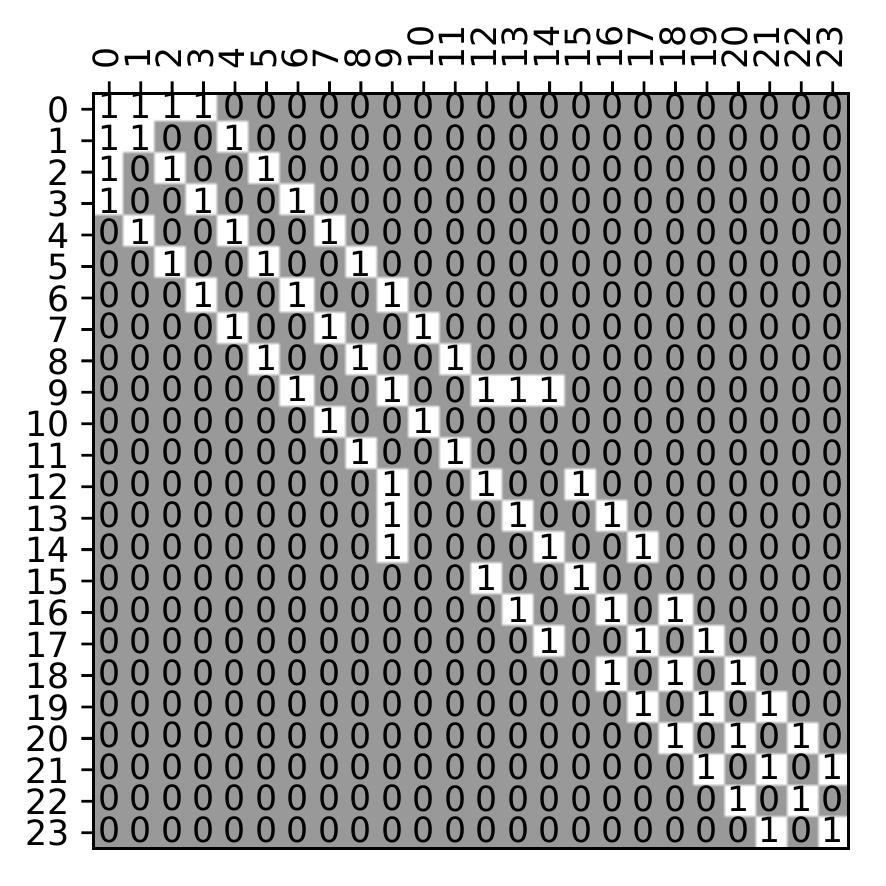}
    		\caption{$\bm{A}^{p2r}$}
    		\label{fig:graph_p2r}
    	\end{subfigure}
	\end{center}
	\vspace{-3mm}
    \caption{Illustration of the aggregated refinement module. (a) Three steps of the proposed refinement strategy. (b) The kinematic tree with $K=24$ joints in the SMPL model. The pelvis joint with 0 index is the root node of the tree. Joints belonging to the same kinematic chain are linked by the line with the same color. (c)(d)(e) Adjacency matrices of the graphs used in three steps for the feature collection, refinement, and conversion.}
\label{fig:graph}
\end{figure}

In our approach, the rotation features extracted in local streams are aggregated to exploit spatial relationships among body joints.
As illustrated in Fig.~\ref{fig:rot2pos2rot}, the position-aided rotation feature refinement involves three consecutive steps, namely rotation feature collection, position feature refinement, and refined feature conversion.
Specifically, the rotation features are first collected into the position feature space where the feature refinement is performed.
After that, the rotation feature refinement is accomplished by converting the refined position features back to the rotation feature space.
All these three steps are performed by customized graph convolution layers.
In particular, we consider the following graph-based convolution layer $\mathcal{G}(\cdot)$ that employs one popular formulation of the  Graph Convolution Networks as proposed in Kipf \etal~\cite{kipf2017semi}.
\begin{equation}
\bm{Z}_{out} = \mathcal{G}(\bm{A}, \bm{Z}_{in})= \sigma(\bm{\hat{A}}\bm{Z}_{in}\bm{W}),
\end{equation}
where $\bm{Z}_{in}$ and $\bm{Z}_{out}$ are input and output features respectively, $\sigma(\cdot)$ is the activation function, $\bm{W}$ is the parameters of convolution kernels, $\bm{\hat{A}}$ denotes the row-normalized matrix of the graph adjacency matrix $\bm{A}$, \ie, $\bm{\hat{A}}={\bm{D}}^{-\frac{1}{2}}\bm{A}{\bm{D}}^{-\frac{1}{2}}$ if $\bm{A}$ is a symmetric matrix, and otherwise $\bm{\hat{A}}=\bm{D}^{-1}\bm{A}$, where $\bm{D}$ is the diagonal node degree matrix of $\bm{A}$ with $\bm{D}_{ii}=\sum_{j}\bm{A}_{ij}$.
For simplicity, we also refer to the graph with adjacency matrix of $\bm{A}$ as graph $\bm{A}$.

\textbf{Step 1: Rotation Feature Collection.} 
Note that the rotation of each body joint could be viewed as sequential data along the kinematic chain.
This is inspired by the fact that the human could act in a recurrent manner according to the kinematic tree shown in Fig.\ref{fig:smpl_joint}.
The position of a specific body joint can be calculated from the collection of the relative rotations and bone lengths of those joints belonging to the same kinematic chain.
At the feature level, we propose to learn the mapping from rotation feature space to position feature space.
To that end, one graph convolution layer is customized to gather information from body joints along the kinematic chain and learn such mapping.
Formally, let $\bm{X}\in \mathbb{R}^{K\times C}$ denote the rotation features extracted from $K$ sets of partial IUV maps with $C$ being the feature dimension.
The position features $\bm{Y}\in \mathbb{R}^{K\times C}$ of $K$ joints is obtained by feeding $\bm{X}$ to the graph convolution, \ie,
\begin{equation}
\bm{Y} = \mathcal{G}(\bm{A}^{r2p}, \bm{X}),
\label{eq:step1}
\end{equation}
where the graph with adjacency matrix $\bm{A}^{r2p}$ is customized as a collection graph for mapping rotation features into the position feature space, in which $\bm{A}^{r2p}_{ij}=1$ if the $j$-th joint is one of the ancestors of the $i$-th joint along the kinematic chain, and otherwise $\bm{A}^{r2p}_{ij}=0$.
The adjacency matrix $\bm{A}^{r2p}$ of the collection graph is depicted in Fig.~\ref{fig:graph_r2p}.

\textbf{Step 2: Position Feature Refinement.} 
Since there are strong spatial correlations among neighboring body joints, utilizing such structured constraints could effectively improve the features learned at each joint.
Towards this goal, a graph-based convolution network is employed to exploit spatial relationships between joints.
Specifically, the position features $\bm{Y}$ are fed into $L$ graph convolution layers with the following layer-wise formulation:
\begin{equation}
\bm{Y}^{(l)} = \mathcal{G}(\bm{A}^{rf}, \bm{Y}^{(l-1)}),
\label{eq:step2}
\end{equation}
where $\bm{Y}^{l}$ denotes the position features obtained from the $l$-th layer with $\bm{Y}^{0}=\bm{Y}$, and the graph with adjacency matrix $\bm{A}^{rf}=\bm{I}+\bm{\tilde{A}}^{rf}$ serves as a refinement graph for feature refinement, in which $\bm{\tilde{A}}^{rf}_{ij}=1$ if the $i$-th and $j$-th joints are neighboring, and otherwise $\bm{\tilde{A}}^{rf}_{ij}=0$.
After graph convolutions, the refined position features $\bm{\hat{Y}}$ are obtained by adding $\bm{Y}^{L}$ with the original position features $\bm{Y}$ in a residual manner, \ie, $\bm{\hat{Y}}=\bm{Y} + \bm{Y}^{L}$.
Fig.~\ref{fig:graph_rf} shows an example of the adjacency matrix $\bm{A}^{rf}$ which considers both one-hop and two-hop neighbors.
Note that $\bm{A}^{rf}$ could have various forms according to the neighbor definition of body joints.

Inspired by previous work~\cite{yan2018spatial,zhao2019semantic}, we also add a learnable edge weighting mask on the graph convolution of this step since messages from different joints should have different contributions to the feature refinement of the target joint.
In this way, we have the adjacency matrix $\bm{A}^{rf}$ improved as
\begin{equation}
\bm{A}^{rf} = \bm{I} + \bm{M} \circ \bm{\tilde{A}}^{rf},
\end{equation}
where $\circ$ denotes the element-wise product, $\bm{M} \in [0, 1]^{K\times K}$ is the learnable edge weighting matrix serving as an attention mask of the graph to balance the contributions of neighboring features to the target feature.

\textbf{Step 3: Refined Feature Conversion.} 
The last step of refinement is to convert the refined features back to the original rotation feature space.
Since the rotation and position of body joints are two mutual representation of 3D human pose, after the refinement of position features, the rotation features can be refined accordingly\footnote{Strictly speaking, the joint rotations can not be fully retrieved from the joint positions due to the fewer DoFs specified in position-based poses. This issue is mild at the feature level since features could be more redundant.}.
Specifically, for the $k$-th body joint, its rotation features can be refined by aggregating messages from the refined position features of three consecutive body joints, \ie, the joint itself and its parent and child joints.
Similar to the first step, the mapping from position features to rotation features is also learned via a graph-based convolution layer, where the difference lies in the adjacency matrix of the graph.
Formally, the refined position features $\bm{\hat{Y}}$ are fed into the graph to obtain features in the rotation space,
resulting in the refined rotation features $\bm{\hat{X}}$ for the final prediction of joint pose parameters, \ie,
\begin{equation}
\bm{\hat{X}} = \mathcal{G}(\bm{A}^{p2r}, \bm{\hat{Y}}),
\label{eq:step3}
\end{equation}
where the graph with adjacency matrix $\bm{A}^{p2r} = \bm{I} + \bm{\tilde{A}}^{p2r}$ is customized as a conversion graph for mapping position features to rotation features, in which $\bm{\tilde{A}}^{p2r}_{ij}=1$ if the $j$-th joint is the parent or child joint of the $i$-th joint, and otherwise $\bm{\tilde{A}}_{ij}=0$.
The adjacency matrix $\bm{A}^{p2r}$ of the conversion graph is depicted in Fig.~\ref{fig:graph_p2r}.

\textbf{Supervision in Refinement.} 
The rotation and position feature spaces are built under corresponding supervisions during training.
As illustrated in Fig.~\ref{fig:rot2pos2rot}, the rotation features $\bm{X}$ and $\bm{\hat{X}}$ are used to predict joint rotations, while the position features $\bm{Y}$ and $\bm{\hat{Y}}$ are used to predict joint positions.
$L_1$ based rotation and position supervisions are imposed on these predictions correspondingly, which compose the objective $\mathcal{L}_{refine}$ involved in the refinement procedure.
Note that these intermediate predictions are unnecessary during testing.

%% file: tex/Experiments.tex
\section{Experiments}
\label{experiments}

\subsection{Implementation Details}
The FCN for IUV estimation in our framework adopts the architecture of HRNet-W48~\cite{wang2020deep}, which is one of the most recent state-of-the-art networks for dense estimation tasks.
The FCN receives the $224\times224$ input and produces $56\times56$ feature maps for estimating global and local IUV maps with the same resolution.
The IUV estimation network is initialized with the model pretrained on the COCO keypoint detection dataset~\cite{lin2014microsoft}, which is helpful for robust joint-centric RoI pooling and partial IUV estimation.
Two ImageNet-pretrained ResNet-18~\cite{he2016deep} are employed as the backbone networks for global and rotation feature extraction respectively.
During training, data augmentation techniques, including color jittering and flipping, are applied randomly to input images.
Random rotation is used when in-the-wild datasets are involved for training.
The IUV estimation task is first trained for 5k iterations before involving the parameter prediction task.
The $\alpha_k$s in Eq. (2) are first learned using ground-truth IUV maps as inputs and then frozen as constants for other experiments, while $\delta$ is empirically set to 0.1.
The hyper-parameters $\lambda$s are decided based on the scales of values in objectives.
The dropping rate $\gamma$ for PartDrop is adopted as 0.3 in our experiments.
For more robust pose recovery from the estimated partial IUV, we perform random jittering on the estimated 2D joint position and the scale of partial IUV maps during training.
Following previous work~\cite{kanazawa2018end,kolotouros2019learning}, the predicted poses are initialized from the mean pose parameters.
For faster runtime, the local streams are implemented to run in a parallel manner. Specifically, the partial IUV maps of all body joints are concatenated batch-wise and then fed into the backbone feature extractor. Moreover, individual rotation feature extraction is implemented based on group convolution.
By default, we adopt the Adam~\cite{kingma2014adam} optimizer with an initial learning rate of $1\times 10^{-4}$ to train our model, and reduce the learning rate to $1\times 10^{-5}$ after 30k iterations.
The learning process converges after around 60k iterations and takes about 25 hours on a single TITAN Xp GPU.
During testing, due to the fundamental depth-scale ambiguity, we follow previous work~\cite{kanazawa2018end,omran2018neural} to center the person within the image and perform scaling such that the inputs have the same setting as training.
Our experiments are implemented in PyTorch~\cite{paszke2019pytorch}.
More implementation details could be found in the publicly available code.

\subsection{Datasets and Evaluation Metrics}

\textbf{Human3.6M.} Human3.6M~\cite{ionescu2014human3} is a large-scale dataset which consists of 3.6 millions of video frames captured in the controlled environment, and currently the most commonly used benchmark dataset for 3D human pose estimation.
Kanazawa \etal~\cite{kanazawa2018end} generated the ground truth SMPL parameters by applying MoSH~\cite{loper2014mosh} to the sparse 3D MoCap marker data.
Following the common protocols~\cite{pavlakos2017coarse,pavlakos2018learning,kanazawa2018end}, we use five subjects (S1, S5, S6, S7, S8) for training and two subjects (S9, S11) for evaluation.
We also down-sample the original videos from 50fps to 10fps to remove redundant frames, resulting in 312,188 frames for training and 26,859 frames for testing.

\textbf{UP-3D.} 
UP-3D~\cite{lassner2017unite} is a collection dataset of existing 2D human pose datasets (\ie, LSP~\cite{johnson2010clustered}, LSP-extended~\cite{johnson2011learning}, MPII HumanPose~\cite{andriluka20142d}, and FashionPose~\cite{dantone2014body}), 
containing 5,703 images for training, 1,423 images for validation, and 1,389 images for testing.
The SMPL parameter annotations of these real-world images are augmented in a semi-automatic way by using an extended version of SMPLify~\cite{lassner2017unite}.

\textbf{COCO.}
The COCO dataset~\cite{lin2014microsoft} contains a large scale of images and person instances labeled with 17 keypoints.
Based on the COCO dataset, DensePose-COCO~\cite{alp2018densepose} further provides the dense correspondences from 2D images to the 3D surface of the human body model for 50K humans.
Different from our rendered IUV maps, the correspondence annotations in DensePose-COCO only consist of approximately 100-150 points per person, which are a sparse subset of the foreground pixels of human images.
In our experiments, we discard those persons without 2D keypoint annotations, resulting in 39,210 samples for training.
Since there are no ground-truth shape and pose parameters for COCO, we evaluate our method quantitatively on the keypoint localization task using its validation set, which includes 50,197 samples.

\textbf{3DPW.} 
The 3DPW dataset~\cite{von2018recovering} is a recent in-the-wild dataset providing accurate shape and pose ground truth annotations.
This dataset captured IMU-equipped actors in challenging outdoor scenes with various activities.
Following previous work~\cite{kanazawa2019learning,kolotouros2019learning}, we do not use its data for training but perform evaluations on its defined test set only.
There are 35,515 samples extracted from videos for testing.

\textbf{Fitted SMPL labels from SPIN.}
Kolotouros \etal~\cite{kolotouros2019learning} proposed SPIN to incorporate a fitting procedure within the training of a SMPL regressor.
The regressor provided better initialization for the fitting of human models to 2D keypoints, and the resulting SMPL parameters could be more accurate than those fitted in a static manner.
For evaluation on 3DPW~\cite{von2018recovering}, our model would be supervised with the final fitted SMPL labels from SPIN~\cite{kolotouros2019learning} for in-the-wild datasets including LSP~\cite{johnson2010clustered}, LSP-Extended~\cite{johnson2011learning}, MPII~\cite{andriluka20142d}, COCO~\cite{lin2014microsoft}, and MPI-INF-3DHP~\cite{mehta2017monocular}.

\textbf{Evaluation Metrics.}
Following previous work~\cite{pavlakos2018learning,varol2018bodynet,rong2019delving}, for evaluating the reconstruction performance, we adopt the mean Per-vertex Error (PVE) as the primary metric, which is defined as the average point-to-point Euclidean distance between the predicted model vertices and the ground truth model vertices.
Besides the PVE metric, we further adopt PVE-S and PVE-P as secondary metrics for separately evaluate the shape and pose prediction results.
The PVE-S computes the per-vertex error with the pose parameters of ground truth and predicted models set as zeros (\ie, models under the rest pose~\cite{loper2015smpl}), while the \mbox{PVE-P} computes the analogous per-vertex error with the shape parameters set as zeros.
For the Human3.6M dataset, the widely used Mean Per Joint Position Error (MPJPE) and the MPJPE after rigid alignment of the prediction with ground truth using Procrustes Analysis (MPJPE-PA) are also adopted to quantitatively evaluate the 3D human pose estimation performance.
The above three metrics will be reported in millimeters (mm) by default.

For the keypoint localization task on COCO, the commonly-used metric is the Average Precision (AP) defined by its organizers\footnote{\url{https://cocodataset.org/\#keypoints-eval}}.
The keypoint localization AP is calculated based on the Object Keypoint Similarity (OKS), which plays the same role as the IoU in object detection.
We report results using the mean AP, and the variants of AP including $\textrm{AP}_{50}$ (AP at OKS = 0.50), $\textrm{AP}_{75}$ (AP at OKS = 0.75), $\textrm{AP}_{M}$ for medium objects, and $\textrm{AP}_{L}$ for large objects.

\subsection{Comparison with State-of-the-art Methods}

\begin{table}[th]
  \centering
  \caption{Quantitative comparison with state-of-the-art methods on the Human3.6M dataset.}
    \begin{tabular}{lccc}
    \toprule
    Method & \multicolumn{1}{l}{PVE} & \multicolumn{1}{l}{MPJPE} & \multicolumn{1}{l}{MPJPE-PA} \\
    \midrule
	Zhou \etal~\cite{zhou2016deep} & - & 107.3 & - \\
	Tung \etal~\cite{tung2017self} & -  & -  & 98.4 \\
	SMPLify~\cite{bogo2016keep} & 202.0  & -  & 82.3 \\
	SMPLify++~\cite{lassner2017unite} & -  & -  & 80.7 \\
	Pavlakos \etal~\cite{pavlakos2018learning} & 155.5  & -  & 75.9 \\
	HMR~\cite{kanazawa2018end} & -  & 88.0    & 56.8 \\
	NBF~\cite{omran2018neural}   & -  & -  & 59.9 \\
	Xiang \etal~\cite{xiang2019monocular} & - & 65.6 & - \\
	Arnab \etal~\cite{arnab2019exploiting} & - & 77.8 & 54.3 \\
	CMR~\cite{kolotouros2019convolutional} & - & - & 50.1 \\
	HoloPose~\cite{guler2019holopose} & - & 60.3 & 46.5 \\
	TexturePose~\cite{pavlakos2019texturepose} & - & - & 49.7 \\
	DenseRaC~\cite{xu2019denserac} & - & 76.8  & 48.0 \\
    SPIN~\cite{kolotouros2019learning} & - & - & \textbf{41.1} \\
    \midrule
	DaNet-LSTM~\cite{zhang2019danet}  & 75.1 & 61.5 &  48.6 \\
	Ours  & \textbf{66.5} & \textbf{54.6} & 42.9 \\
    \bottomrule
    \end{tabular}%
    \vspace{-3mm}
  \label{tab:h36m}%
\end{table}%

\addtolength{\tabcolsep}{-3.5pt}
\begin{table*}[t]
  \centering
  \caption{Quantitative comparison of MPJPE-PA across different actions on the Human3.6M dataset.}
    \begin{tabular}{lcccccccccccccccc}
    \toprule
    Method & Direct. & Discuss & Eating & Greet & Phone & Photo & Pose  & Purch. & Sitting & SitingD. & Smoke & Wait  & WalkD. & Walk  & WalkT. & Avg. \\
    \midrule
    Pavlakos \etal~\cite{pavlakos2017coarse} & 47.5 & 50.5  & 48.3  & 49.3  & 50.7  & 55.2  & 46.1  & 48.0  & 61.1  & 78.1  & 51.1  & 48.3  & 52.9  & 41.5  & 46.4  & 51.9 \\
    Martinez \etal~\cite{martinez2017simple} & 39.5 & 43.2  & 46.4  & 47.0  & 51.0  & 56.0  & 41.4  & 40.6  & 56.5  & 69.4  & 49.2  & 45.0  & 49.5  & 38.0  & 43.1  & 47.7 \\
    \midrule
    SMPLify~\cite{bogo2016keep} & 62.0 & 60.2  & 67.8  & 76.5  & 92.1  & 77.0  & 73.0  & 75.3  & 100.3 & 137.3 & 83.4  & 77.3  & 79.7  & 86.8  & 81.7  & 82.3 \\
    HMR~\cite{kanazawa2018end} & 53.2  & 56.8  & 50.4  & 62.4  & 54.0  & 72.9  & 49.4  & 51.4  & 57.8  & 73.7  & 54.4  & 50.0  & 62.6  & 47.1  & 55.0  & 57.2 \\
    CMR~\cite{kolotouros2019convolutional} & 41.8  & 44.8  & 42.6  & 46.6  & 45.9  & 57.2  & 40.8  & 40.6  & 52.2  & 66.0  & 46.6  & 42.8  & 51.7  & 36.9  & 44.6  & 48.2 \\
    SPIN~\cite{kolotouros2019learning}  & 37.6  & 42.4  & \textbf{38.8} & 42.6  & \textbf{40.4} & \textbf{45.9} & 36.1  & 36.7  & 48.7  & 58.6  & 41.2  & 37.9  & 46.6  & \textbf{33.8} & \textbf{38.4} & 41.1 \\
    \midrule
    DaNet-LSTM~\cite{zhang2019danet} & 43.3  & 48.8  & 50.6  & 48.3  & 47.3  & 55.5  & 41.6  & 42.7  & 53.8  & 61.5  & 47.4  & 43.2  & 53.3  & 40.8  & 47.9  & 48.6 \\
    Ours  & 37.9  & 44.3  & 41.2  & 43.3  & 42.1  & 48.7  & 36.2  & 38.9  & 47.4  & 53.7  & 41.1  & 39.9  & 46.0  & 34.6  & 41.3  & 42.9 \\
    Ours-6D & \textbf{35.7} & \textbf{40.4} & 39.0  & \textbf{40.3} & 40.5  & 47.4  & \textbf{35.1} & \textbf{34.9} & \textbf{45.2} & \textbf{51.7} & \textbf{39.6} & \textbf{37.8} & \textbf{43.4} & 34.4  & 39.8  & \textbf{40.5} \\
    \bottomrule
    \end{tabular}%
  \label{tab:h36m_action}%
\end{table*}%
\addtolength{\tabcolsep}{3.5pt}

\begin{figure*}[t]
	\centering
	\begin{tikzpicture}[remember picture,overlay]
	\node[font=\fontsize{8pt}{8pt}\selectfont, rotate=90] at (0,4) {Image};
	\node[font=\fontsize{8pt}{8pt}\selectfont, rotate=90] at (0,2.2) {HMR~\cite{kanazawa2018end}};
	\node[font=\fontsize{8pt}{8pt}\selectfont, rotate=90] at (0,0.2) {NBF~\cite{omran2018neural}};
	\node[font=\fontsize{8pt}{8pt}\selectfont, rotate=90] at (0,-1.6) {CMR~\cite{kolotouros2019convolutional}};
	\node[font=\fontsize{8pt}{8pt}\selectfont, rotate=90] at (0,-3.6) {Ours};
	\end{tikzpicture}
	\foreach \idx in {0,1,2,3,4,5,6,7,8} {
		\begin{subfigure}[h]{0.095\textwidth}
			\centering
			\foreach \sub in {1,2,3,4,5} {
    			\pgfmathsetmacro\imidx{int(\idx * 5+\sub)}
    			\includegraphics[width=1.1\textwidth]{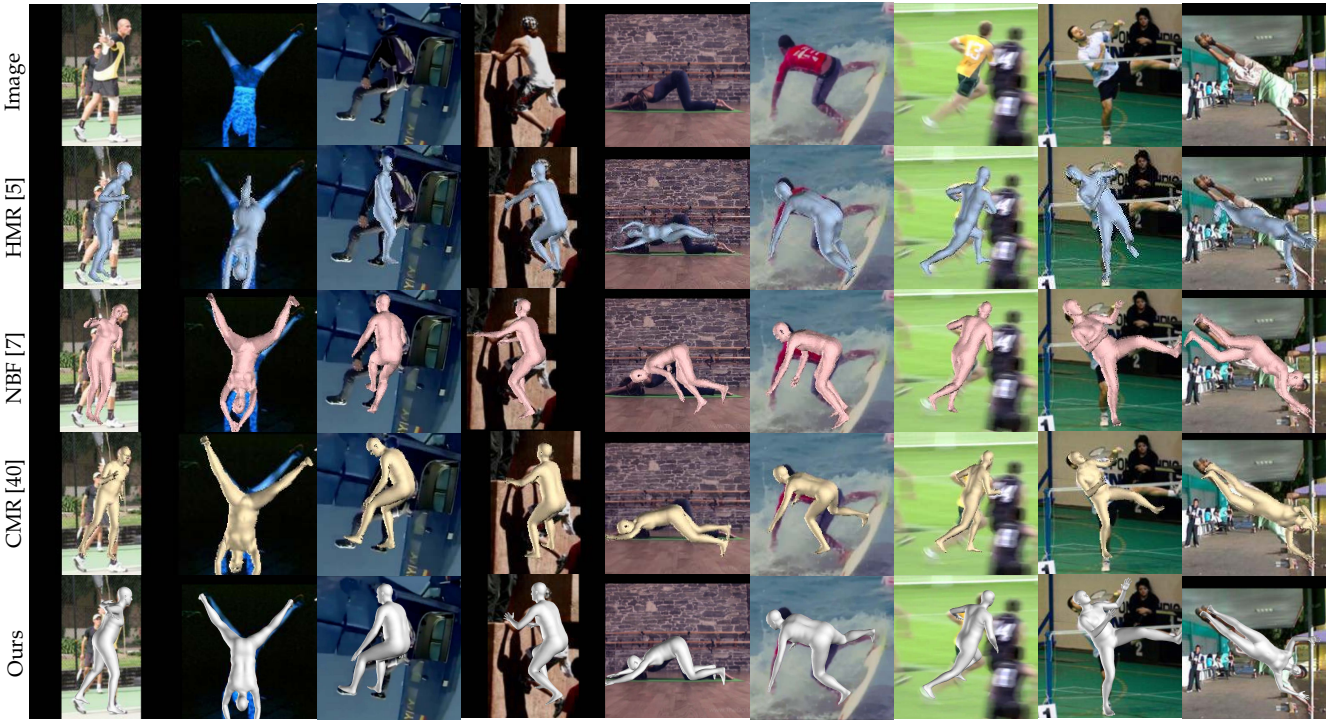}
    		}
		\end{subfigure}
	}
	\vspace{-3mm}
	\caption{Qualitative comparison of reconstruction results on the UP-3D dataset. }
	\label{fig:up3dDemo}
\end{figure*}

\subsubsection{Comparison on the Indoor Dataset.}
\textbf{Evaluation on Human3.6M.}
We evaluate the 3D human mesh recovery as well as pose estimation performance for quantitative comparison on Human3.6M, where our model is trained on its training set.
Table~\ref{tab:h36m} reports the comparison results with previous methods that output more than sparse 3D keypoint positions.
For regression-based methods in Table~\ref{tab:h36m}, different architectures have been designed to predict the shape and pose parameters.
Among them, HMR~\cite{kanazawa2018end} adopts a single CNN and an iterative regression module to produce all parameters.
Pavlakos \etal~\cite{pavlakos2018learning} decompose the shape and pose prediction tasks, while their pose parameters are predicted from 2D joints positions.
NBF~\cite{omran2018neural} adopts segmentation as the intermediate representation and learns all parameters from it.
CMR~\cite{kolotouros2019convolutional} directly regresses 3D meshes with a graph-based convolutional network.
All these architectures estimate pose parameters through a single stream with an exception that HoloPose~\cite{guler2019holopose} regresses poses using a part-based model.
As can be seen from Table~\ref{tab:h36m}, our network significantly outperforms the above-mentioned architectures.
It's worth noting that the methods reported in Table~\ref{tab:h36m} are not strictly comparable since they may use different datasets for training.
Among existing state-of-the-art approaches, we have a very competitive result which is only inferior to SPIN in Table~\ref{tab:h36m}.
SPIN has the same architecture as HMR except that it uses the 6D continuous representation~\cite{zhou2019continuity} for 3D rotations.
SPIN aims to incorporate regression- and optimization-based methods, while our work focuses on the design of a stronger regressor.
Hence, our method is complementary to SPIN since we can combine them together by simply plugging our network into SPIN.

For more comprehensive comparison, Table~\ref{tab:h36m_action} reports pose estimation performance across different actions on Human3.6M.
Compared with SPIN and other methods, our method can be more robust to challenging actions such as Sitting and Sitting Down.
We believe these benefits come from our decomposition design which enables our network to capture more detailed information for joint poses and produce more accurate reconstruction results.
We can also see from the last row of Table~\ref{tab:h36m_action} that, by simply replacing rotation matrices with the 6D representations~\cite{zhou2019continuity} for pose parameters as SPIN do, our method can achieve results on par with or even better than SPIN.

\begin{table}[t]
  \centering
  \caption{Quantitative comparison of PVE with state-of-the-art methods on the UP-3D dataset.}
    \begin{tabular}{l|ccc|c}
    \toprule
    Method & LSP   & MPII  & FashionPose & Full \\
    \midrule
    SMPLify++~\cite{lassner2017unite} & 174.4 & 184.3 & 108.0   & 169.8 \\
    HMR~\cite{kanazawa2018end} & - & - & - & 149.2 \\
    NBF~\cite{omran2018neural} & - & - & - & 134.6 \\
    Pavlakos \etal~\cite{pavlakos2018learning} & 127.8 & 110.0   & 106.5 & 117.7 \\
    BodyNet~\cite{varol2018bodynet} & 102.5 & -     & -     & - \\
    Rong \etal~\cite{rong2019delving} & - & - & - & 122.2 \\
    \midrule
    DaNet-LSTM~\cite{zhang2019danet}  & 90.4 & 83.0 & 61.8 & 83.7 \\
    Ours  & \textbf{88.5} & \textbf{82.1} & \textbf{60.8} & \textbf{82.3} \\
    \bottomrule
    \end{tabular}%
  \label{tab:up3d}%
\end{table}%

\begin{figure*}[t]
	\centering
	\begin{tikzpicture}[remember picture,overlay]
	\node[font=\fontsize{8pt}{8pt}\selectfont, rotate=90] at (0,4.2) {Image};
	\node[font=\fontsize{8pt}{8pt}\selectfont, rotate=90] at (0,2.3) {HMR~\cite{kanazawa2018end}};
	\node[font=\fontsize{8pt}{8pt}\selectfont, rotate=90] at (0,0.4) {Rong \etal~\cite{rong2019delving}};
	\node[font=\fontsize{8pt}{8pt}\selectfont, rotate=90] at (0,-1.6) {SPIN~\cite{kolotouros2019learning}};
	\node[font=\fontsize{8pt}{8pt}\selectfont, rotate=90] at (0,-3.5) {Ours};
	\end{tikzpicture}
	\foreach \idx in {1,2,3,4,5,6,7,8,9} {
		\begin{subfigure}[h]{0.095\textwidth}
			\centering
			\foreach \sub in {0,1,2,3,4} {
    			\pgfmathsetmacro\imidx{int(\sub * 9 + \idx)}
    			\includegraphics[width=1.1\textwidth]{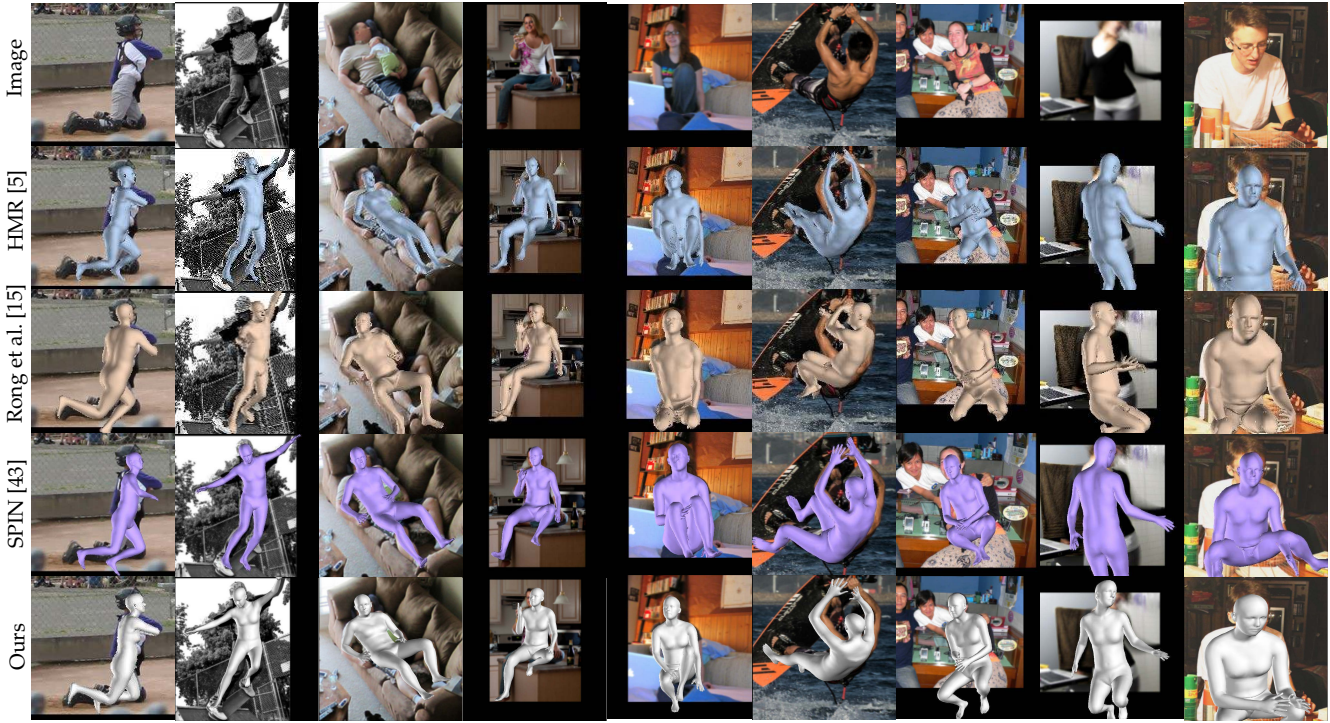}
    		}
		\end{subfigure}
	}
	\vspace{-3mm}
	\caption{Qualitative comparison of reconstruction results on the COCO dataset. }
	\label{fig:coco_dpDemo}
\end{figure*}

\begin{table}[t]
  \centering
  \caption{Quantitative comparison of keypoint localization AP with state-of-the-art methods on the COCO validation set. Results of HMR, CMR, and SPIN are obtained based on their publicly released code and models.}
    \begin{tabular}{l|ccccc}
    \toprule
    Method & AP    & $\textrm{AP}_{50}$ & $\textrm{AP}_{75}$ & $\textrm{AP}_{M}$ & $\textrm{AP}_{L}$ \\
    \midrule
    OpenPose~\cite{cao2019openpose} & 65.3  & 85.2  & 71.3  & 62.2  & 70.7 \\
    SimpleBaseline~\cite{xiao2018simple} & 74.3  & 89.6  & 81.1  & 70.5  & 79.7 \\
    HRNet~\cite{sun2019deep} & \textbf{76.3} & \textbf{90.8} & \textbf{82.9} & \textbf{72.3} & \textbf{83.4} \\
    \midrule
    HMR~\cite{kanazawa2018end} & 18.9  & 47.5  & 11.7  & 21.5  & 17.0 \\
    CMR~\cite{kolotouros2019convolutional} & 9.3   & 26.9  & 4.2   & 11.3  & 8.1 \\
    SPIN~\cite{kolotouros2019learning} & 17.3  & 39.1  & 13.5  & 19.0    & 16.6 \\
    SPIN-HRNet~\cite{kolotouros2019learning} & 21.2  & 45.3  & 18.0  & 22.5  & 20.9 \\
    \midrule
    DaNet-LSTM~\cite{zhang2019danet} & 28.5  & 58.7  & 24.6  & 30.8  & 27.1 \\
    DaNet-GCN & 31.9  & 65.5  & 27.5  & 33.2  & 31.2 \\
    ~ + Dropout & 30.6  & 64.6  & 25.7  & 32.0  & 30.0 \\
    ~ + DropBlock & 32.0  & 66.9  & 27.4  & 33.8  & 30.9 \\
    ~ + PartDrop (Ours) & \textbf{33.8} & \textbf{68.6} & \textbf{29.9} & \textbf{36.0} & \textbf{32.3} \\
    \bottomrule
    \end{tabular}%
  \label{tab:coco}%
\end{table}%

\subsubsection{Comparison on In-the-wild Datasets}
Reconstructing 3D human model on real-world images is much more challenging due to factors such as extreme poses and heavy occlusions.
In our network, the aggregated refinement module and PartDrop training strategy are proposed to enhance its robustness and generalization.
We conduct evaluation experiments on UP-3D, COCO, and 3DPW to demonstrate the efficacy of our method.

\textbf{Evaluation on UP-3D.}
For comparison on the UP-3D dataset, we report quantitative results in the PVE of the reconstructed meshes in Table~\ref{tab:up3d}.
In comparison with previous methods, our method outperforms them across all subsets of UP-3D by a large margin.
Our closest competitor BodyNet~\cite{varol2018bodynet} has the PVE value of 102.5 on LSP, while ours is 88.5.
Moreover, BodyNet~\cite{varol2018bodynet} uses both 2D and 3D estimation as the intermediate representation, which is much more time-consuming than ours.
Reconstruction results on UP-3D are visualized in Fig.~\ref{fig:up3dDemo}.
Compared with other methods, our DaNet could produce more satisfactory results under challenging scenarios.

\begin{figure}[ht]
	\centering
	\begin{tikzpicture}[remember picture,overlay]
	\node[font=\fontsize{8pt}{8pt}\selectfont] at (0.9,-2.1) {Image};
	\node[font=\fontsize{8pt}{8pt}\selectfont] at (2.3,-2.1) {Est. IUV};
	\node[font=\fontsize{8pt}{8pt}\selectfont] at (3.8,-2.1) {w./o. drop};
	\node[font=\fontsize{8pt}{8pt}\selectfont] at (5.25,-2.1) {Dropout};
	\node[font=\fontsize{8pt}{8pt}\selectfont] at (6.7,-2.1) {DropBlock};
	\node[font=\fontsize{8pt}{8pt}\selectfont] at (8.2,-2.1) {PartDrop};
	\end{tikzpicture}
	\foreach \idx in {1,2,3,4,5,6} {
		\begin{subfigure}[h]{0.071\textwidth}
			\centering
			\foreach \sub in {0,1,2} {
    			\pgfmathsetmacro\imidx{int(\sub * 6 + \idx)}
    			\includegraphics[width=1.1\textwidth]{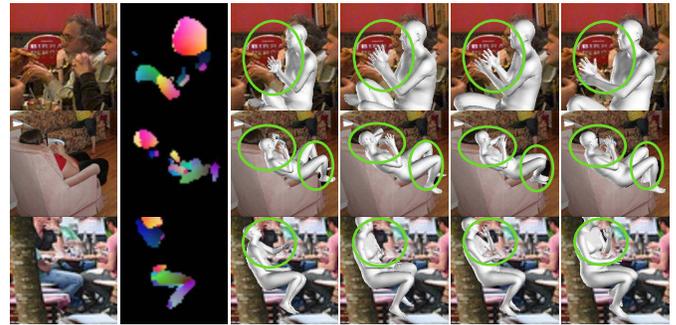}
    		}
		\end{subfigure}
	}
    \caption{Comparison of different dropping out strategies in challenging cases. From left to right: input images, estimated IUV maps, results of models trained without dropping, with Dropout, DropBlock, and PartDrop strategies.}
\label{fig:coco_partdrop}
\end{figure}

\textbf{Evaluation on COCO.}
For evaluation on COCO, we train our model on the mixture of training data from DensePose-COCO and Human3.6M datasets, and perform both qualitative and quantitative comparison on the COCO validation set.
We first show qualitative reconstruction results in Fig.~\ref{fig:coco_dpDemo}, and make comparisons with HMR~\cite{kanazawa2018end}, Rong \etal~\cite{rong2019delving}, and SPIN~\cite{kolotouros2019learning}.
As we can see, our method has better generalization in real-world scenarios with more accurate and well-aligned reconstruction performances.
Our method can produce reasonable results even in cases of extreme poses, occlusions, and incomplete human bodies, while competitors fail or produce visually displeasing results.

To perform quantitative evaluations on COCO, we project keypoints from the estimated SMPL models on the image plane, and compute the Average Percision (AP) based on the keypoint similarity with the ground truth annotations.
We report keypoint localization APs of different approaches in Table~\ref{tab:coco}, where we also include 2D human pose estimation approaches~\cite{cao2019openpose,xiao2018simple,wang2020deep} for comparison.
It can be seen that, in terms of keypoint localization results, approaches for 3D human mesh recovery lag far behind those for 2D human pose estimation.
Among approaches for human mesh recovery, our model achieves significantly higher APs than previous ones.
Compared with the recent state-of-the-art method SPIN~\cite{kolotouros2019learning}, our model improves the mean AP and $\textrm{AP}_{50}$ by 16.5\% and 29.5\%, respectively.
We attribute such remarkable improvements to our decompose-and-aggregate design.
To validate this, we upgrade the backbone of SPIN to HRNet-W64-C~\cite{wang2020deep}, a more powerful classification network, and denote it as SPIN-HRNet.
As shown in Table~\ref{tab:coco}, though SPIN-HRNet has a stronger backbone with more parameters than our whole network, it brings much less gains over SPIN (3.9\% improvement in mean AP from 17.3\% to 21.2\%).
In contrast, our network decomposes the perception tasks and aggregates them efficiently, making our SMPL regressor more effective to handle challenging cases in real-world scenes.

\begin{table}[t]
  \centering
  \caption{Quantitative comparison with state-of-the-art methods on the 3DPW dataset.}
    \begin{tabular}{clccc}
    \toprule
    \multicolumn{2}{c}{Method} & PVE   & MPJPE & MPJPE-PA \\
    \midrule
    \multirow{5}[2]{*}{\begin{sideways}\rotatebox[origin=c]{0}{Temporal}\end{sideways}} & Kanazawa \etal~\cite{kanazawa2019learning} & 139.3 & 116.5 & 72.6 \\
          & Doersch \etal~\cite{doersch2019sim2real} & -     & -     & 74.7 \\
          & Arnab \etal~\cite{arnab2019exploiting} & -     & -     & 72.2 \\
          & Sun \etal~\cite{sun2019human} & -     & -     & 69.5 \\
          & VIBE~\cite{kocabas2020vibe} & 113.4 & 93.5  & 56.5 \\
    \midrule
    \multirow{7}[4]{*}{\begin{sideways}\rotatebox[origin=c]{0}{Frame-based}\end{sideways}} & HMR~\cite{kanazawa2018end} & -     & 130.0   & 76.7 \\
          & CMR~\cite{kolotouros2019convolutional} & -     & -     & 70.2 \\
          & Rong \etal~\cite{rong2019delving} & 152.9 & -     & - \\
          & SPIN~\cite{kolotouros2019learning} & 114.8 & 96.9  & 59.2 \\
          & SPIN-HRNet~\cite{kolotouros2019learning} & 112.4 & 95.4  & 58.5 \\
\cmidrule{2-5}          & DaNet-LSTM~\cite{zhang2019danet} & 114.6 & 92.2  & 56.9 \\
          & Ours  & \textbf{110.8} & \textbf{85.5} & \textbf{54.8} \\
    \bottomrule
    \end{tabular}%
  \label{tab:3dpw}%
\end{table}%

Table~\ref{tab:coco} also presents the comparison of our approach against our previous model DaNet-LSTM~\cite{zhang2019danet}.
DaNet-LSTM has the same network architecture with ours except that its aggregation procedure is performed sequentially along kinetic chains via LSTM.
Based on DaNet-LSTM, we introduce the graph-based aggregation module and PartDrop strategy in this work.
The graph-based aggregation performs feature refinement in parallel for all body parts, while PartDrop regularizes the network and encourages learning features from complementary body parts.
These newly introduced designs can help to improve the robustness and generalization of our model.
As shown in Table~\ref{tab:coco}, both two new components contribute to higher performance in this challenging dataset.
By replacing the LSTM-based aggregation module with the graph-based one, our DaNet-GCN obtains a 6.8\% improvement in $\textrm{AP}_{50}$.
By adopting the PartDrop strategy for training, we further have a 3.1\% improvement in $\textrm{AP}_{50}$.
Taking these two updates together, our approach improves the $\textrm{AP}_{50}$ by 9.9\% over DaNet-LSTM from 58.7\% to 68.6\%.
We can also see from Table~\ref{tab:coco} that other dropping out strategies such as Dropout and DropBlock do not work well as PartDrop and even degrade the performance.
One intuitive explanation for this is that our PartDrop can better imitate the corrupted IUV maps in challenging cases.
As we can observe from Fig.~\ref{fig:coco_partdrop} that the body parts are missing irregularly from the estimated IUV maps due to occlusions.
PartDrop helps to produce more natural and well-aligned results in comparison with its alternatives.

\begin{figure}[t]
	\centering
	\begin{tikzpicture}[remember picture,overlay]
	\node[font=\fontsize{8pt}{8pt}\selectfont, rotate=90] at (0,2.0) {Image};
	\node[font=\fontsize{8pt}{8pt}\selectfont, rotate=90] at (0,0.3) {SPIN~\cite{kolotouros2019learning}};
	\node[font=\fontsize{8pt}{8pt}\selectfont, rotate=90] at (0,-1.45) {Ours};
	\end{tikzpicture}
	\foreach \idx in {1,3,4,5,6} {
		\begin{subfigure}[h]{0.085\textwidth}
			\centering
			\foreach \sub in {0,1,2} {
    			\pgfmathsetmacro\imidx{int(\sub * 6 + \idx)}
    			\includegraphics[width=1.1\textwidth]{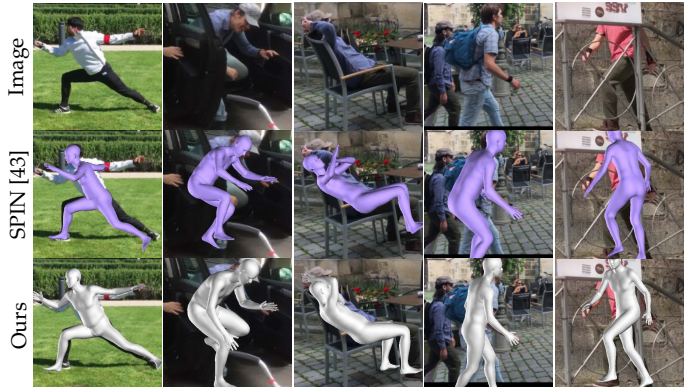}
    		}
		\end{subfigure}
	}
	\vspace{-3mm}
	\caption{Qualitative comparison of reconstruction results on the 3DPW dataset.}
	\label{fig:3dpwDemo}
\end{figure}

\textbf{Evaluation on 3DPW.}
In Table~\ref{tab:3dpw}, we report the results of our approach and other state-of-the-art approaches on the 3DPW test set.
Here, we use the same datasets and training strategy as SPIN~\cite{kolotouros2019learning} and do not use any data from 3DPW for training.
Besides, the valid SMPL parameters fitted in SPIN are adopted as ground-truth labels for those in-the-wild training datasets.
As shown in Table~\ref{tab:3dpw}, our approach reduces the MPJPE-PA by 4.4 mm over SPIN to 54.8 mm, achieving the best performance among frame-based and even temporal approaches.
Table~\ref{tab:3dpw} also includes SPIN-HRNet for comparison, where we can see that there is only a 0.7 mm reduction in MPJPE-PA over SPIN.
Fig.~\ref{fig:3dpwDemo} depicts the qualitative results of our approach.
We can observe that our model has better generalization performances on 3DPW in comparison with SPIN.

\subsubsection{Running Time}
During inference, our method takes about 93ms on a Titan Xp GPU, where the IUV estimation accounts for 60ms while the parameter prediction accounts for the rest 33ms.
The running time and platform of different models are included in Table~\ref{tab:cpr_runtime} for comparison.
Numbers are obtained from respective literature or evaluated using their official implementation.
Overall, our method has a moderate computation cost among regression-based reconstruction methods.

\begin{table}[htbp]
  \centering
  \caption{Comparison of running time (ms) with state-of-the-art models.}
    \begin{tabular}{lcl}
    \toprule
    Method & Run Time & \multicolumn{1}{c}{GPU} \\
    \midrule
    HMR~\cite{kanazawa2018end}   & 40    & GTX 1080 Ti \\
    Pavlakos \etal~\cite{pavlakos2018learning} & 50    & Titan X \\
    NBF~\cite{omran2018neural}   & 110   & Titan Xp \\
    BodyNet~\cite{varol2018bodynet} & 280   & Modern GPU \\
    CMR~\cite{kolotouros2019convolutional}   & 33    & RTX 2080 Ti \\
    DenseRaC~\cite{xu2019denserac} & 75    & Tesla V100 \\
    \midrule
    Ours  & 93    & Titan Xp \\
    \bottomrule
    \end{tabular}%
  \label{tab:cpr_runtime}%
\end{table}%

\subsection{Ablation Study}
To evaluate the effectiveness of the key components proposed in our method, we conduct ablation experiments on Human3.6M under various settings.
We will begin with our baseline network by removing the local streams, aggregated refinement module, and PartDrop strategy in our method.
In other words, the baseline simply uses the global stream of DaNet to predict all parameters.
Moreover, it adopts ResNet101~\cite{he2016deep} as the backbone network for parameter predictions such that the model size of the baseline is comparable to that of the networks used in ablation experiments.

\subsubsection{Intermediate Representation}
To show the superiority of adopting the IUV map as the intermediate representation, our baseline network adopts its alternatives for the shape and pose prediction tasks.
Specifically, the IUV maps are replaced by the convolutional feature maps outputted from the last layer of the FCN or the part segmentation (\ie, $\textit{Index}$ channels of IUV maps).
Note that there is actually no intermediate representation for the approach adopting feature maps as ``intermediate representation''.
As observed from Table~\ref{tab:abla_inter}, the approach adopting IUV maps as intermediate representations achieves the best performance.
In our experiments, we found that the approach without using any intermediate representation is more prone to overfitting to the training set.

\begin{table}[t]
  \centering
  \caption{Performance of approaches adopting different intermediate representations on the Human3.6M dataset.}
    \begin{tabular}{lccc}
    \toprule
    Method & PVE   & MPJPE & MPJPE-PA \\
    \midrule
    ConvFeat & 98.9 & 82.5 & 60.3 \\
    Segmentation & 90.4 & 74.6 & 57.1 \\
    IUV   & \textbf{87.8} & \textbf{71.6} & \textbf{55.4} \\
    \bottomrule
    \end{tabular}%
  \label{tab:abla_inter}%
\end{table}%

\begin{figure}[t]
    \begin{subfigure}[t]{0.23\textwidth}
 		\centering
		\includegraphics[width=1.0\textwidth]{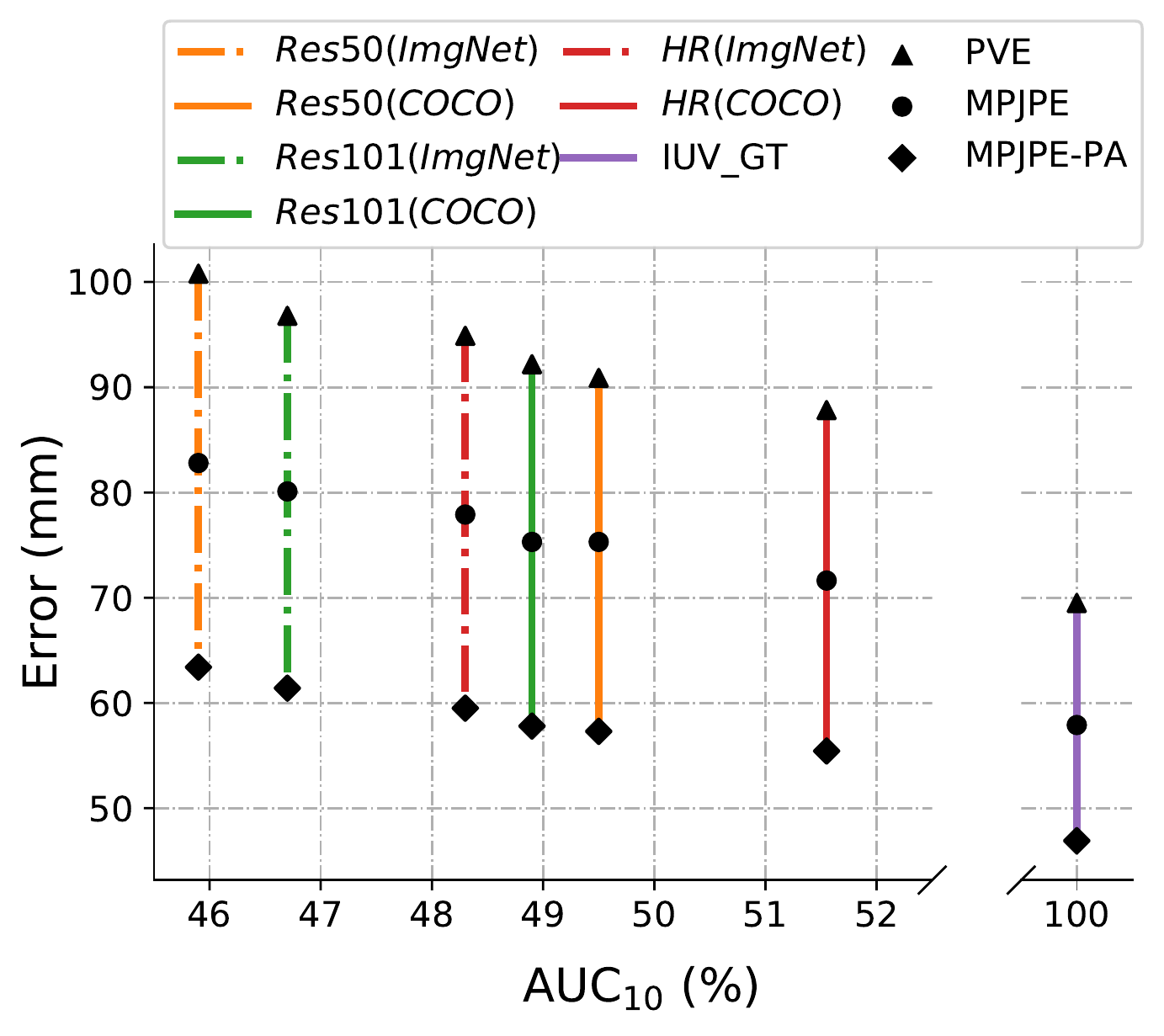}
		\caption{}
		\label{fig:uvi_auc_err}
    \end{subfigure}
    \begin{subfigure}[t]{0.23\textwidth}
 		\centering
		\includegraphics[width=1.0\textwidth]{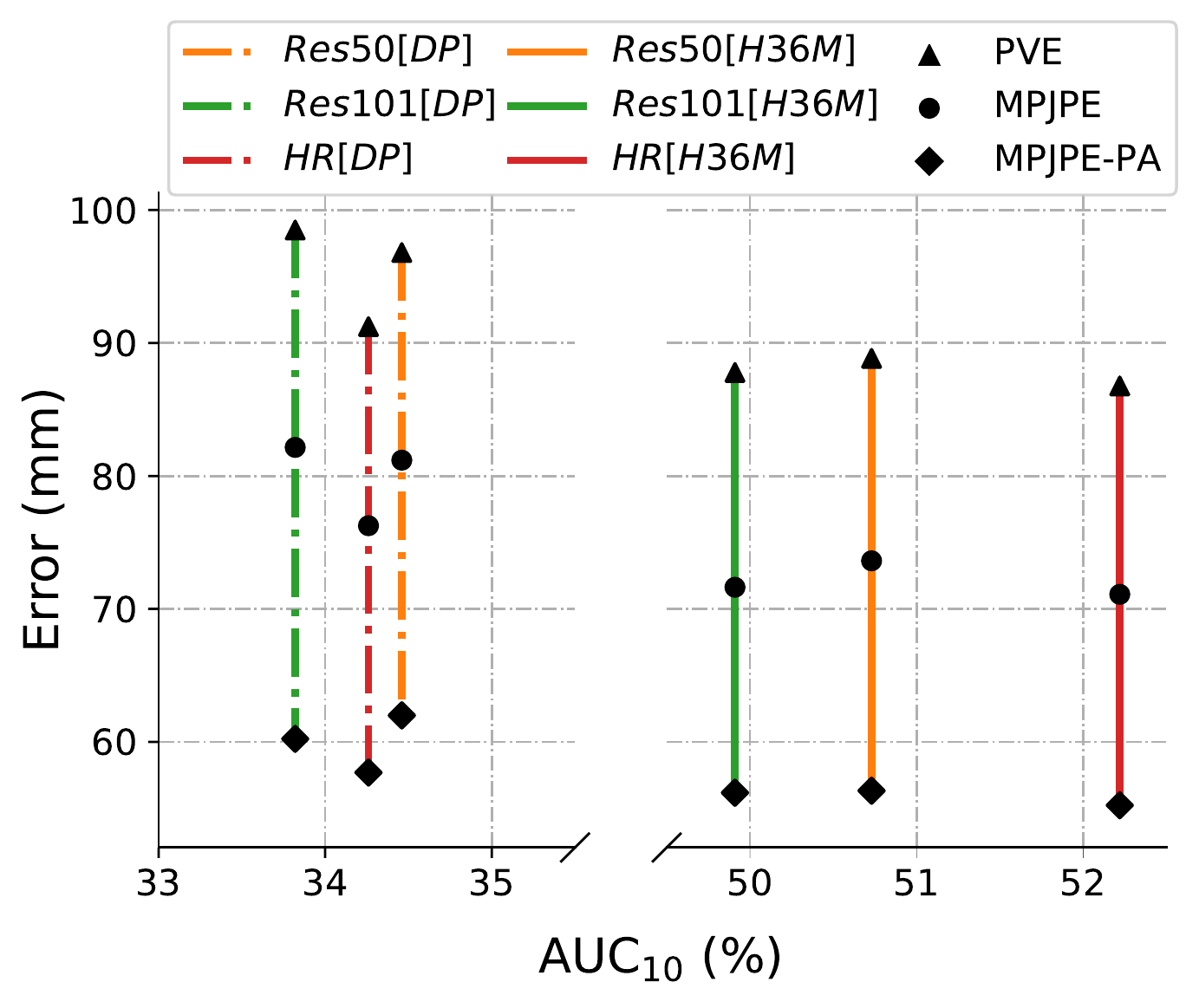}
		\caption{}
		\label{fig:uvi_retr_err}
    \end{subfigure}
    \vspace{-3mm}
    \caption{Reconstruction performance on Human3.6M versus the IUV estimation quality for approaches adopting IUV estimators with different architectures and training strategies. (a) Higher IUV estimation qualities generally contribute to better reconstruction performance. IUV estimators are all trained on Human3.6M but initialized with different models. (b) The IUV estimators trained on Human3.6M with dense supervisions have higher IUV estimation qualities. IUV estimators are all pretrained on COCO and then trained on different datasets. Different IUV estimators are denoted as $\dagger(\star)$ or $\dagger[\ast]$, where $\dagger$ is the architecture, $\star$ and $\ast$ denote the pretrained and training datasets. $IUV\_GT$ stands for taking ground-truth IUV as input. $ImgNet$, $DP$, and $H36M$ abbreviate ImageNet, DensePose-COCO, and Human3.6M, respectively.
    }
\label{fig:abla_dense}
\end{figure}

\textbf{Effect of IUV Estimation Quality.}
We further conduct experiments to investigate the impact of the quality of dense estimation on the final shape and pose prediction performance.
To this end, different architectures or initializations of the IUV estimators are adopted in ablation experiments to produce IUV maps with different qualities.
Specifically, the IUV estimator adopts the pose estimation networks~\cite{xiao2018simple} built upon ResNet-50 and ResNet-101 as alternative architectures, and these models are pretrained on ImageNet~\cite{deng2009imagenet} or COCO~\cite{lin2014microsoft}.
Following the protocol of DensePose~\cite{alp2018densepose}, we measure the quality of dense correspondence estimations via the pointwise evaluation~\cite{alp2018densepose}, where the area under the curve at the threshold of 10cm (\ie, $\text{AUC}_{10}$) is adopted as the metric.
Fig.~\ref{fig:uvi_auc_err} reports the reconstruction results of ablation approaches versus their qualities of IUV estimations.
As we can see, networks with better IUV estimations consistently achieve better reconstruction performance.
To investigate the performance upper bound of adopting IUV maps as intermediate representations, we also report the results of the approach using ground truth IUV maps as input with the removal of the IUV estimator.
As shown in the rightmost result of Fig.~\ref{fig:uvi_auc_err}, the approach learning from the ground truth IUV maps achieves much better performance than using the estimated one outputted from networks,
which means that there is still a large margin for improvement by adopting IUV maps as intermediate representations.

In contrast to the concurrent work~\cite{kolotouros2019convolutional,rong2019delving,xu2019denserac} obtaining IUV maps from the pretrained network of DensePose~\cite{alp2018densepose}, our approach augments the annotation of Human3.6M with the rendered IUV maps so that our IUV estimator can be trained on Human3.6M with dense supervision, which enables our network to have a higher quality of IUV estimation.
To verify this, the IUV estimator is firstly trained on DensePose-COCO or Human3.6M, and then frozen to generate IUV maps for the training of the reconstruction task on Human3.6M.
As can be seen from Fig.~\ref{fig:uvi_retr_err}, approaches with the IUV estimators trained on Human3.6M consistently achieve better performances on both IUV estimation and model reconstruction tasks.

\begin{figure*}[t!]
	\begin{center}
        \centering
		\includegraphics[width=1.0\textwidth]{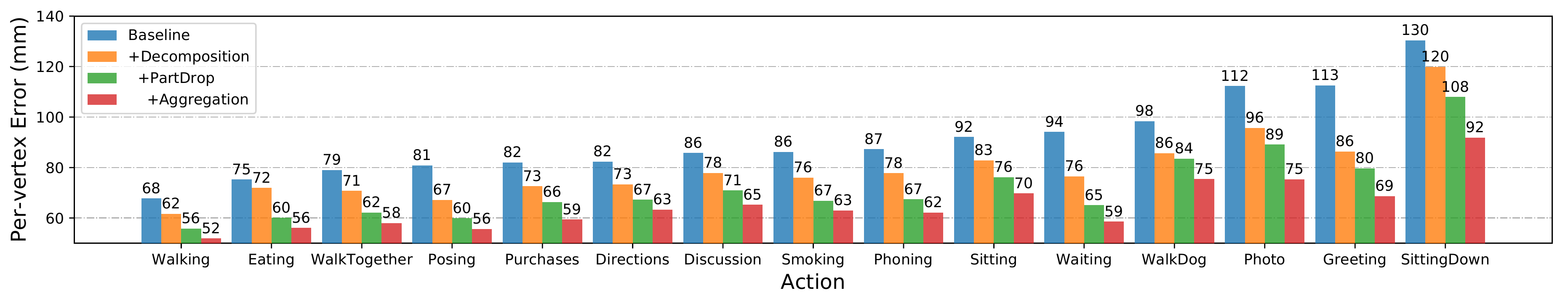}
	\end{center}
	\vspace{-5mm}
    \caption{Reconstruction performance of ablation approaches across different actions on the Human3.6M dataset.}
\label{fig:action_err}
\end{figure*}

\begin{figure}[th]
	\begin{center}
        \begin{subfigure}[b]{0.4\textwidth}
    		\centering
    		\includegraphics[width=1.0\textwidth]{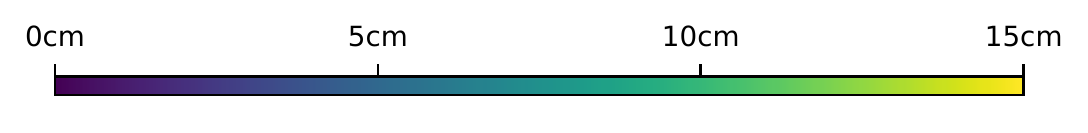}
    	\end{subfigure}
    	
    	\begin{subfigure}[b]{0.13\textwidth}
     		\centering
    		\includegraphics[height=20mm]{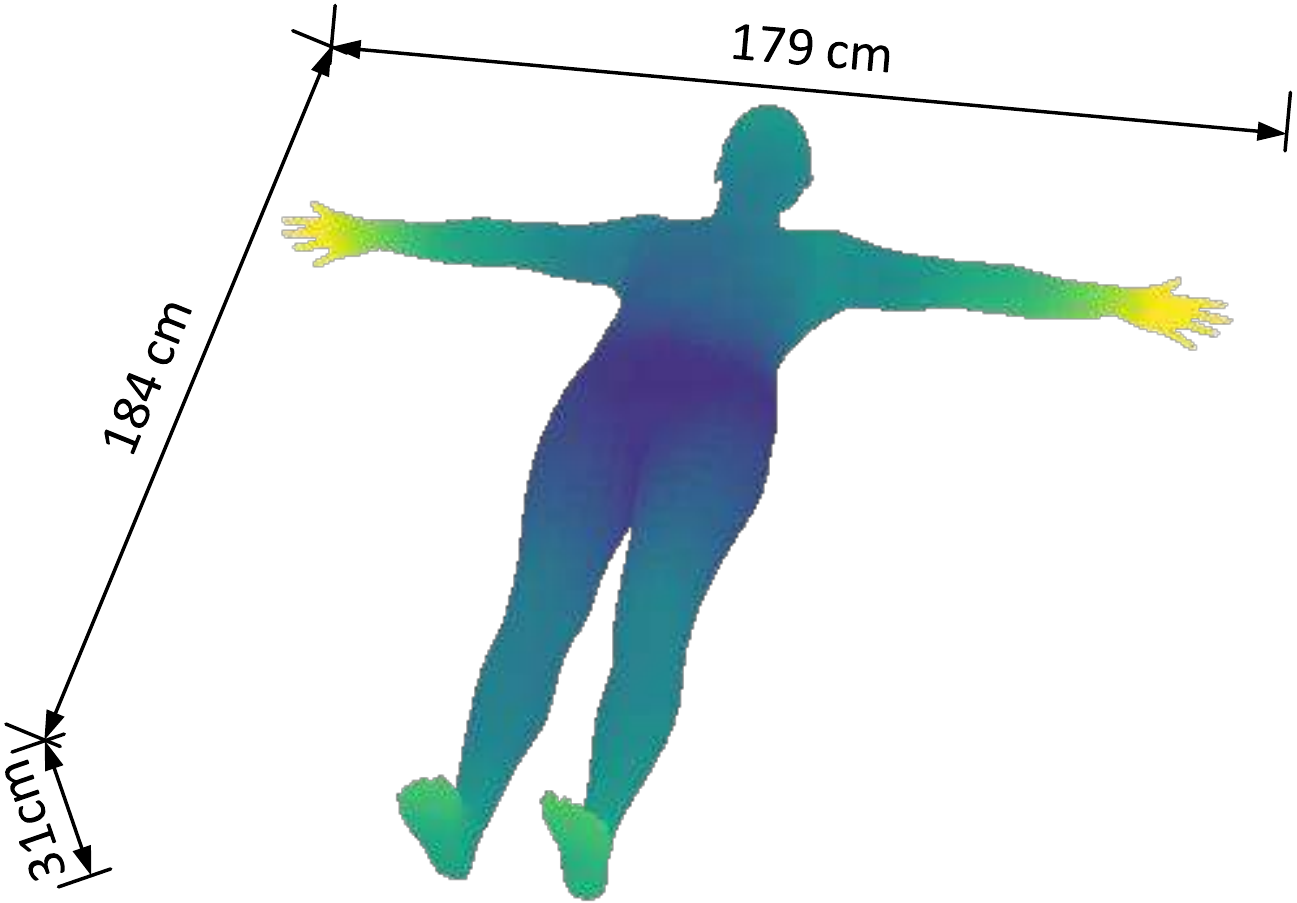}
    		\caption{}
    		\label{fig:surf_err_base}
    	\end{subfigure}
     	\hspace{1mm}
    	\begin{subfigure}[b]{0.1\textwidth}
    		\centering
    		\includegraphics[height=20mm]{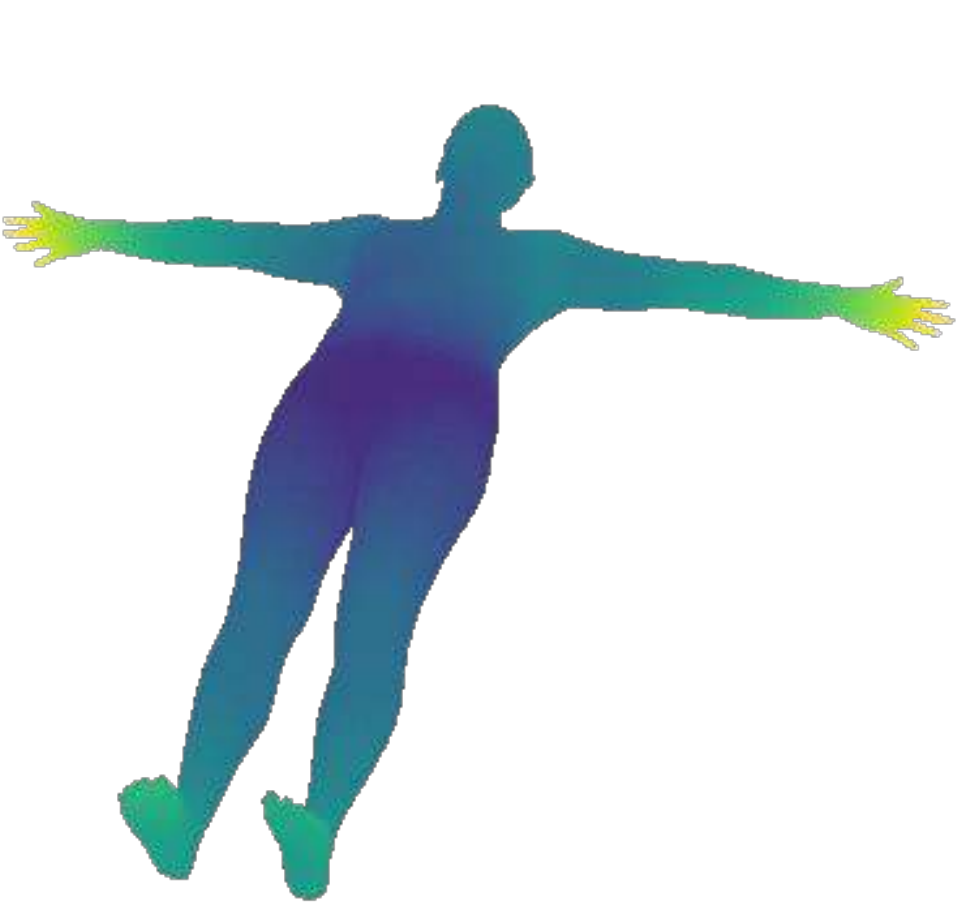}
    		\caption{}
    		\label{fig:surf_err_local}
    	\end{subfigure}
    	\hspace{1mm}
    	\begin{subfigure}[b]{0.1\textwidth}
    		\centering
    		\includegraphics[height=20mm]{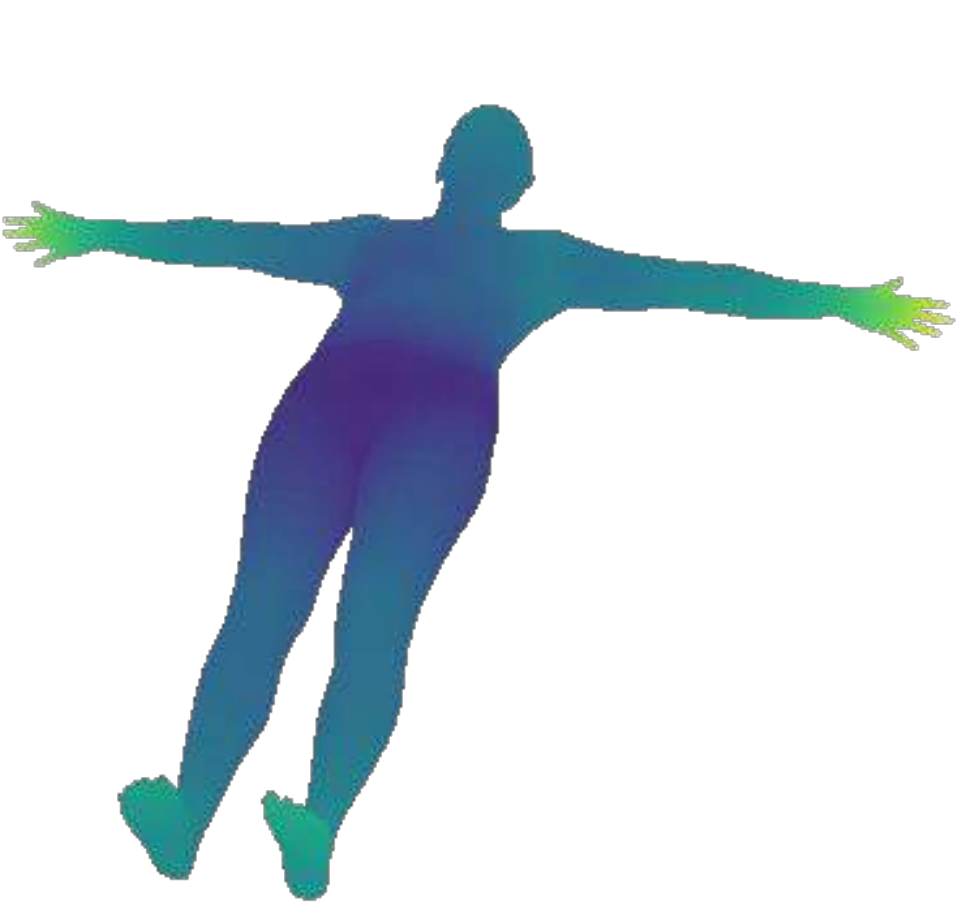}
    		\caption{}
    		\label{fig:surf_err_drop}
    	\end{subfigure}
    	\hspace{1mm}
    	\begin{subfigure}[b]{0.1\textwidth}
    		\centering
    		\includegraphics[height=20mm]{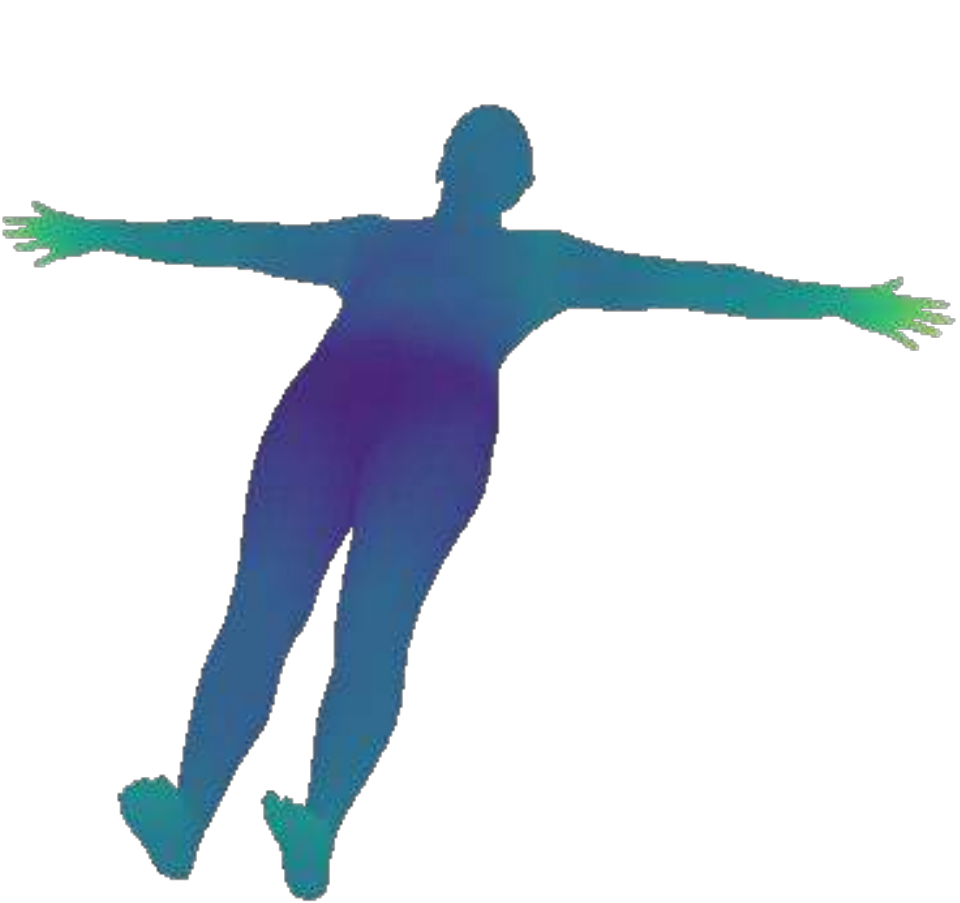}
    		\caption{}
    		\label{fig:surf_err_refine}
    	\end{subfigure}
	\end{center}
	\vspace{-3mm}
    \caption{Comparison of the average per-vertex error upon the model surface for ablation approaches on the Human3.6M dataset. (a) The baseline approach using one stream only. (b) The approach using multiple streams for decomposed perception. (c) The approach using decomposed perception and PartDrop strategies. (d) Our final approach with the aggregated refinement.
    }
\label{fig:surface_err}
\end{figure}

\subsubsection{Decomposed Perception}
The decomposed perception provides fined-grained information for detailed pose estimation.
To validate the effectiveness of such a design, we report the performance of the approaches using one-stream and multiple streams in Table~\ref{tab:abla_partial}, where the \textit{D-Net} denotes the variant of our DaNet without using the aggregated refinement module and PartDrop strategy.
Results in PVE-S and PVE-P are also reported in Table~\ref{tab:abla_partial} for separately studying the efficacy of the decomposed design on the shape and pose predictions.
It can be seen that the reconstruction performance metric PVE is actually dominated by the PVE-P metric.
Comparison of the first and second rows in Table~\ref{tab:abla_partial} shows that using multiple streams has barely effects on the shape prediction but brings a significant improvement in the pose prediction (\ie, the PVE-P value drops more than 14\%).
We also report results to validate the use of different ratios $\alpha_k$ and the simplification of partial IUV maps.
In the 3rd and 4th rows of Table~\ref{tab:abla_partial}, \textit{D-Net-ES} adopts equal scales with all $\alpha_k$s set to 0.5, while \textit{D-Net-AP} adopts partial IUV maps with all body parts.
As can be seen, such modifications degrade the performance, which is due to two facts that (i) the proportions of body parts are different and (ii) the rotational states of different body joints are relatively independent and involving irrelevant body parts could disturb the inference of the target joint rotations.

To visualize the reconstruction performance on different body areas, Fig.~\ref{fig:surface_err} depicts the average per-vertex error with respect to the surface areas of the human model.
As shown in Fig.~\ref{fig:surf_err_base}, for the baseline network, the per-vertex errors of limb parts (hands, feet) are much higher than that of the torso.
By comparing Figs.~\ref{fig:surf_err_base} and~\ref{fig:surf_err_local}, we can conclude that our decomposed perception design alleviates the above issue and achieves much better reconstruction performance on limb parts.
Reconstruction performances across different actions on Human3.6M are also reported in Fig.~\ref{fig:action_err} for comprehensive evaluations.
We can see that the decomposed perception design reduces reconstruction errors consistently for all actions.

\begin{table}[t]
  \centering
  \caption{Performance of approaches using different perception strategies on the Human3.6M dataset.}
    \begin{tabular}{lccccc}
    \toprule
    Method & PVE   & PVE-S & PVE-P & MPJPE & MPJPE-PA \\
    \midrule
    Baseline & 87.8 & 38.0 & 76.3 & 71.6 & 55.4 \\
    D-Net & \textbf{74.3} & \textbf{36.3} & \textbf{64.0} & \textbf{61.8} & \textbf{48.5} \\
    D-Net-ES & 76.1 & 36.6 & 65.5 & 63.1 & 49.8 \\
    D-Net-AP & 76.8 & 36.8 & 65.8 & 63.4 & 49.5 \\
    \bottomrule
    \end{tabular}%
  \label{tab:abla_partial}%
\end{table}%

\begin{figure}[t]
	\begin{center}
    	\begin{subfigure}[b]{0.15\textwidth}
     		\centering
    		\includegraphics[width=1.2\textwidth]{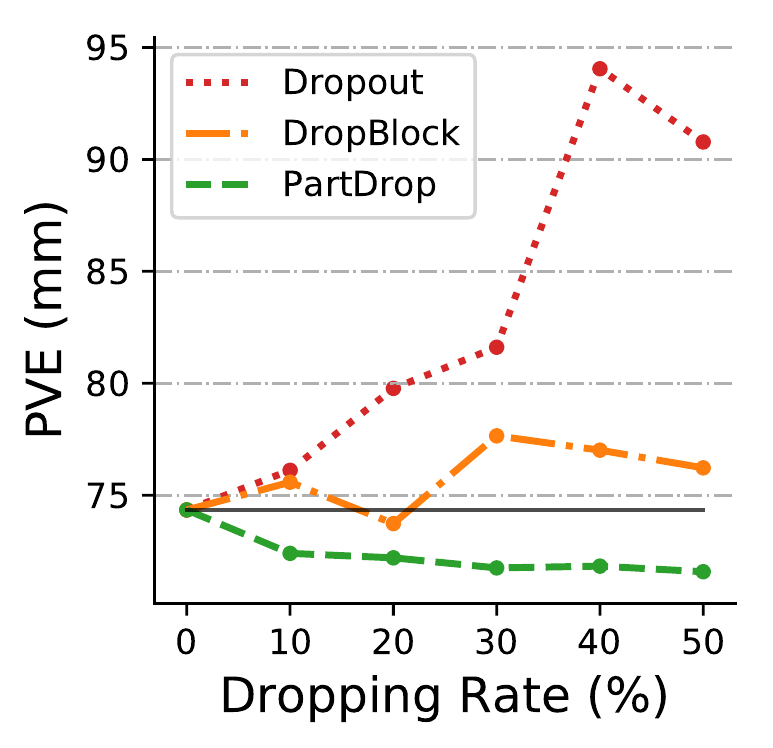}
    		\caption{Full}
    		\label{fig:drop_curve_pve}
    	\end{subfigure}
     	\hspace{1mm}
    	\begin{subfigure}[b]{0.15\textwidth}
    		\centering
    		\includegraphics[width=1.2\textwidth]{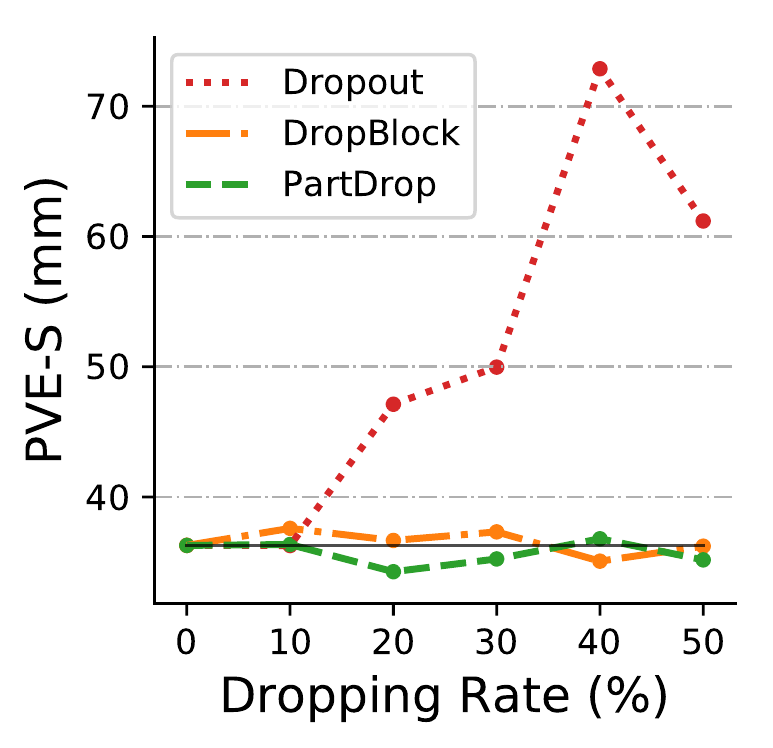}
    		\caption{Shape}
    		\label{fig:drop_curve_t}
    	\end{subfigure}
    	\hspace{1mm}
    	\begin{subfigure}[b]{0.15\textwidth}
    		\centering
    		\includegraphics[width=1.2\textwidth]{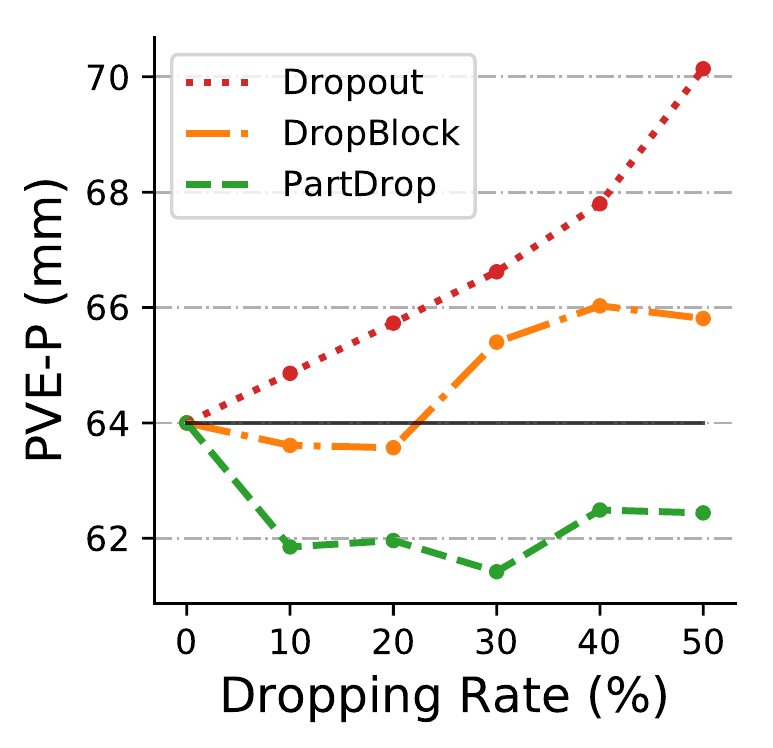}
    		\caption{Pose}
    		\label{fig:drop_curve_p}
    	\end{subfigure}
	\end{center}
	\vspace{-3mm}
    \caption{Comparison of reconstruction performance for approaches using different dropping out strategies on the Human3.6M dataset. (a)(b)(c) report results with metrics of PVE, PVE-S, and PVE-P to reveal the quality of the full model recovery, shape recovery, and pose recovery across different dropping rates, respectively.}
\label{fig:drop_curve}
\end{figure}

\subsubsection{Part-based Dropout}
The proposed Part-based Dropout (PartDrop) strategy drops IUV values in contiguous regions at the granularity level of body parts.
Such a dropping out strategy can effectively regularize the neural network by removing semantic information from foreground areas of intermediate representations.
In this subsection, we conduct experiments to validate its effectiveness and evaluate the impact of the dropping rate on the reconstruction performance.

To validate the superiority of our PartDrop strategy, we adopt DropBlock~\cite{ghiasi2018dropblock} and Dropout~\cite{srivastava2014dropout} as alternative strategies to drop values from intermediate representations during training.
For DropBlock, following the setting of~\cite{ghiasi2018dropblock}, the size of the block to be dropped is set to 7 in our experiments.
For fair comparison, only the foreground pixels are involved in counting the dropping rate.
Fig.~\ref{fig:drop_curve} reports the performance of the full model reconstruction as well as its shape and pose components under different strategies across different dropping rates.
It can be seen that the performance gains brought by dropping out strategies mainly come from the pose prediction tasks since the evaluation metric PVE is dominated by its pose component \mbox{PVE-P}.
Among three strategies, Dropout is the worst and its performance deteriorates quickly when increasing the rate of dropping out.
DropBlock works better than Dropout and brings marginal gains when the dropping rate is less than 20\%.
Though we can see from the PVE-S curves in Fig.~\ref{fig:drop_curve_t} that DropBlock has comparable results with PartDrop on shape prediction when the dropping rate is larger than 40\%, its pose prediction results degrade significantly as shown in Fig.~\ref{fig:drop_curve_p}.
We hypothesize that the removal of a large area of block makes DropBlock similar to PartDrop for the global perception but does harm to the local perception for pose prediction.
Compared with these two alternatives, the proposed PartDrop is more robust to the dropping rate and achieves the best results at a dropping rate around 30\%.
The above comparison of unit-wise, block-wise, and part-wise dropping strategies suggest that removing features in a structured manner is crucial to our reconstruction task, where PartDrop performs best among them.
The efficacy of PartDrop can be also validated from the reconstruction error reduction shown in Fig.~\ref{fig:action_err} and Fig.~\ref{fig:surf_err_drop}.

\begin{figure}[t]
	\centering
	\begin{tikzpicture}[remember picture,overlay]
	\node[font=\fontsize{8pt}{8pt}\selectfont, rotate=90] at (0,2.7) {Image};
	\node[font=\fontsize{8pt}{8pt}\selectfont, rotate=90] at (0,1.1) {w/o Ref.};
	\node[font=\fontsize{8pt}{8pt}\selectfont, rotate=90] at (0,-0.5) {Direct Ref.};
	\node[font=\fontsize{8pt}{8pt}\selectfont, rotate=90] at (0,-2.1) {Pos.-aided Ref.};
	\end{tikzpicture}
	\foreach \idx in {1,2,3,4,5} {
		\begin{subfigure}[h]{0.08\textwidth}
			\centering
			\foreach \imtype in {h36m_img,h36m_pd,h36m_drf,h36m_rf} {
    			\includegraphics[width=1.1\textwidth]{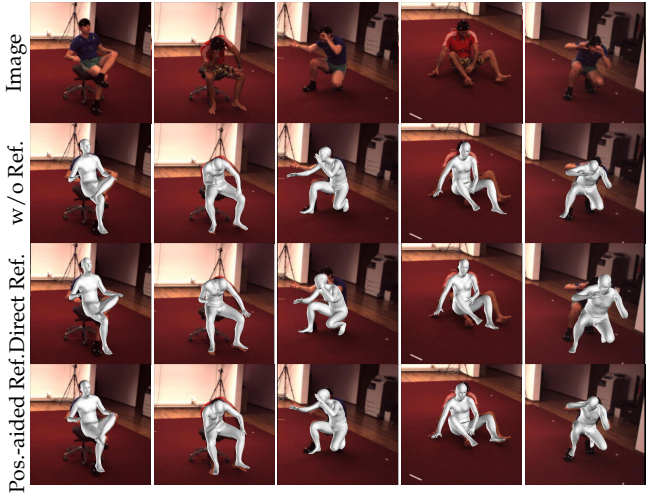}
    		}
		\end{subfigure}
	}
	\vspace{-3mm}
	\caption{Example results of approaches without refinement, or using direct / position-aided refinement strategies.}
	\vspace{-3mm}
	\label{fig:h36m_abla_Demo}
\end{figure}

\begin{table}[t]
  \centering
  \caption{Performance of approaches using different feature refinement strategies on the Human3.6M dataset.}
    \begin{tabular}{l|ccc}
    \toprule
    Refinement Strategy & PVE   & MPJPE & MPJPE-PA \\
    \midrule
    w/o Ref. & 71.7 & 59.1 & 46.1 \\
    \midrule
    Direct Ref. & 70.3 & 58.1 & 45.5 \\
    Pos.-implicit Ref. & 69.2 & 56.5 & 44.7 \\
    Pos.-aided Ref. & \textbf{66.5} & \textbf{54.6} & \textbf{42.9} \\
    \bottomrule
    \end{tabular}%
  \label{tab:abla_refine}%
\end{table}%

\begin{figure}[t]
	\begin{center}
        \centering
        \hspace{-8mm}
    	\begin{subfigure}[b]{0.15\textwidth}
     		\centering
    		\includegraphics[height=1.0\textwidth]{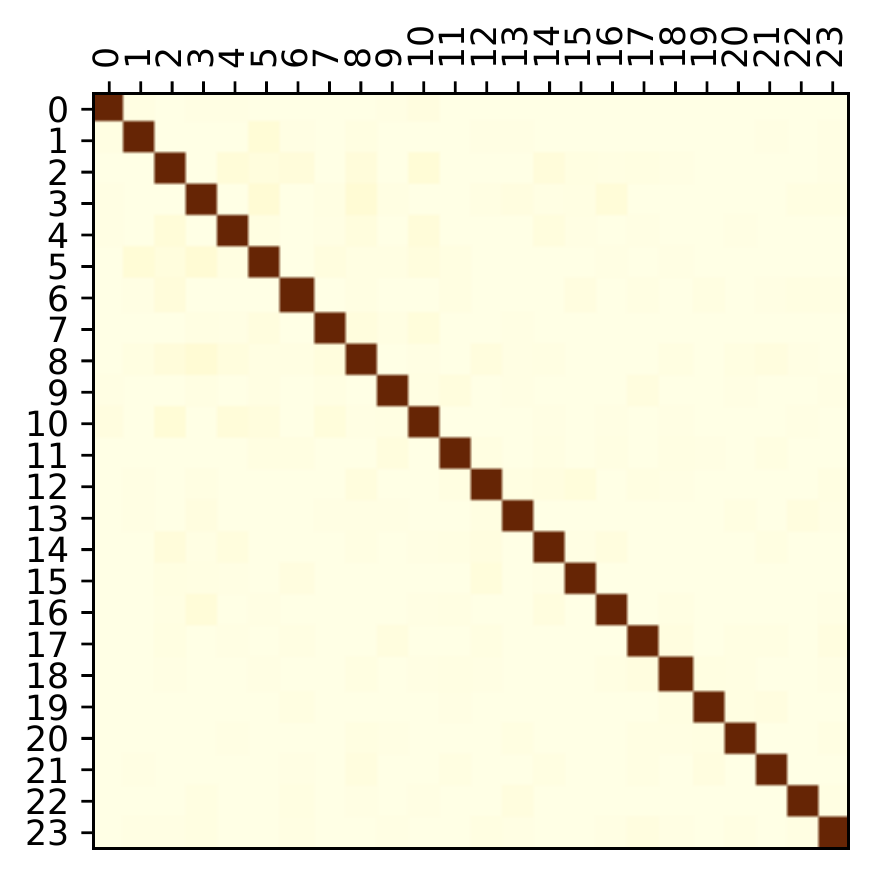}
    		\caption{}
    		\label{fig:rot_feat_corr}
    	\end{subfigure}
    	\begin{subfigure}[b]{0.15\textwidth}
    		\centering
    		\includegraphics[height=1.0\textwidth]{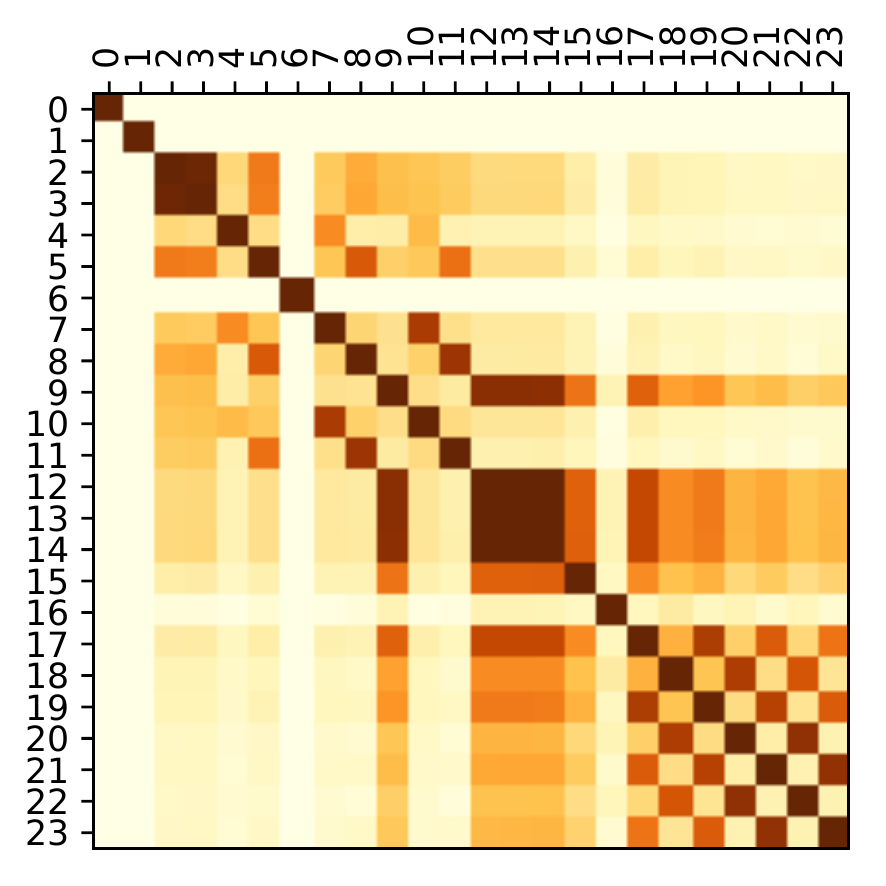}
    		\caption{}
    		\label{fig:imp_pos_feat_corr}
    	\end{subfigure}
    	\begin{subfigure}[b]{0.15\textwidth}
    		\centering
    		\includegraphics[height=1.0\textwidth]{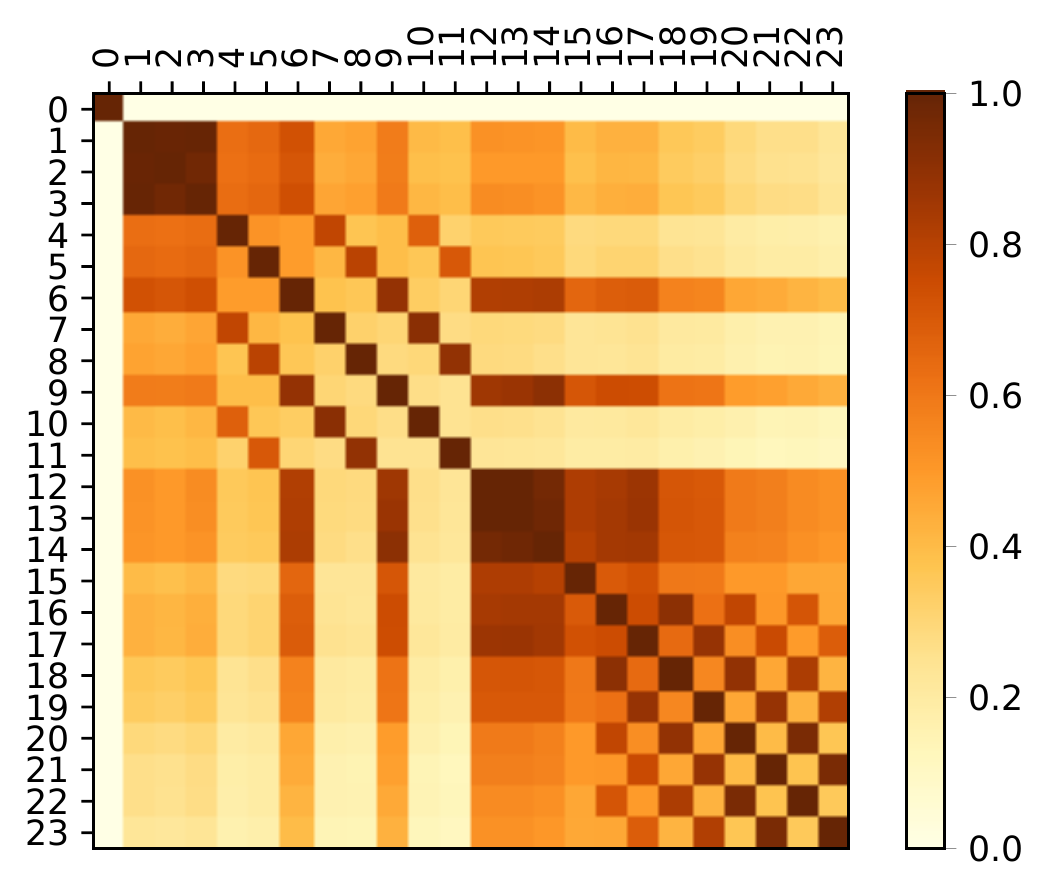}
    		\caption{}
    		\label{fig:pos_feat_corr}
    	\end{subfigure}
	\end{center}
	\vspace{-3mm}
    \caption{Correlation matrices of the features extracted from (a) rotation, (b) implicit position, and (c) position feature spaces.}
\label{fig:rot_pos_feat_corr}
\end{figure}

\subsubsection{Aggregated Refinement}
Our aggregated refinement module is proposed to impose spacial structure constraints upon rotation-based pose features.
As observed from Fig.~\ref{fig:surf_err_refine} and Fig.~\ref{fig:action_err}, the aggregation in DaNet effectively reduces the reconstruction errors across all surface areas and human actions considerably.

A straightforward strategy to refine the feature would be conducting refinement between the rotation features directly.
In such a \textit{direct} refinement strategy, the first and third steps of our refinement procedure are removed and the rotation features are directly refined by the graph convolution layers of the second step.
The features outputted from the last refinement layer are also added with the original rotation features in a residual manner and then used to predict joint rotations.
For fair comparison, the refinement layer number of the direct strategy is equal to the number of the layers involved in the three steps of the position-aided strategy.

\textbf{Rotation Feature Space \vs Position Feature Space.}
The proposed position-aided refinement strategy performs refinement in the position feature space instead of the rotation feature space.
The graphs $\bm{A}^{r2p}$ and $\bm{A}^{p2r}$ of the first and last refinement steps are customized to connect the rotation and position feature spaces.
The graph $\bm{A}^{r2p}$ collects rotation features to the position feature space, while the graph $\bm{A}^{p2r}$ converts position features back to the rotation feature space.
To validate their functions, we discard position supervisions from the objective $\mathcal{L}_{refine}$ during refinement.
We refer to this strategy as the \textit{position-implicit} refinement strategy since the position feature space is built in an implicit manner.
The only difference between the direct and position-implicit refinement strategies is that, in the latter one, there are two mapping operations performed before and after the refinement.
We report the results of the approaches using direct, position-implicit, position-aided strategies in Table~\ref{tab:abla_refine} for comparison.
It can be seen that the position-implicit strategy achieves inferior results than the position-aided strategy but better results than the direct strategy, which means that the implicit position space still works better than the rotation space for feature refinement.
Example results of the approach using the direct or position-aided refinement strategy are also depicted in Fig.~\ref{fig:h36m_abla_Demo} for comparison.
We can see that the position-aided refinement helps to handle challenging cases and produce more realistic and well-aligned results, while the direct refinement brings marginal to no improvement.

The reason behind the inferior performance of the direct refinement is that the correlation between rotation features is weak, and the messages of neighboring rotation features are generally irrelevant to refine the target rotation feature.
Our refinement module builds an auxiliary position feature space for feature refinement, making it much more efficient than that in the original rotation feature space.
To verify this, we extract the features before refinement from the rotation, implicit position, and position spaces of the three strategies mentioned above, and compute the correlations between features of different body joints.
Fig.~\ref{fig:rot_pos_feat_corr} shows the comparison of correlation matrices of these three types of features.
As observed from Fig.~\ref{fig:rot_feat_corr}, the correlation matrix of rotation features approximates to an identity matrix, meaning that the correlations between the rotation features of different joints are rather weak even for two adjacent joints.
By contrast, for implicit position features in Fig.~\ref{fig:imp_pos_feat_corr} and position features in Fig.~\ref{fig:pos_feat_corr}, the correlations between features of adjacent joints are much higher, making it more feasible to refine features with the messages from neighboring joints.

\begin{table}[t]
  \centering
  \caption{Ablation study of using learnable graph edge and PartDrop strategies on the Human3.6M dataset.}
    \begin{tabular}{l|ccc}
    \toprule
    Method & PVE   & MPJPE & MPJPE-PA \\
    \midrule
    D-Net & 74.3 & 61.8 & 48.5 \\
    ~ + PartDrop & 71.7 & 59.1 & 46.1 \\
    \midrule
    D-Net+Direct & 72.1 & 59.4 & 46.9 \\
    ~ + LearntEdge & 72.7 & 59.6 & 47.0 \\
    ~~~ + PartDrop & 70.3 & 58.1 & 45.5 \\
    \midrule
    D-Net+Pos.-aided & 70.8 & 57.1 & 45.9 \\
    ~ + LearntEdge & 68.9 & 55.8 & 44.9 \\
    ~~~ + PartDrop & \textbf{66.5} & \textbf{54.6} & \textbf{42.9} \\
    \bottomrule
    \end{tabular}%
  \label{tab:abla_weight}%
\end{table}%

\begin{figure}[t]
	\begin{center}
        \centering
        \hspace{-8mm}
    	\begin{subfigure}[b]{0.11\textwidth}
    		\centering
    		\includegraphics[height=1.1\textwidth]{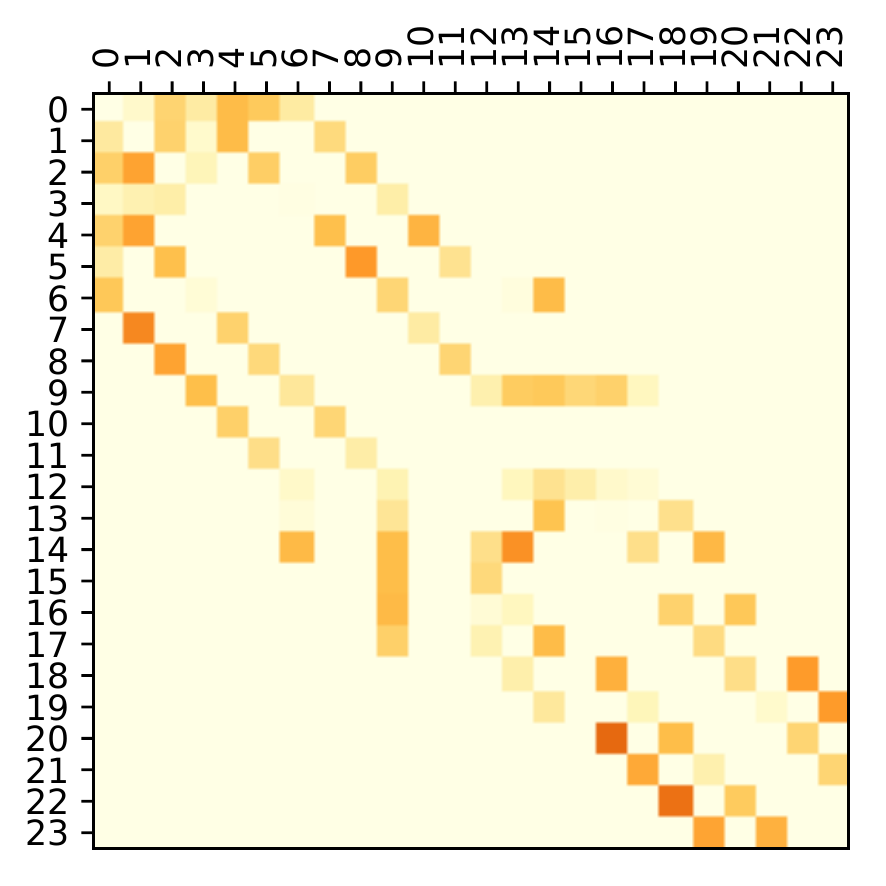}
    		\caption{}
    		\label{fig:direct_mask}
    	\end{subfigure}
    	\begin{subfigure}[b]{0.11\textwidth}
     		\centering
    		\includegraphics[height=1.1\textwidth]{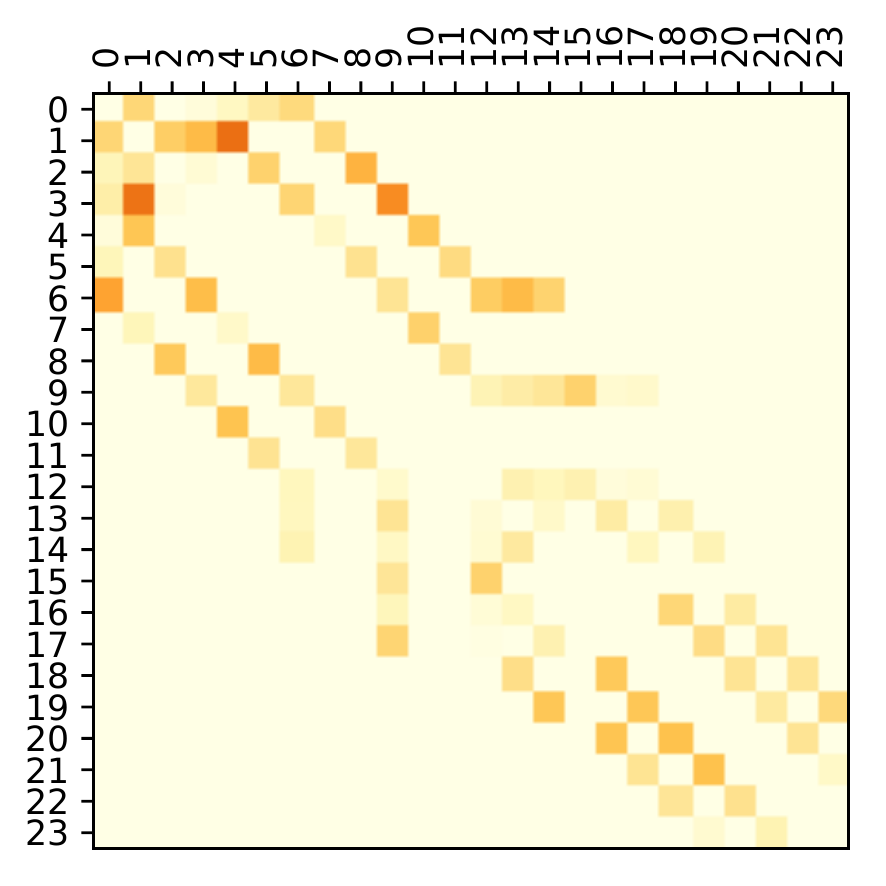}
    		\caption{}
    		\label{fig:direct_mask_drop}
    	\end{subfigure}
    	\begin{subfigure}[b]{0.11\textwidth}
     		\centering
    		\includegraphics[height=1.1\textwidth]{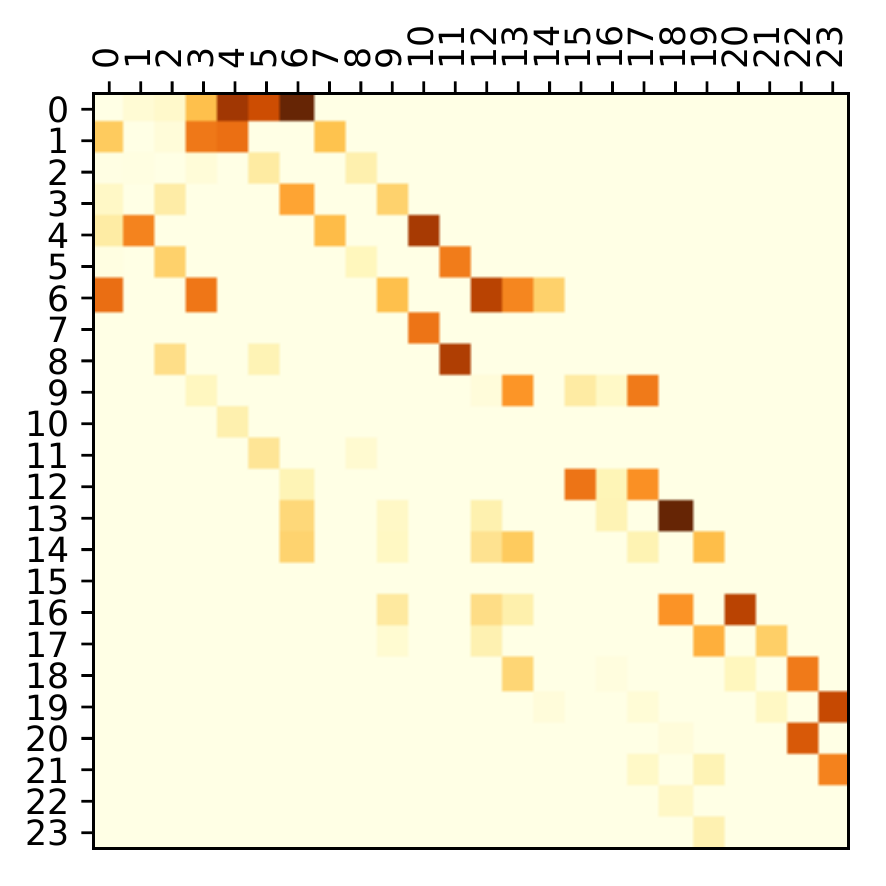}
    		\caption{}
    		\label{fig:aided_mask}
    	\end{subfigure}
    	\begin{subfigure}[b]{0.11\textwidth}
     		\centering
    		\includegraphics[height=1.1\textwidth]{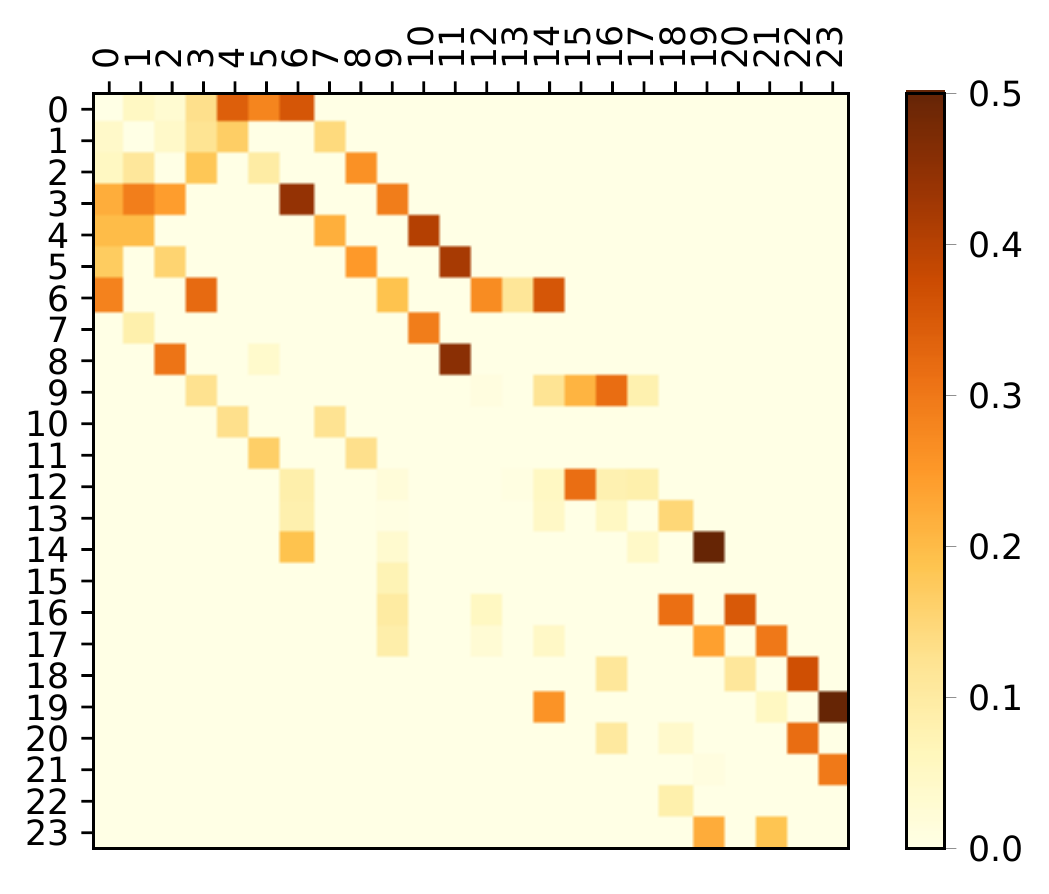}
    		\caption{}
    		\label{fig:aided_mask_drop}
    	\end{subfigure}
	\end{center}
	\vspace{-3mm}
    \caption{Visualization of learned edge weighting matrices under different training settings. (a)(b) Direct refinement without and with PartDrop. (c)(d) Position-aided refinement without and with PartDrop.}
\label{fig:mask_vis}
\end{figure}

\textbf{Benefit from Learnable Graph Edge and PartDrop.}
The learnable edge weighting matrix $\bm{M}$ of the refinement graph contributes to better balancing the importance of neighboring messages, while the PartDrop strategy helps to encourage the network to leverage more information from neighboring joints.
To verify their effectiveness during feature refinement, Table~\ref{tab:abla_weight} reports the results of the ablation approaches incrementally adopting the learnable edge in the refinement graph and the PartDrop strategy, 
where \textit{D-Net+Direct} and \textit{D-Net+Pos.-aided} adopt the refinement module with the direct and position-aided strategy, respectively.
It can be seen that, for the direct refinement, the performance gains mainly come from the PartDrop strategy.
In contrast, for the position-aided refinement, the performance gains are attributed to both the learnable edge and the PartDrop strategy.
Fig.~\ref{fig:mask_vis} depicts the learned edge weighting matrices of different ablation approaches.
As observed, the learned edge weighting matrices of the direct refinement are relatively flat with lower values.
When using the PartDrop strategy, the learnable values of most edges in the refinement graph rise for the position-aided refinement, while such a phenomenon is not observed for the direct refinement.
We conjecture that the PartDrop strategy brings gains from two perspectives.
First, PartDrop regularizes the backbone feature extractor to focus on more complementary regions in intermediate representations for better feature exploitation.
Second, PartDrop encourages the refinement module to borrow more information from neighbors in the position feature space for better feature refinement.

%% file: bare_jrnl_compsoc.bbl
\begin{thebibliography}{10}
\providecommand{\url}[1]{#1}
\csname url@samestyle\endcsname
\providecommand{\newblock}{\relax}
\providecommand{\bibinfo}[2]{#2}
\providecommand{\BIBentrySTDinterwordspacing}{\spaceskip=0pt\relax}
\providecommand{\BIBentryALTinterwordstretchfactor}{4}
\providecommand{\BIBentryALTinterwordspacing}{\spaceskip=\fontdimen2\font plus
\BIBentryALTinterwordstretchfactor\fontdimen3\font minus
  \fontdimen4\font\relax}
\providecommand{\BIBforeignlanguage}[2]{{%
\expandafter\ifx\csname l@#1\endcsname\relax
\typeout{** WARNING: IEEEtran.bst: No hyphenation pattern has been}%
\typeout{** loaded for the language `#1'. Using the pattern for}%
\typeout{** the default language instead.}%
\else
\language=\csname l@#1\endcsname
\fi
#2}}
\providecommand{\BIBdecl}{\relax}
\BIBdecl

\bibitem{loper2015smpl}
M.~Loper, N.~Mahmood, J.~Romero, G.~Pons-Moll, and M.~J. Black, ``Smpl: A
  skinned multi-person linear model,'' \emph{ACM Transactions on Graphics},
  vol.~34, no.~6, pp. 248:1--248:16, 2015.

\bibitem{bogo2016keep}
F.~Bogo, A.~Kanazawa, C.~Lassner, P.~Gehler, J.~Romero, and M.~J. Black, ``Keep
  it smpl: Automatic estimation of 3d human pose and shape from a single
  image,'' in \emph{Proceedings of the European Conference on Computer
  Vision}.\hskip 1em plus 0.5em minus 0.4em\relax Cham: Springer, 2016, pp.
  561--578.

\bibitem{lassner2017unite}
C.~Lassner, J.~Romero, M.~Kiefel, F.~Bogo, M.~J. Black, and P.~V. Gehler,
  ``Unite the people: Closing the loop between 3d and 2d human
  representations,'' in \emph{Proceedings of the IEEE Conference on Computer
  Vision and Pattern Recognition}, 2017, pp. 6050--6059.

\bibitem{tung2017self}
H.-Y. Tung, H.-W. Tung, E.~Yumer, and K.~Fragkiadaki, ``Self-supervised
  learning of motion capture,'' in \emph{Advances in Neural Information
  Processing Systems}, 2017, pp. 5236--5246.

\bibitem{kanazawa2018end}
A.~Kanazawa, M.~J. Black, D.~W. Jacobs, and J.~Malik, ``End-to-end recovery of
  human shape and pose,'' in \emph{Proceedings of the IEEE Conference on
  Computer Vision and Pattern Recognition}, 2018, pp. 7122--7131.

\bibitem{pavlakos2018learning}
G.~Pavlakos, L.~Zhu, X.~Zhou, and K.~Daniilidis, ``Learning to estimate 3d
  human pose and shape from a single color image,'' in \emph{Proceedings of the
  IEEE Conference on Computer Vision and Pattern Recognition}, 2018, pp.
  459--468.

\bibitem{omran2018neural}
M.~Omran, C.~Lassner, G.~Pons-Moll, P.~Gehler, and B.~Schiele, ``Neural body
  fitting: Unifying deep learning and model based human pose and shape
  estimation,'' in \emph{International Conference on 3D Vision}.\hskip 1em plus
  0.5em minus 0.4em\relax IEEE, 2018, pp. 484--494.

\bibitem{anguelov2005scape}
D.~Anguelov, P.~Srinivasan, D.~Koller, S.~Thrun, J.~Rodgers, and J.~Davis,
  ``Scape: shape completion and animation of people,'' in \emph{ACM
  Transactions on Graphics}, vol.~24, no.~3.\hskip 1em plus 0.5em minus
  0.4em\relax ACM, 2005, pp. 408--416.

\bibitem{chen2014articulated}
X.~Chen and A.~L. Yuille, ``Articulated pose estimation by a graphical model
  with image dependent pairwise relations,'' in \emph{Advances in Neural
  Information Processing Systems}, 2014, pp. 1736--1744.

\bibitem{chu2016structured}
X.~Chu, W.~Ouyang, H.~Li, and X.~Wang, ``Structured feature learning for pose
  estimation,'' in \emph{Proceedings of the IEEE Conference on Computer Vision
  and Pattern Recognition}, 2016, pp. 4715--4723.

\bibitem{zhang2019danet}
H.~Zhang, J.~Cao, G.~Lu, W.~Ouyang, and Z.~Sun, ``Danet:
  Decompose-and-aggregate network for 3d human shape and pose estimation,'' in
  \emph{Proceedings of the 27th ACM International Conference on
  Multimedia}.\hskip 1em plus 0.5em minus 0.4em\relax ACM, 2019, pp. 935--944.

\bibitem{sigal2008combined}
L.~Sigal, A.~Balan, and M.~J. Black, ``Combined discriminative and generative
  articulated pose and non-rigid shape estimation,'' in \emph{Advances in
  Neural Information Processing Systems}, 2008, pp. 1337--1344.

\bibitem{guan2009estimating}
P.~Guan, A.~Weiss, A.~O. Balan, and M.~J. Black, ``Estimating human shape and
  pose from a single image,'' in \emph{Proceedings of the IEEE International
  Conference on Computer Vision}.\hskip 1em plus 0.5em minus 0.4em\relax IEEE,
  2009, pp. 1381--1388.

\bibitem{guler2019holopose}
R.~A. Guler and I.~Kokkinos, ``Holopose: Holistic 3d human reconstruction
  in-the-wild,'' in \emph{Proceedings of the IEEE Conference on Computer Vision
  and Pattern Recognition}, 2019, pp. 10\,884--10\,894.

\bibitem{rong2019delving}
Y.~Rong, Z.~Liu, C.~Li, K.~Cao, and C.~C. Loy, ``Delving deep into hybrid
  annotations for 3d human recovery in the wild,'' in \emph{Proceedings of the
  IEEE International Conference on Computer Vision}, 2019, pp. 5340--5348.

\bibitem{kanazawa2019learning}
A.~Kanazawa, J.~Y. Zhang, P.~Felsen, and J.~Malik, ``Learning 3d human dynamics
  from video,'' in \emph{Proceedings of the IEEE Conference on Computer Vision
  and Pattern Recognition}, 2019, pp. 5614--5623.

\bibitem{arnab2019exploiting}
A.~Arnab, C.~Doersch, and A.~Zisserman, ``Exploiting temporal context for 3d
  human pose estimation in the wild,'' in \emph{Proceedings of the IEEE
  Conference on Computer Vision and Pattern Recognition}, 2019, pp. 3395--3404.

\bibitem{liang2019shape}
J.~Liang and M.~C. Lin, ``Shape-aware human pose and shape reconstruction using
  multi-view images,'' in \emph{Proceedings of the IEEE International
  Conference on Computer Vision}, 2019, pp. 4352--4362.

\bibitem{pavlakos2019texturepose}
G.~Pavlakos, N.~Kolotouros, and K.~Daniilidis, ``Texturepose: Supervising human
  mesh estimation with texture consistency,'' in \emph{Proceedings of the IEEE
  International Conference on Computer Vision}, 2019, pp. 803--812.

\bibitem{kocabas2020vibe}
M.~Kocabas, N.~Athanasiou, and M.~J. Black, ``Vibe: Video inference for human
  body pose and shape estimation,'' in \emph{Proceedings of the IEEE Conference
  on Computer Vision and Pattern Recognition}, 2020, pp. 5253--5263.

\bibitem{joo2018total}
H.~Joo, T.~Simon, and Y.~Sheikh, ``Total capture: A 3d deformation model for
  tracking faces, hands, and bodies,'' in \emph{Proceedings of the IEEE
  Conference on Computer Vision and Pattern Recognition}, 2018, pp. 8320--8329.

\bibitem{pavlakos2019expressive}
G.~Pavlakos, V.~Choutas, N.~Ghorbani, T.~Bolkart, A.~A. Osman, D.~Tzionas, and
  M.~J. Black, ``Expressive body capture: 3d hands, face, and body from a
  single image,'' in \emph{Proceedings of the IEEE Conference on Computer
  Vision and Pattern Recognition}, 2019, pp. 10\,975--10\,985.

\bibitem{zhu2019detailed}
H.~Zhu, X.~Zuo, S.~Wang, X.~Cao, and R.~Yang, ``Detailed human shape estimation
  from a single image by hierarchical mesh deformation,'' in \emph{Proceedings
  of the IEEE Conference on Computer Vision and Pattern Recognition}, 2019, pp.
  4491--4500.

\bibitem{martinez2017simple}
J.~Martinez, R.~Hossain, J.~Romero, and J.~J. Little, ``A simple yet effective
  baseline for 3d human pose estimation,'' in \emph{Proceedings of the IEEE
  International Conference on Computer Vision}, 2017, pp. 2640--2649.

\bibitem{nie2017monocular}
B.~X. Nie, P.~Wei, and S.-C. Zhu, ``Monocular 3d human pose estimation by
  predicting depth on joints,'' in \emph{Proceedings of the IEEE International
  Conference on Computer Vision}.\hskip 1em plus 0.5em minus 0.4em\relax IEEE,
  2017, pp. 3467--3475.

\bibitem{moreno20173d}
F.~Moreno-Noguer, ``3d human pose estimation from a single image via distance
  matrix regression,'' in \emph{Proceedings of the IEEE Conference on Computer
  Vision and Pattern Recognition}, 2017, pp. 2823--2832.

\bibitem{lee2018propagating}
K.~Lee, I.~Lee, and S.~Lee, ``Propagating lstm: 3d pose estimation based on
  joint interdependency,'' in \emph{Proceedings of the European Conference on
  Computer Vision}.\hskip 1em plus 0.5em minus 0.4em\relax Cham: Springer,
  2018, pp. 119--135.

\bibitem{dibra2016hs}
E.~Dibra, H.~Jain, C.~{\"O}ztireli, R.~Ziegler, and M.~Gross, ``Hs-nets:
  Estimating human body shape from silhouettes with convolutional neural
  networks,'' in \emph{International Conference on 3D Vision}.\hskip 1em plus
  0.5em minus 0.4em\relax IEEE, 2016, pp. 108--117.

\bibitem{smith2019towards}
B.~M. Smith, V.~Chari, A.~Agrawal, J.~M. Rehg, and R.~Sever, ``Towards accurate
  3d human body reconstruction from silhouettes,'' in \emph{International
  Conference on 3D Vision}.\hskip 1em plus 0.5em minus 0.4em\relax IEEE, 2019,
  pp. 279--288.

\bibitem{shotton2012efficient}
J.~Shotton, R.~Girshick, A.~Fitzgibbon, T.~Sharp, M.~Cook, M.~Finocchio,
  R.~Moore, P.~Kohli, A.~Criminisi, A.~Kipman \emph{et~al.}, ``Efficient human
  pose estimation from single depth images,'' \emph{IEEE Transactions on
  Pattern Analysis and Machine Intelligence}, vol.~35, no.~12, pp. 2821--2840,
  2013.

\bibitem{gabeur2019moulding}
V.~Gabeur, J.-S. Franco, X.~Martin, C.~Schmid, and G.~Rogez, ``Moulding humans:
  Non-parametric 3d human shape estimation from single images,'' in
  \emph{Proceedings of the IEEE International Conference on Computer Vision},
  2019, pp. 2232--2241.

\bibitem{tekin2017learning}
B.~Tekin, P.~M{\'a}rquez-Neila, M.~Salzmann, and P.~Fua, ``Learning to fuse 2d
  and 3d image cues for monocular body pose estimation,'' in \emph{Proceedings
  of the IEEE International Conference on Computer Vision}, 2017, pp.
  3941--3950.

\bibitem{pavlakos2017coarse}
G.~Pavlakos, X.~Zhou, K.~G. Derpanis, and K.~Daniilidis, ``Coarse-to-fine
  volumetric prediction for single-image 3d human pose,'' in \emph{Proceedings
  of the IEEE Conference on Computer Vision and Pattern Recognition}, 2017, pp.
  7025--7034.

\bibitem{varol2018bodynet}
G.~Varol, D.~Ceylan, B.~Russell, J.~Yang, E.~Yumer, I.~Laptev, and C.~Schmid,
  ``Bodynet: Volumetric inference of 3d human body shapes,'' in
  \emph{Proceedings of the European Conference on Computer Vision}, 2018, pp.
  20--36.

\bibitem{jackson20183d}
A.~S. Jackson, C.~Manafas, and G.~Tzimiropoulos, ``3d human body reconstruction
  from a single image via volumetric regression,'' in \emph{Proceedings of the
  European Conference on Computer Vision}, 2018.

\bibitem{zheng2019deephuman}
Z.~Zheng, T.~Yu, Y.~Wei, Q.~Dai, and Y.~Liu, ``Deephuman: 3d human
  reconstruction from a single image,'' in \emph{Proceedings of the IEEE
  International Conference on Computer Vision}, 2019, pp. 7739--7749.

\bibitem{luo2018orinet}
C.~Luo, X.~Chu, and A.~L. Yuille, ``Orinet: {A} fully convolutional network for
  3d human pose estimation,'' in \emph{British Machine Vision Conference},
  2018, p.~92.

\bibitem{xiang2019monocular}
D.~Xiang, H.~Joo, and Y.~Sheikh, ``Monocular total capture: Posing face, body,
  and hands in the wild,'' in \emph{Proceedings of the IEEE Conference on
  Computer Vision and Pattern Recognition}, 2019, pp. 10\,965--10\,974.

\bibitem{alp2018densepose}
R.~Alp~G{\"u}ler, N.~Neverova, and I.~Kokkinos, ``Densepose: Dense human pose
  estimation in the wild,'' in \emph{Proceedings of the IEEE Conference on
  Computer Vision and Pattern Recognition}, 2018, pp. 7297--7306.

\bibitem{kolotouros2019convolutional}
N.~Kolotouros, G.~Pavlakos, and K.~Daniilidis, ``Convolutional mesh regression
  for single-image human shape reconstruction,'' in \emph{Proceedings of the
  IEEE Conference on Computer Vision and Pattern Recognition}, 2019, pp.
  4501--4510.

\bibitem{xu2019denserac}
Y.~Xu, S.-C. Zhu, and T.~Tung, ``Denserac: Joint 3d pose and shape estimation
  by dense render-and-compare,'' in \emph{Proceedings of the IEEE International
  Conference on Computer Vision}, 2019, pp. 7760--7770.

\bibitem{tan2018indirect}
V.~Tan, I.~Budvytis, and R.~Cipolla, ``Indirect deep structured learning for 3d
  human body shape and pose prediction,'' in \emph{British Machine Vision
  Conference}, 2017, pp. 1--11.

\bibitem{kolotouros2019learning}
N.~Kolotouros, G.~Pavlakos, M.~J. Black, and K.~Daniilidis, ``Learning to
  reconstruct 3d human pose and shape via model-fitting in the loop,'' in
  \emph{Proceedings of the IEEE International Conference on Computer Vision},
  2019, pp. 2252--2261.

\bibitem{sun2019human}
Y.~Sun, Y.~Ye, W.~Liu, W.~Gao, Y.~Fu, and T.~Mei, ``Human mesh recovery from
  monocular images via a skeleton-disentangled representation,'' in
  \emph{Proceedings of the IEEE International Conference on Computer Vision},
  2019, pp. 5349--5358.

\bibitem{kipf2017semi}
T.~N. Kipf and M.~Welling, ``Semi-supervised classification with graph
  convolutional networks,'' in \emph{International Conference on Learning
  Representations}, 2017, pp. 1--14.

\bibitem{yao2019densebody}
P.~Yao, Z.~Fang, F.~Wu, Y.~Feng, and J.~Li, ``Densebody: Directly regressing
  dense 3d human pose and shape from a single color image,'' \emph{arXiv
  preprint arXiv:1903.10153}, 2019.

\bibitem{pishchulin2013poselet}
L.~Pishchulin, M.~Andriluka, P.~Gehler, and B.~Schiele, ``Poselet conditioned
  pictorial structures,'' in \emph{Proceedings of the IEEE Conference on
  Computer Vision and Pattern Recognition}, 2013, pp. 588--595.

\bibitem{yang2011articulated}
Y.~Yang and D.~Ramanan, ``Articulated pose estimation with flexible
  mixtures-of-parts,'' in \emph{Proceedings of the IEEE Conference on Computer
  Vision and Pattern Recognition}.\hskip 1em plus 0.5em minus 0.4em\relax IEEE,
  2011, pp. 1385--1392.

\bibitem{zuffi2015stitched}
S.~Zuffi and M.~J. Black, ``The stitched puppet: A graphical model of 3d human
  shape and pose,'' in \emph{Proceedings of the IEEE Conference on Computer
  Vision and Pattern Recognition}, 2015, pp. 3537--3546.

\bibitem{tompson2014joint}
J.~J. Tompson, A.~Jain, Y.~LeCun, and C.~Bregler, ``Joint training of a
  convolutional network and a graphical model for human pose estimation,'' in
  \emph{Advances in Neural Information Processing Systems}, 2014, pp.
  1799--1807.

\bibitem{fang2018learning}
H.-S. Fang, Y.~Xu, W.~Wang, X.~Liu, and S.-C. Zhu, ``Learning pose grammar to
  encode human body configuration for 3d pose estimation,'' in
  \emph{Proceedings of the AAAI Conference on Artificial Intelligence}, 2018,
  pp. 6821--6828.

\bibitem{zhao2019semantic}
L.~Zhao, X.~Peng, Y.~Tian, M.~Kapadia, and D.~N. Metaxas, ``Semantic graph
  convolutional networks for 3d human pose regression,'' in \emph{Proceedings
  of the IEEE Conference on Computer Vision and Pattern Recognition}, 2019, pp.
  3425--3435.

\bibitem{akhter2015pose}
I.~Akhter and M.~J. Black, ``Pose-conditioned joint angle limits for 3d human
  pose reconstruction,'' in \emph{Proceedings of the IEEE Conference on
  Computer Vision and Pattern Recognition}, 2015, pp. 1446--1455.

\bibitem{zhou2016deep}
X.~Zhou, X.~Sun, W.~Zhang, S.~Liang, and Y.~Wei, ``Deep kinematic pose
  regression,'' in \emph{Proceedings of the European Conference on Computer
  Vision}.\hskip 1em plus 0.5em minus 0.4em\relax Cham: Springer, 2016, pp.
  186--201.

\bibitem{sun2017compositional}
X.~Sun, J.~Shang, S.~Liang, and Y.~Wei, ``Compositional human pose
  regression,'' in \emph{Proceedings of the IEEE International Conference on
  Computer Vision}, 2017, pp. 2602--2611.

\bibitem{zhou2017towards}
X.~Zhou, Q.~Huang, X.~Sun, X.~Xue, and Y.~Wei, ``Towards 3d human pose
  estimation in the wild: a weakly-supervised approach,'' in \emph{Proceedings
  of the IEEE International Conference on Computer Vision}, 2017, pp. 398--407.

\bibitem{yang20183d}
W.~Yang, W.~Ouyang, X.~Wang, J.~Ren, H.~Li, and X.~Wang, ``3d human pose
  estimation in the wild by adversarial learning,'' in \emph{Proceedings of the
  IEEE Conference on Computer Vision and Pattern Recognition}, 2018, pp.
  5255--5264.

\bibitem{srivastava2014dropout}
N.~Srivastava, G.~Hinton, A.~Krizhevsky, I.~Sutskever, and R.~Salakhutdinov,
  ``Dropout: a simple way to prevent neural networks from overfitting,''
  \emph{The Journal of Machine Learning Research}, vol.~15, no.~1, pp.
  1929--1958, 2014.

\bibitem{tompson2015efficient}
J.~Tompson, R.~Goroshin, A.~Jain, Y.~LeCun, and C.~Bregler, ``Efficient object
  localization using convolutional networks,'' in \emph{Proceedings of the IEEE
  Conference on Computer Vision and Pattern Recognition}, 2015, pp. 648--656.

\bibitem{ghiasi2018dropblock}
G.~Ghiasi, T.-Y. Lin, and Q.~V. Le, ``Dropblock: A regularization method for
  convolutional networks,'' in \emph{Advances in Neural Information Processing
  Systems}, 2018, pp. 10\,727--10\,737.

\bibitem{devries2017improved}
T.~DeVries and G.~W. Taylor, ``Improved regularization of convolutional neural
  networks with cutout,'' \emph{arXiv preprint arXiv:1708.04552}, 2017.

\bibitem{alp2017densereg}
R.~Alp~Guler, G.~Trigeorgis, E.~Antonakos, P.~Snape, S.~Zafeiriou, and
  I.~Kokkinos, ``Densereg: Fully convolutional dense shape regression
  in-the-wild,'' in \emph{Proceedings of the IEEE Conference on Computer Vision
  and Pattern Recognition}, 2017, pp. 6799--6808.

\bibitem{loper2014opendr}
M.~M. Loper and M.~J. Black, ``Opendr: An approximate differentiable
  renderer,'' in \emph{Proceedings of the European Conference on Computer
  Vision}.\hskip 1em plus 0.5em minus 0.4em\relax Cham: Springer, 2014, pp.
  154--169.

\bibitem{kato2018neural}
H.~Kato, Y.~Ushiku, and T.~Harada, ``Neural 3d mesh renderer,'' in
  \emph{Proceedings of the IEEE Conference on Computer Vision and Pattern
  Recognition}, 2018, pp. 3907--3916.

\bibitem{jaderberg2015spatial}
M.~Jaderberg, K.~Simonyan, A.~Zisserman \emph{et~al.}, ``Spatial transformer
  networks,'' in \emph{Advances in Neural Information Processing Systems},
  2015, pp. 2017--2025.

\bibitem{sun2018integral}
X.~Sun, B.~Xiao, F.~Wei, S.~Liang, and Y.~Wei, ``Integral human pose
  regression,'' in \emph{Proceedings of the European Conference on Computer
  Vision}, 2018, pp. 529--545.

\bibitem{he2016deep}
K.~He, X.~Zhang, S.~Ren, and J.~Sun, ``Deep residual learning for image
  recognition,'' in \emph{Proceedings of the IEEE Conference on Computer Vision
  and Pattern Recognition}, 2016, pp. 770--778.

\bibitem{zhou2019continuity}
Y.~Zhou, C.~Barnes, J.~Lu, J.~Yang, and H.~Li, ``On the continuity of rotation
  representations in neural networks,'' in \emph{Proceedings of the IEEE
  Conference on Computer Vision and Pattern Recognition}, 2019, pp. 5745--5753.

\bibitem{yan2018spatial}
S.~Yan, Y.~Xiong, and D.~Lin, ``Spatial temporal graph convolutional networks
  for skeleton-based action recognition,'' in \emph{Proceedings of the AAAI
  Conference on Artificial Intelligence}, 2018, pp. 7444--7452.

\bibitem{wang2020deep}
J.~Wang, K.~Sun, T.~Cheng, B.~Jiang, C.~Deng, Y.~Zhao, D.~Liu, Y.~Mu, M.~Tan,
  X.~Wang \emph{et~al.}, ``Deep high-resolution representation learning for
  visual recognition,'' \emph{IEEE Transactions on Pattern Analysis and Machine
  Intelligence}, 2020.

\bibitem{lin2014microsoft}
T.-Y. Lin, M.~Maire, S.~Belongie, J.~Hays, P.~Perona, D.~Ramanan,
  P.~Doll{\'a}r, and C.~L. Zitnick, ``Microsoft coco: Common objects in
  context,'' in \emph{Proceedings of the European Conference on Computer
  Vision}.\hskip 1em plus 0.5em minus 0.4em\relax Cham: Springer, 2014, pp.
  740--755.

\bibitem{kingma2014adam}
D.~P. Kingma and J.~Ba, ``Adam: A method for stochastic optimization,''
  \emph{International Conference on Learning Representations}, 2015.

\bibitem{paszke2019pytorch}
A.~Paszke, S.~Gross, F.~Massa, A.~Lerer, J.~Bradbury, G.~Chanan, T.~Killeen,
  Z.~Lin, N.~Gimelshein, L.~Antiga \emph{et~al.}, ``Pytorch: An imperative
  style, high-performance deep learning library,'' in \emph{Advances in Neural
  Information Processing Systems}, 2019, pp. 8024--8035.

\bibitem{ionescu2014human3}
C.~Ionescu, D.~Papava, V.~Olaru, and C.~Sminchisescu, ``Human3. 6m: Large scale
  datasets and predictive methods for 3d human sensing in natural
  environments,'' \emph{IEEE Transactions on Pattern Analysis and Machine
  Intelligence}, vol.~36, no.~7, pp. 1325--1339, 2014.

\bibitem{loper2014mosh}
M.~Loper, N.~Mahmood, and M.~J. Black, ``Mosh: Motion and shape capture from
  sparse markers,'' \emph{ACM Transactions on Graphics}, vol.~33, no.~6, p.
  220, 2014.

\bibitem{johnson2010clustered}
S.~Johnson and M.~Everingham, ``Clustered pose and nonlinear appearance models
  for human pose estimation.'' in \emph{British Machine Vision Conference},
  2010, pp. 12.1--12.11.

\bibitem{johnson2011learning}
------, ``Learning effective human pose estimation from inaccurate
  annotation,'' in \emph{Proceedings of the IEEE Conference on Computer Vision
  and Pattern Recognition}.\hskip 1em plus 0.5em minus 0.4em\relax IEEE, 2011,
  pp. 1465--1472.

\bibitem{andriluka20142d}
M.~Andriluka, L.~Pishchulin, P.~Gehler, and B.~Schiele, ``2d human pose
  estimation: New benchmark and state of the art analysis,'' in
  \emph{Proceedings of the IEEE Conference on Computer Vision and Pattern
  Recognition}, 2014, pp. 3686--3693.

\bibitem{dantone2014body}
M.~Dantone, J.~Gall, C.~Leistner, and L.~Van~Gool, ``Body parts dependent joint
  regressors for human pose estimation in still images,'' \emph{IEEE
  Transactions on Pattern Analysis and Machine Intelligence}, vol.~36, no.~11,
  pp. 2131--2143, 2014.

\bibitem{von2018recovering}
T.~von Marcard, R.~Henschel, M.~J. Black, B.~Rosenhahn, and G.~Pons-Moll,
  ``Recovering accurate 3d human pose in the wild using imus and a moving
  camera,'' in \emph{Proceedings of the European Conference on Computer
  Vision}, 2018, pp. 601--617.

\bibitem{mehta2017monocular}
D.~Mehta, H.~Rhodin, D.~Casas, P.~Fua, O.~Sotnychenko, W.~Xu, and C.~Theobalt,
  ``Monocular 3d human pose estimation in the wild using improved cnn
  supervision,'' in \emph{International Conference on 3D Vision}.\hskip 1em
  plus 0.5em minus 0.4em\relax IEEE, 2017, pp. 506--516.

\bibitem{cao2019openpose}
Z.~Cao, G.~H. Martinez, T.~Simon, S.-E. Wei, and Y.~A. Sheikh, ``Openpose:
  Realtime multi-person 2d pose estimation using part affinity fields,''
  \emph{IEEE Transactions on Pattern Analysis and Machine Intelligence}, 2019.

\bibitem{xiao2018simple}
B.~Xiao, H.~Wu, and Y.~Wei, ``Simple baselines for human pose estimation and
  tracking,'' in \emph{Proceedings of the European Conference on Computer
  Vision}, 2018, pp. 466--481.

\bibitem{sun2019deep}
K.~Sun, B.~Xiao, D.~Liu, and J.~Wang, ``Deep high-resolution representation
  learning for human pose estimation,'' in \emph{Proceedings of the IEEE
  Conference on Computer Vision and Pattern Recognition}, 2019, pp. 5693--5703.

\bibitem{doersch2019sim2real}
C.~Doersch and A.~Zisserman, ``Sim2real transfer learning for 3d human pose
  estimation: motion to the rescue,'' in \emph{Advances in Neural Information
  Processing Systems}, 2019, pp. 12\,949--12\,961.

\bibitem{deng2009imagenet}
J.~Deng, W.~Dong, R.~Socher, L.-J. Li, K.~Li, and L.~Fei-Fei, ``Imagenet: A
  large-scale hierarchical image database,'' in \emph{Proceedings of the IEEE
  Conference on Computer Vision and Pattern Recognition}.\hskip 1em plus 0.5em
  minus 0.4em\relax IEEE, 2009, pp. 248--255.

\end{thebibliography}
